\pdfoutput=1

\documentclass{article}

\usepackage{arxiv}

\usepackage{pdfpages}
\usepackage{anyfontsize}
\usepackage{setspace}
\usepackage{amsmath}
\usepackage{amssymb}
\usepackage{siunitx}
\usepackage{makecell}
\usepackage{xfrac}
\usepackage{algorithm}
\usepackage[noend]{algpseudocode}
\usepackage{listings}
\usepackage{caption}
\usepackage{subcaption}
\usepackage[toc,page]{appendix}
\usepackage[toc,shortcuts]{glossaries}

\usepackage[utf8]{inputenc} 
\usepackage[T1]{fontenc}    
\usepackage{hyperref}       
\usepackage{url}            
\usepackage{booktabs}       
\usepackage{amsfonts}       
\usepackage{nicefrac}       
\usepackage{microtype}      
\usepackage{graphicx}
\usepackage{doi}
\usepackage{natbib}

\title{A novel evaluation methodology for supervised Feature Ranking algorithms}

\author{ \href{https://orcid.org/0000-0003-3304-3800}{\includegraphics[scale=0.06]{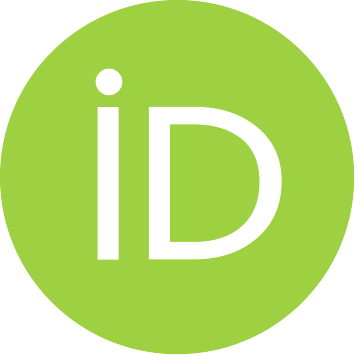}\hspace{1mm}Jeroen G. S. Overschie}\thanks{Acknowledgements go to George Azzopardi and Ahmad Alsahaf.} \\
	Faculty of Science and Engineering\\
	University of Groningen\\
	Groningen, 9747 AG \\
	\texttt{j.g.s.overschie@student.rug.nl} \\
}

\date{July 1, 2021}

\renewcommand{\headeright}{Technical Report}
\renewcommand{\undertitle}{Technical Report}
\renewcommand{\shorttitle}{A novel evaluation methodology}

\hypersetup{
pdftitle={A novel evaluation methodology for supervised Feature Ranking algorithms},
pdfsubject={cs.SE},
pdfauthor={Jeroen G. S.~Overschie},
pdfkeywords={Feature Ranking, Feature Selection, Evaluation Metrics, Benchmark, Machine Learning},
}

\makenoidxglossaries

\newacronym{rfe}{RFE}{Recursive Feature Elimination}
\newacronym{cv}{CV}{Cross-Validation}
\newacronym{dt}{DT}{Decision Tree}
\newacronym{lasso}{LASSO}{Least Absolute Shrinkage and Selection Operator}
\newacronym{pca}{PCA}{Principle Component Analysis}
\newacronym{lda}{LDA}{Linear Discriminant Analysis}
\newacronym{knn}{k-NN}{k- Nearest Neighbors}
\newacronym{nb}{NB}{Naïve Bayes}
\newacronym{svd}{SVD}{Singular Value Decomposition}
\newacronym{ml}{ML}{Machine Learning}
\newacronym{ai}{AI}{Artificial Intelligence}
\newacronym{pcc}{PCC}{Pearson product-moment Correlation Coefficient}
\newacronym{hpc}{HPC}{High-Performance Computing}
\newacronym{gt}{GT}{Ground-Truth}

\newglossaryentry{rq}
{
    name={RQ},
    description={``RQ (Redis Queue) is a simple Python library for queueing jobs and processing them in the background with workers. It is backed by Redis and it is designed to have a low barrier to entry.'' - \href{source}{https://python-rq.org/}}
}

\newglossaryentry{apriori}
{
    name={apriori},
    description={\textit{Apriori} information refers to any possible information about the dataset feature importances one has before running any learning algorithms. Common ways to obtain such information are human domain experts or the usage of synthetic datasets}
}

\newglossaryentry{ib1}
{
  name={IB1},
  description={Instance-based classification algorithm originally proposed by Aha et al. (1991) \citep{aha_instance-based_1991}},
  first={\glsentrytext{ib1}},
  plural={IB1},
  firstplural={\glsentrytext{ib1}}
}

\newglossaryentry{part}
{
  name={PART},
  description={Partial Decision Tree algorithm},
  first={\glsentrytext{part}},
  plural={PART},
  firstplural={\glsentrytext{part}}
}

\newglossaryentry{rba}
{
  name={RBA},
  description={Relief-Based Feature Selection Algorithms},
  first={\glsentrydesc{rba} (\glsentrytext{rba})},
  plural={RBAs},
  firstplural={\glsentrydesc{rba}s (\glsentryplural{rba})}
}

\newglossaryentry{svm}
{
  name={SVM},
  description={Support Vector Machine},
  first={\glsentrydesc{svm} (\glsentrytext{svm})},
  plural={SVMs},
  firstplural={\glsentrydesc{svm}s (\glsentryplural{svm})}
}

\begin{document}
\maketitle

\begin{abstract}
Both in the domains of Feature Selection and Interpretable AI, there exists a desire to `rank' features based on their importance. Such feature importance rankings can then be used to either: (1) reduce the dataset size or (2) interpret the Machine Learning model. In the literature, however, such Feature Rankers are not evaluated in a systematic, consistent way. Many papers have a different way of arguing which feature importance ranker works best. This paper fills this gap, by proposing a new evaluation methodology. By making use of synthetic datasets, feature importance scores can be known beforehand, allowing more systematic evaluation. To facilitate large-scale experimentation using the new methodology, a benchmarking framework was built in Python, called fseval. The framework allows running experiments in parallel and distributed over machines on HPC systems. By integrating with an online platform called Weights and Biases, charts can be interactively explored on a live dashboard. The software was released as open-source software, and is published as a package on the PyPi platform. The research concludes by exploring one such large-scale experiment, to find the strengths and weaknesses of the participating algorithms, on many fronts.
\end{abstract}

\keywords{Feature Ranking, Feature Selection, Evaluation Metrics, Benchmark, Machine Learning}

\clearpage

\tableofcontents
\newpage

\section{Introduction}\label{section:introduction}

In this day and age, more data is available than ever before in many domains \citep{sagiroglu_big_2013}. In the biomedical domain sensory devices such as MRI or PET scanners are getting ever more accurate - requiring more storage space to store higher resolution data. In the financial domain, markets are operating at increasingly low time intervals - requiring storage and analysis of data at a higher time resolution than before. The internet too, sees increasing amounts of traffic world-wide and is producing immense amounts of data every day. Applications of Machine Learning are common in all these domains: training predictive models by learning from example data is able to give us interesting insights that can improve both economy and the quality of human lives. By the nature of models that learn from examples, performance is often better when a larger amount of examples is available. But it shall get clear that larger quantities of data presents itself not as solely beneficial; but rather- a mixed blessing.

The field of Machine Learning has seen vast increases in dataset sizes: both in terms of sample size and amount of dimensions. Although the availability of more data presents practitioners with opportunities to create better performing models, more data will have to be processed - causing an increased computational burden in the learning process. This increase is nonlinear: due to the curse of dimensionality, the computational burden can get large, quickly. Even though the field has for long relied on computer processing speed steadily increasing, obeying Moore's law, the rate of advancement will inevitably start declining - and in fact already has \citep{theis_end_2017}. Self-evidently, many technological advancements can still be realised, either in the silicon world or in a post-silicon world, in which perhaps forms of quantum computing might become predominant \citep{britt_high-performance_2017}. But what is certain, is that besides the technological opportunities the chip-making industry still has, there also exist strict physical limitations as to how fast computer processing can get. So, besides leveraging faster hardware, we are also going to have to make our software smarter. Instead of increasing just our computational power, methods for reducing the computational burden in the first place are desired. Thus, the need for preprocessing techniques and data reduction algorithms is instigated.

\textbf{Feature selection} is such a domain that focuses on reducing the overall learning processing burden \citep{guyon_introduction_2003}. By figuring out what dimensions are relevant to the learning task, which ones are redundant, and which ones are completely noisy with respect to the learning task, a smaller dataset can be obtained by means of a preprocessing step. This is contrary to the domain of dimensionality reduction, in which data is projected onto a space of smaller dimensionality - but losing the exact representation of the data distribution. In Feature Selection, we aim to obtain a binary decision about which dimensions to keep, in the original data space. Aside from reducing dataset size by cutting off dimensions, in some cases the generalization ability of a learning model can even be improved: the learning model can better learn the data distribution by using less but more meaningful dimensions with less noise. This makes the benefit of learning on a subset of the available features two-fold: model fitting and prediction can be both faster and more accurate. To select relevant dataset features, a wide variety of strategies exist. One is to assign a scoring to each dimension and keep only the most relevant ones - such a ranking is called a feature ranking.

\textbf{Feature ranking} is a broader domain in contrast to feature selection, in which the sole purpose is not to only reduce the dataset size, but to construct a hierarchical order on the importance of features given a specific learning task \citep{duch_comparison_2004}. Many techniques can be employed to create such feature rankings, of which some are substantially faster than others, but might yield sub-optimal results: the choice of a suitable feature ranking algorithm is not trivial in most scenarios. Once such a feature ranking has been constructed, it can be used for various applications. First, a feature selection can be made by removing features that rank below a certain feature importance score threshold; one such naive method would be to cut off any feature that ranks below the average feature importance score. Secondly, the feature ranking can also be used for better Machine Learning model interpretation; in which the feature importance scores help humans better understand the predictions and reasoning of the models - by knowing which features the model found important, it can better be understood how the model made the decision that it has. This second application is part of the bigger domain of Interpretable Artificial Intelligence, or more commonly \textit{Interpretable \gls{ai}}, which has in recent times become ever more relevant \citep{ghosh_interpretable_2020}. Interpretable \gls{ai} aims to explain models that were before considered `black box' models, making them more transparent to the user.

\textbf{Evaluation} on the performance of feature ranking algorithms has been conducted in many different ways. Most authors used a `validation' estimator, which was trained on a chosen subset of the dataset to then see how the estimator performed given this feature subset. A ranker is desired, then, that ranks the most predictive features highest, preferably in a reasonable amount of time. In this way, we get a feature subset that is as small as possible, that gives us the highest possible predictive power. This evaluation technique, however, might not be systematic enough. Across papers, many different validation estimators are used, which make the results across papers subsequently incomparable to one another. Researchers might also benefit from more extensive and systematic evaluation by use of synthetic datasets. By manually controlling many aspects of the dataset, such as noise levels, the complexity of the data distribution to be learned and the amount of informative features, a comprehensive evaluation on the feature ranking algorithm behavior and characteristics can be made. In this way, by employing synthetic datasets, the exact informative features to be ranked as relevant can be known \textit{\gls{apriori}}, i.e. before conducting the feature ranking operation. Such new and possibly useful metrics can be employed to evaluate feature rankings independent of any validation estimator.

\textbf{In this paper}, a comprehensive comparative experiment on feature rankers is conducted using both real-world and synthetic datasets, employing new evaluation metrics on the synthetic datasets by knowing the relevant feature apriori. Both classical and more recent feature ranking algorithms are included, using methods that originally reside in both the statistical and feature selection domains. By systematically generating synthetic datasets that are specifically designed to vary in various relevant data properties, various characteristics of the feature ranking algorithms can be assessed and estimated. To employ such a large-scale benchmark, a software framework was built to facilitate such testing. The framework was released as open-source, freely available software.

The \textbf{research question} which is to be answered is as follows. ``How do we evaluate Feature Ranking algorithms in a way that is systematic, comprehensive, and emphasizes the differences between the algorithms in a meaningful way?'' Currently, Feature Ranking and Feature Selection algorithms are evaluated in many ways. This research aims to find a methodology that is both meaningful and applicable to many algorithms.
The goal is to arrive at a scientifically sound benchmarking method for feature ranking methods, taking advantage of \textit{\gls{apriori}} information wherever possible. Up to our knowledge, no current literature exists that is aimed at the evaluation of Feature- Ranking and Selection algorithms, that takes such a comprehensive approach and proposes novel evaluation metrics. An important sub-goal in constructing such a benchmarking method is the development of a pipeline implementing such a new benchmarking method and analysing its results.

\textbf{The scope} of this research can be defined as follows. Feature Ranking methods are evaluated that work on tabular datasets and require example data including prediction targets, i.e. only \textit{supervised} algorithms are considered. Furthermore, all considered datasets are \textit{tabular}, that is, no underlying data structures such as linked- or streaming data is assumed. The considered dataset prediction tasks are \textit{regression} and \textit{classification} - which limits the considered ranking methods to these tasks as well. Eight out of fourteen classification datasets perfectly balanced classes, the other six have varying levels of imbalance (see Appendix). All considered Feature Ranking methods are \textit{global} rankers. Meaning that all methods construct rankings for the full dataset, i.e., no instance-based methods are included. Lastly, all three ranking types are supported and are considered in the research, i.e. the research considers feature importance-, feature support- and feature ranking vectors.

\textbf{The contribution} of this paper is multiple-fold. Firstly, the inclusion of many feature ranking algorithms and many datasets makes it possible to make more meaningful comparisons between feature ranking algorithms, which would not have been possible across existing papers due to the lack of a single evaluation standard. Secondly, the proposal of a new evaluation standard also makes it possible for other authors to conduct experiments in reproducible manner. This subsequently enables readers to compare results across papers - saving time but also allowing more thorough comparative analysis on which feature ranker is best suited for a given dataset. Third, the new evaluation standard was implemented and packaged in an easy-to-use open-source software package, distributed as a Python  pip package manager distribution on the PyPi platform\footnote{\href{https://pypi.org/project/fseval/}{https://pypi.org/project/fseval/}}.

Chapters in this paper are structured as follows. First, motivations for both Feature Ranking and the evaluation thereof are given in Chapter~\ref{section:motivations}. Second, an analysis on previous work in the literature is conducted, in Chapter~\ref{section:related-work}. Third, various methods for creating feature rankings are presented, such that a grasp can be obtained on the overall mechanics of the methods. This is done is Chapter~\ref{section:methods}. Fourth, an insight into how feature rankings are evaluated is gained; and a new method is presented afterwards, in Chapter~\ref{section:evaluation}. Next, comments are made on the construction of a benchmarking pipeline in Chapter~\ref{section:pipeline}. Then, this new standard is applied in an experiment, which setup and results are explained in Chapter~\ref{section:experiments}. Lastly, the paper is concluded by a discussion in Chapter~\ref{section:discussion} and a conclusion in Chapter~\ref{section:conclusion}.

\clearpage
\section{Motivations}\label{section:motivations}

\subsection{Motivations for Feature Ranking}
The desire to create a feature ranking from some features in a dataset stems from two main applications: the domain of feature selection and the domain of interpretable AI. To better understand why one would want to create a feature ranking, a brief exploration is made on both domains, motivating the concept throughout.

\subsubsection{Feature Selection}
The field of Machine Learning is plagued by the curse of dimensionality \citep{koppen_curse_2009}: as the dimensionality of datasets grows larger, the computational burden for learning algorithms gets exponentially larger. This is due to the fact that when data gets of increasingly higher dimensionality, the volume of the data shape grows faster than the amount of samples grows along, causing the data volume to become sparse.
There are many real-world domains that naturally deal with datasets of such shape. In the biomedical world, for example, collecting data samples might require arduous amounts of human effort \citep{hu_feature_2018}. There might be the case where one data sample represents one human and any such sample is very laborious to collect: but the samples that are collected are of very high dimensionality. In this case, we deal with a scenario where the amount of dimensions far exceeds the amount of samples ($n \gg p$ where $n$ is the amount of samples and $p$ the amount of dimensions).

To avoid such a curse of dimensionality, feature selection can be used. The goal is to reduce as much features as possible while retaining the most relevant information in the dataset. The benefits are a decreased computational burden; though in some cases the learning algorithm generalization performance might actually be improved over a scenario when no feature selection is used. This is most often due to the removal of noise that would have interfered with the learning process.

\begin{figure}[h]
    \centering
    \includegraphics[width=\linewidth]{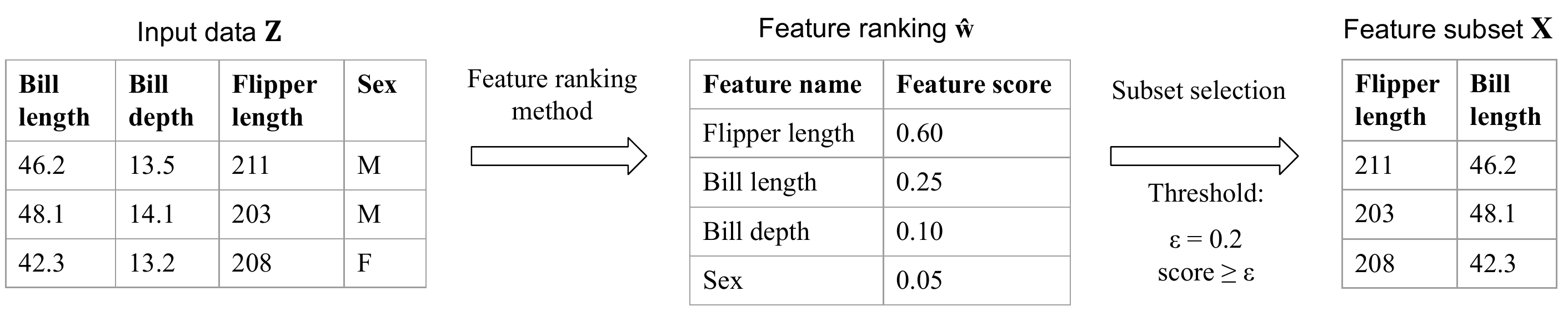}
    \caption{An example process of feature selection by thresholding a feature ranking at some point $\epsilon$. The resulting feature subset $\mathbf{X}$ presumably contains the strongest predictors given the target variable. Feature names come from a real dataset on antarctic penguins \citep{horst_palmerpenguins_2020}.}
    \label{fig:schematic-feature-selection}
\end{figure}

One such example process of feature selection by using a feature ranking can be seen in Figure~\ref{fig:schematic-feature-selection}. As can be seen in the figure, a ranking can be used to reduce the overall dataset size by using a \textit{threshold} operation. In such a threshold operation, any feature that ranks below the given threshold $\epsilon$ is removed. Many, if not most, feature selection methods rely on feature ranking methods under the hood to create a feature subset of reduced size.

For this reason, the more general problem of constructing feature rankings is omni-relevant in the feature selection domain. To go from feature ranking to feature selection, the only extra step necessary is a sufficient threshold point, which many algorithms arrive at using different methods. A straightforward way to cut off low-ranking features is to exclude any feature that ranks below the mean score - or, alternatively, exclude only the highest ranking quartile, etcetera. More sophisticated schemes exist, however. One might also choose to iteratively run a feature ranking process and remove the lowest ranking score in each iteration. Such a process is called \gls{rfe}. That said, it is clear that feature ranking and feature selection go in unison.

\subsubsection{Interpretable Artificial Intelligence}
The domain of Interpretable AI has arisen due to the need to better understand and reason about the decisions that any Machine Learning model makes. Traditionally, the sole goal in building a learning algorithm would be to best predict some target variable, given some set of samples to learn from. With the advent of sophisticated learning algorithms such as Neural Networks and especially Deep Neural Networks, however, the many computational layers that separate inputs from answers tend to obfuscate the decision process \citep{rai_explainable_2020}. Whereas in classical statistical models practitioners sought to unveil a clear relationship between the independent- and dependent variables, some models in the field of Machine Learning have grown so complicated that no human can reason on its output. Because such algorithms tend to compare metaphorically to a black box one cannot possibly see through, one also refers to such algorithms as \textit{black boxes}.

Whilst at the same time models have gotten progressively more sophisticated and thus less transparent, the market for deploying Machine Learning has been steadily getting bigger. Many institutions seek to benefit from the possibilities of recent developments in \gls{ai} and \gls{ml} - including many companies, schools or the government. In all such applications, for every decision made by a Machine Learning model, an argumentation as how the model came to such a conclusion is desired. Even, in some scenarios the decision to be made at hand weighs so heavily that a \gls{ml} model that cannot explain itself might not be usable at all. For example, one might deploy a \gls{ml} model that scores employees based on their performance. The goal is then to fire the lowest ranking employees and keep highest performing ones. But fully trusting such a model might be a dangerous practice. If a model provides only bare explanations on how it came to such a decision, employers might have a hard time interpreting the model and employees might have a hard time understanding its decisions. As a matter of fact, faulty models of such kind have already been deployed out in the open - with considerable complications with respect to model transparency as a result \citep{oneil_weapons_2016}.

For this reason, decisions made by computers are required to be explainable - allowing practitioners to assess not only \gls{ml} model decisions, but also assess the overall usefulness of the model in general. After all, many learning models are designed such to always generate a decision, no matter how noisy the input. That said, the black box of AI can generally be opened up in two ways. First, one could opt for a less sophisticated in the first place - one that reveals its inner workings and supports its decisions by an elaborate scheme, much in line with classical statistics. Although a simpler model could sometimes suffice in places where practitioners currently opt for more complicated ones, a solution for models of any kind of flexibility is desired. Therefore, a second option is to delve into any such model to reveal a reasoning about its decision-making process - which is the option that the field of Interpretable AI bothers itself with.

\begin{figure}[ht]
    \centering
    \includegraphics[width=\linewidth]{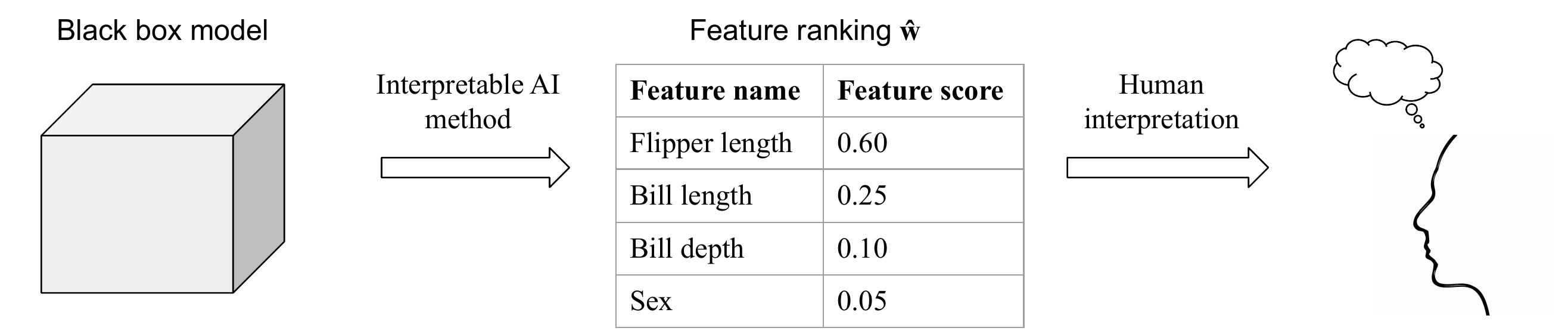}
    \caption{An example process of Interpretable AI. In this case, the input to the interpretable AI algorithm is a trained black box model, which has an otherwise hard to understand decision making process. The interpretable AI algorithm then extracts feature importance scores for better interpretability.}
    \label{fig:schematic-interpretable-ai}
\end{figure}

Feature rankings are one of the facets which can be used to facilitate a better understanding of a Machine Learning model \citep{hansen_interpretability_2019}: by unveiling which variables the \gls{ml} model finds important, a better understanding can be gained on its decision-making process. In some scenarios, it might, for example, be the case that the Machine Learning model in question unexpectedly weighs a feature as very important, even though a human expert can know apriori that the feature is not of value to the predictive task at hand. In this way, faulty models can be detected and prevented from being used in critical settings. In the Interpretable AI jargon, the process of scoring and explaining feature importance goes under different names. In the literature, feature rankings are referred to as \textit{feature impact} scores, \textit{feature importance} scores, or \textit{feature effects}, which are synonymous in this case. All terms indicate the process of quantifying feature relevance with respect to the learning task at hand. The general process of an interpretable AI algorithm can be seen as in Figure~\ref{fig:schematic-interpretable-ai}.

As could be seen in Figure~\ref{fig:schematic-interpretable-ai}, one way of making a black box model more interpretable is to extract a feature ranking from the trained model - an interpretable AI algorithm reads in the black box model and tries to explain it \citep{lundberg_unified_2017}. Other interpretable AI methods take a more direct approach \citep{arik_tabnet_2020}, in which the interpretable AI algorithm functions as a prediction method itself. The latter approach is employed only in more recent years; in which learning algorithms are designed from the start to be interpretable. However, independent of whether a direct- or indirect approach to explaining the decision process is taken, the desire to rank features is applicable to both. That said, we can therefore conclude that the construction of truthful feature rankings are an important aspect in interpretable AI.

\subsection{Motivations for Evaluating Feature Ranking methods}
Given the need for Feature Ranking methods, the desire to evaluate such methods systematically is clear. Practitioners throughout the field of Machine Learning desire to know the behavior of different methods under a large range of varying conditions, such as the sample size-, dimensionality-, or domain of the dataset. One also needs to take into account the available computational resources and the desired degree of accuracy of the solution. Finding out what evaluation methods and metrics best represent the optimality of the solution is essential, for else the presentation of scientific findings might be misleading compared to real-world performance.

In this way, individuals and institutions are motivated to actively research Feature Ranking methods and evaluate them such to determine which method works best, under what conditions. As is more often the case in the comparison of Machine Learning algorithms, however, is the fact that methods are not always easily analytically compared. That is, solely a theoretical analysis does not suffice to create a meaningful comparison. Although one can reason about the performance of many methods before conducting concrete data analyses, a real-world experiment is essential to any reputable publication.

\clearpage
\section{Related work}\label{section:related-work}
The domain of feature ranking and selection has a large availability of literature, spread out over many subtopics. What is more rare, however, is to find papers that explicitly research and reason about the usage of certain evaluation metrics. In general, papers tend to stick to a certain evaluation method when the majority employs the given technique - but a chance for conducting a more thorough analysis might be missed nonetheless.

In the feature selection domain, the evaluation and comparison of feature selection algorithms is a non-trivial problem. Among a wide range of metrics, no consensus exists among researchers, leaving many papers to present outcomes in different ways \citep{guyon_introduction_2003}. In absence of a single consistent evaluation pipeline across the field, many scholars adhere to methods that are `widely used' \citep{solorio-fernandez_review_2020} \citep{li_feature_2017}.

Recommendations for metrics have been given in previous papers, most often when discussing future work. Arguments are made for relevant aspects to evaluate, such as in Chandrashekar (2014) \citep{chandrashekar_survey_2014}:

\begin{quote}\textit{"a feature selection algorithm can be selected based on the following considerations: simplicity, stability, number of reduced features, classification accuracy, storage and computational requirements"}\end{quote}

Of these aspects most proposals focus mainly on number of reduced features, classification accuracy and computational requirements. In the regression case, the classification accuracy would be replaced by a commonly used regression counterpart, the R2-score. Let us explore what validation estimators and corresponding metrics are used in papers across the field. Afterwards, evaluation aspects are covered that are not present in the aforementioned set.

\subsection{Evaluation metrics}
A \textbf{validation estimator} is often used to evaluate supervised feature selection methods; assessing the quality of a feature subset by running some predictor over the feature subset selected by the feature selection algorithm, obtaining the easily interpretable \textbf{prediction accuracy} metric in the classification case. Predictors often used in the literature include \gls{knn} \citep{al-tashi_review_2020} \citep{mafarja_dragonfly_2020}, \glspl{svm} \citep{chandrashekar_survey_2014}, \glspl{dt} \citep{li_feature_2017} and \gls{nb} \citep{koller_toward_1996}. Metrics used are often classification accuracy or in some cases average error rate \citep{khurma_evolopy-fs_2020}, validated using some $n$-fold cross validation, commonly 5- or 10-fold. See Table~\ref{table:evaluation-metrics-table}.

\renewcommand\theadalign{bl}
\begin{table}[ht]
    \centering
    \begin{tabular}{| l | l | l | l | l | l | l |}
    \hline
    \thead{Name method} & \thead{Validation \\ estimators} & \thead{Acc-\\uracy} & \thead{Stab-\\ility} & \thead{Time\\ $t(s)$} & \thead{Time\\ $\Omega(n)$} & \thead{Apriori\\info} \\
    \hline
    FOCUS \citep{almuallim_learning_1991}                       & \gls{dt}                              & \checkmark              &                &                & \checkmark               &                   \\
    \hline
    Relief \citep{kira_feature_1992}                      & \gls{dt}                              & \checkmark              &                & \checkmark               & \checkmark               &                   \\
    \hline
    Relief-F \citep{kononenko_estimating_1994}                    & \makecell[tl]{\acs{pcc}} &               &                &                &                & \checkmark                  \\
    \hline
    INTERACT \citep{zhao_searching_2007}                    & \gls{dt}, \gls{svm}                         & \checkmark              &                & \checkmark               & \checkmark               &                   \\
    \hline
    Fisher \citep{gu_generalized_2012}                      & \gls{knn}                            & \checkmark              &                &                & \checkmark               &                   \\
    \hline
    MutInf \citep{zaffalon_robust_2014}                      & \gls{nb}                              & \checkmark              & \checkmark               & \checkmark               & \checkmark               &                   \\
    \hline
    \makecell[tl]{Joint MutInf Maximization \citep{bennasar_feature_2015}}   & \gls{nb}, \gls{knn}                        & \checkmark              & \checkmark               &                &                &                   \\
    \hline
    \makecell[tl]{Interaction Weight-based FS \citep{zeng_novel_2015}} & \makecell[tl]{\gls{dt}, \gls{ib1}, \gls{part}}                   & \checkmark              &                & \checkmark               & \checkmark               & \checkmark                  \\
    \hline
    \makecell[tl]{Infinite FS \citep{roffo_infinite_2015}}  & \gls{svm}                             & \checkmark              & \checkmark               & \checkmark               & \checkmark               &                   \\
    \hline
    MultiSURF \citep{urbanowicz_relief-based_2018}                   &           -                      &               &                &                & \checkmark               & \checkmark    \\
    \hline     
    \end{tabular}
    \caption{A comparison table of evaluation metrics used in Feature- Ranking or Selection paper proposals. Many different evaluation metrics are used, illustrating there exist no consensus on definite meaningful evaluation metrics in the field.}
    \label{table:evaluation-metrics-table}
\end{table}

\textbf{Stability}, on the other hand, is not widely used in theoretical or quantitative argumentation. The stability is defined as the ability of an algorithm to produce consistent results given small changes in the sample data. In this context, this can be phrased as the sensitivity of the feature selector to data perturbations. Even though stability was recommended as a relevant evaluation metric by Chandrashekar (2014), not many papers explicitly argue for the stability of their method; the metric is called an \textit{``overlooked problem''} in Chandrashekar (2014). In many papers this metric is still regarded as a future work for solidifying any experimental results - the development of algorithms that achieve both high classification accuracy and high stability is still seen as a `challenging' problem by Tang et al (2015) \citep{tang_feature_2014}.

The trend seems to be turning though, with more authors becoming aware of the importance of stability. Our reliance on machine learning is ever-increasing, and so does the demand for interpretability and reliability of the algorithms. Take for example a biomedical application, in which feature selection is used to select genes in a gene sequencing analysis. Any expert in this domain field will feel more confident if an algorithm produces stable results given a varying sample population. In this way, algorithm stability is of much relevance to real-world applications of \gls{ml}. Stability has been long taken into account into prediction tasks, but not so much in feature selection - see for example Table~\ref{table:evaluation-metrics-table}.

\textbf{Scalability} is another point of interest that only recently caught more attention. The extra demand of algorithms to allow for parallel execution has been imminent as data grew tremendously large. Even, multi-core processing can lack in terms of performance, hence introducing the need for algorithms that can run in distributed fashion. Distributing a dataset over multiple machines poses challenges to some existing methods, though. Some current methods require the full dimensionality of the dataset to be available in-memory \citep{tang_feature_2014}. Yet, other methods require each sample to be visited multiple times, e.g., to apply a sample re-weighting strategy in order to converge. For these reasons, distributing any dataset workload on to multiple workers is a non-trivial problem; no generalized solution exists for cutting the dataset into chunks.

It is up to individual algorithms to find suitable ways of supporting parallel solutions and more importantly, support cases where data is too large to fit in-memory, i.e., apply distributed computing. Recent strategies overcome the issue of working with fewer samples by retaining only those samples that are most representative of the data - eliminating the need for working with a full sample population. Although few in number, there exist proposals for distributed dimensionality reduction methods \citep{li_distributed_2020}, using \textit{divide-and-conquer} techniques. Aggregating disjoint results would make for a performance similar to that of a centralized solution.

\subsection{Synthetic datasets and apriori information}
\textbf{Synthetic datasets} are employed in many papers in the literature. Whereas some datasets are injected by synthetically generated probe variables, others use completed generated benchmark datasets, such that benchmarks can be conducted in an even more controlled environment.
About such partially synthetic datasets has been spoken of in literature since long, using datasets that are real but altered by injecting more data. In \citep{guyon_introduction_2003} the authors speak of \lq probe variables', which are used to discard any variable that scores lower than any of the probes. If the probe variables are set to be random variables, a simple way is obtained to apply a threshold to cut off features from the selected feature subset, i.e., by introducing known noise into the dataset we can construct more thoughtful cut-off thresholds.

Completely synthetic datasets, on the other hand, can allow for more sophisticated metrics to be used. Possibilities for new evaluation metrics are, for example described in \citep{solorio-fernandez_review_2020}: \textit{``Evaluation in terms of the redundancy of the selected features''} and \textit{``Evaluation in terms of the correctness of the selected features''}, the latter of which requires us to know what features are informative \textit{a priori} - which is accomplished with synthetic generation. Controlling all facets relevant to the quantitative analysis manually makes for a \textit{Simulation study}, which is argued for in \citep{urbanowicz_benchmarking_2018} as follows:

\begin{quote}
    \textit{``Simulation studies such as these facilitate proper evaluation and comparison of methodologies because a simulation study can be designed by systematically varying key experimental conditions, and the ground truth of the dataset is known i.e. we know which features are relevant vs. irrelevant, we know the pattern of association between relevant features and endpoint, and we know how much signal is in the dataset.''}
\end{quote}

Indeed, there seems to be a trend toward including synthetically generated datasets in experiments. In a review paper \citep{bolon-canedo_review_2013} the authors argue that synthetically generated datasets can yield statistically sound results because of the fact no inherent noise or redundancy will obstruct the experiment process. In other papers simulation studies are conducted as well \citep{cai_online_2020} \citep{tang_high-dimensional_2020} \citep{li_distributed_2020}, concluded by a small section depicting a `real data' analysis to conclude the point. For these reasons, a recommendation is made to include simulation studies in any comprehensive benchmark on feature ranking methods.

\subsection{The gap in the current literature}
With respect to the above summarized works, there are some aspects missing in the evaluation process. Because this research is set out to fill in the highlighted missing parts, it is important to get a clear idea of the entire set of missing aspects. Therefore, the concerned \textit{literature `gap'} is summarized as follows.

Many papers in the literature evaluate Feature Ranking and Feature Selection algorithms in different ways. Many validation estimators are used across papers, causing the results to become incomparable. Moreover, not every paper evaluates the stability or scalability of the algorithm, like shown in Table~\ref{table:evaluation-metrics-table}. Whereas the stability is a quantification of the algorithm's robustness against data permutations, the scalability means both storage- and time complexity. Lastly, there exist opportunities for systematic evaluation using \textit{\gls{apriori}} information on the feature importances. Few authors currently utilise this opportunity.

Therefore, the above problems are to be addressed in this paper. A concrete quantification of the algorithm's stability, scalability and performance is desired. Thereby, also synthetic data is considered, in which the relevant features are known \gls{apriori}. This paper also fills in the gaps left by some papers which do only theoretically describe their experimental setup: the new methodology was implemented in a pipeline, available as open-source software. However, first, a look is taken at how to make Feature Rankings at all.

\clearpage
\section{Methods for Feature Ranking}\label{section:methods}
In the following, general theory related to the construction of feature rankings is discussed. The theory is required to be discussed because in order to best understand the evaluation process, an understanding of the construction process is a must.

\subsection{Terminology}
Among the subject of reducing dataset dimensions, there exists a common terminology that is used among the literature. Whilst some terminology is synonymous, other seemingly related terms mean different concepts entirely.

\textbf{Feature Selection} and Feature Ranking are two terms often used interchangeably. Since often feature selection is done by first constructing a feature ranking and then cutting off features ranked lower than some threshold $\epsilon$, the construction of a feature ranking is often times an integral part of feature selection. Although feature selection methods exist that do not construct feature rankings, the two are synchronous in many ways, and are in this paper thus related to one another.

The term Feature Selection is used to indicate the general process of obtaining a feature subset with reduced size without transforming the data. Note that any feature selection method might transform the data in the algorithm however it likes internally - the stated terminology is only concerned with the eventual output of the feature selection algorithm. In this paper, a broad perspective is taken and not only feature selection methods but feature ranking methods generally are considered.

\textbf{Feature Ranking} is a broader term compared to Feature Selection, mapping onto more domains than just Feature Selection. Because in the construction of a Feature Ranking no assumptions are made on the desired data subset, a score is assigned to each of the dataset features, giving each dimension an `importance' score. Such scores can also be used to interpret and clarify a Machine Learning model: thus making such ranking processes useful to the interpretable AI domain.

\textbf{Feature projection} and Feature Selection are both processes relating to the concept of dimensionality reduction \citep{cunningham_dimension_2007}, however, there exists an important distinction between them. Whilst in the process of feature selection, relevant dimensions are sought and selected without altering their input values, in the process of feature projection (also called \textit{feature extraction}) data transformations are applied, mapping the original data onto a lower-dimensional space. Common methods of feature projection are \textit{\gls{pca}} for the supervised case and \textit{\gls{lda}} for the unsupervised case. Both \gls{pca} and \gls{lda} take a statistical approach to detecting feature interactions, which not always results in an optimal feature set for prediction. Rather, machine learning techniques can be used to select a more optimal subset. Although the two methods are different, the two aim at achieving the same goal and are thus encountered in similar contexts.

\subsection{Types of Feature Rankings}\label{section:methods-ranking-types}
To better understand in what form a Feature Ranking might be defined and how it relates to the process of Feature Selection, exact mathematical definitions of various types of feature rankings are given first. At all times, the prediction- or estimation of any quantity is denoted with a hat notation, i.e. `\textasciicircum', whilst the `true' value of the quantity has no such hat.

\subsubsection{Feature importance}\label{section:feature-importance-definition}
Feature importance scores are defined as a vector of $p$ dimensions, containing real-valued numbers. Let us define the vector like so:

\begin{equation}
\hat{\boldsymbol{w}} = (\hat{w}_1, \hat{w}_2, \ldots, \hat{w}_{p-1}, \hat{w}_p),
\end{equation}

where $\hat{\boldsymbol{w}} \in \mathbb{R}^p$. Such a ranking is obtained, for example, by running a feature ranker on the dataset and having it assigned as score to each dimension. The vector is assumed to be normalized, i.e. each vector element is divided by the vector sum:

\begin{equation}\label{eq:normalize-feature-ranking}
\hat{\boldsymbol{w}} = \frac{\hat{\boldsymbol{r}}}{\sum^p_{i=1} \hat{r}_i},
\end{equation}

given a scoring vector $\hat{\boldsymbol{r}}$, which indicates feature importance on an arbitrary scale. It is self-evident that $\hat{\boldsymbol{w}}$ has the property that $\sum^p_{i=1} \hat{w}_i = 1$, i.e. is a probability vector. An example such vector can be:

\begin{equation}\label{eq:importance-vector-example}
\hat{\boldsymbol{w}} = (0.20, 0.8, 0.0),
\end{equation}

in which it is clear that the ranking algorithm found the second feature to be the most important. In the case where multiple feature importance vectors are considered, e.g., in the case where $B$ bootstraps (Section~\ref{section:bootstrapping}) are considered, the vectors are stacked in a matrix, i.e.:

\begin{equation}\label{eq:feature-importance-matrix}
\mathbf{\hat{W}} \in \mathbb{R}^{B \times p},
\end{equation}

which denotes $B$ feature importance $p$-dimensional vectors arranged in a matrix of reals. The ground-truth $w$ will remain a vector, since there will still be only one such vector available per dataset.

\subsubsection{Feature support}\label{section:feature-support-definition}
Feature support indicates whether certain dimensions are chosen to be included a \textit{feature subset}; i.e. the vector marks elements as being chosen by the feature selection process. In this definition, the feature support vector is synonymous with the definition of a feature subset. Although some feature- ranking and selection processes approximate a suitable feature support vector directly, an algorithm can also make use of a threshold $\epsilon$ to generate a feature support vector from a feature importance vector. The feature support vector is a boolean-valued vector of $p$ dimensions.

\begin{equation}
\hat{\boldsymbol{s}} = (\hat{s}_1, \hat{s}_2, \ldots, \hat{s}_{p-1}, \hat{s}_p),
\end{equation}

where $\hat{\boldsymbol{s}} \in \mathbb{B}^p$. Note $\mathbb{B}$ is the boolean-valued vector space, i.e. its elements lie in the set $\{0, 1\}$. An example such vector can be:

\begin{equation}\label{eq:support-vector-example}
\hat{\boldsymbol{s}} = (1, 1, 0),
\end{equation}

which is the feature support mask obtained from thresholding the feature importance vector $\hat{\boldsymbol{w}}$ in the above example (Equation~\ref{eq:importance-vector-example}) using the threshold $\epsilon > 0.0$, causing one feature to be dropped from the feature subset. Just like for the feature importance vector, the feature support vectors can be arranged in a matrix in the boolean space, like so:

\begin{equation}\label{eq:feature-support-matrix}
\mathbf{\hat{S}} \in \mathbb{B}^{B \times p},
\end{equation}

given $B$ feature support vectors.

A \textbf{sparse representation} can also be constructed, given a feature support vector. A set is created containing only the indices of the selected features, causing a more sparse representation of the feature subset in case of high dimensionality and relatively few selected features. The support vector $\boldsymbol{s}$ and its prediction $\hat{\boldsymbol{s}}$ can be readily converted back- and forth into such a sparse representation. We define the sparse representation as the set $\hat{\mathbb{S}}$:

\begin{equation}
\hat{\mathbb{S}} = \{i \mid i \in \mathbb{Z} \wedge \hat{s}_i = 1 \},
\end{equation}

where $| \hat{\mathbb{S}} | = d$, i.e. $d$ dimensions were selected in the feature subset. Note, that the sparse feature subset is represented as a \textit{set} instead of a vector, meaning that the sequence is no longer considered ordered. To show a concrete example, the vector from Equation~\ref{eq:support-vector-example} is converted to the following set:

\begin{equation}
\hat{\mathbb{S}} = \{ 2, 1 \} \text{ where } | \hat{\mathbb{S}} | = 2,
\end{equation}

containing the indices of the selected features as defined in Eq~\ref{eq:support-vector-example}, in no particular ordering whatsoever. A sparse set $\hat{\mathbb{S}}$ can be converted back to a feature support vector $\hat{\boldsymbol{s}}$ like so:

\begin{equation}
\hat{s}_i = \begin{cases}
  1 & \text{if } i \in \hat{\mathbb{S}}\\
  0 & \text{otherwise}
\end{cases}
\end{equation}

Denote $B$ such sparse feature support sets as $\hat{\mathbb{S}}^{boot}$:

\begin{equation}\label{eq:feature-support-superset}
\hat{\mathbb{S}}^{boot} = \{ \hat{\mathbb{S}}_1, \hat{\mathbb{S}}_2, \ldots, \hat{\mathbb{S}}_{B-1}, \hat{\mathbb{S}}_B \},
\end{equation}

meaning $B$ sparse support sets were arranged in the superset $\hat{\mathbb{S}}^{boot}$.

\subsubsection{Feature rankings}\label{section:feature-rankings-definition}
Feature rankings are similar to feature importance scores, but with less precision. Whereas in a feature importance vector each element is approximated with a real-valued score, in an ordinary feature ranking the only considered facet is the \textbf{order} of the features in terms of importance. Although in most cases a feature importance vector is constructed first, after which a support or ranking vector can be constructed, in some cases only a ranking is available - e.g. in the case of \gls{rfe}. A feature ranking is constructed by assigning each dimension a rank, anywhere in the integer set $\{1, 2, \ldots, p - 1, p\}$. Such, the vector can be expressed as:

\begin{equation}
\hat{\boldsymbol{r}} = (r_1, r_2, \ldots, r_{p-1}, r_p),
\end{equation}

where $\hat{\boldsymbol{r}} \in \mathbb{Z}^p$, the integer-space. An example such vector can be:

\begin{equation}\hat{\boldsymbol{r}} = (2, 3, 1),\end{equation}

which is the feature ranking obtained from converting the feature importance vector $\hat{\boldsymbol{w}}$ in the above example (Eq~\ref{eq:importance-vector-example}) to a ranking. Such rankings are easily converted to importance vectors using Equation~\ref{eq:normalize-feature-ranking}, allowing one to use the same statistical machinery as for feature importance vectors. A higher rank number means a greater importance.

Such an integer-valued ranking can be utilised to select some $k$ best features to use in a subsequent learning task, i.e. to perform a feature subset selection. This is similar to selecting features using a feature importance vector. Now, however, a subset is constructed not based on a threshold value $\epsilon$, but by including a certain number of best features in the subset. Even though the feature- importance and ranking vectors both have the ability to select feature subsets, the feature importance vector carries more meaning, because it more precisely quantifies the relative importance of each feature. This extra meaning can be made to good use during the evaluation process of the feature rankings, especially when using synthetically generated datasets, in which the ground-truth feature relevance is available.

\subsection{Feature Selection Taxonomy}
To better understand what type of feature selectors exist, and how they relate to the types of feature rankings that can be constructed, a taxonomy is considered. Feature Selection can be constructed by running a separate statistical operation on the dataset, before running any learning algorithms, or as part of a learning algorithm itself. In some cases, a learning algorithm that is itself very sophisticated and time-consuming might still be worthwhile to use as a feature selection pre-processing step. This is because if time is saved by having the prediction estimator process less features, one might still have to spend less time in his learning process. In this way, one might enjoy gains in processing efficiency by using a feature selector.

Due to this distinction in the approach used in a feature ranking algorithm, a subdivision can be made to separate methods into more specific categories. As such, a common taxonomy in the field is created: subdividing feature ranking methods into the categories of filter-, wrapper- and embedded methods \citep{chandrashekar_survey_2014}.

\subsubsection{Filter methods}
Filter methods use some scoring mechanism to compute \lq usefulness' for each feature, without applying a learning algorithm. Having applied some ranking criterion, often a feature ranking is produced, after which some thresholding operation can be applied to select features. Although filter methods are often computationally light and do not overfit due to the absence of a learning algorithm, filter methods might miss out on more complex feature interactions, causing a non-optimal subset as a result. Also, choosing a suitable threshold to use can be difficult.

Examples of filter methods include the \textit{Fast Correlation Based Filter} \citep{yu_feature_2003}, which uses a quick and easy-to-compute statistical measure to select features according to some predefined threshold $\epsilon$ (in Yu et al. denoted as $\delta$). To illustrate the type of statistical quantities generally used in filter methods, the statistical quantity in Yu et al (2003) can be denoted like so:

\begin{equation}
S U(X, Y)=2\left[\frac{I G(X \mid Y)}{H(X)+H(Y)}\right],
\end{equation}

where $I G(X|Y)$ is the \textit{information gain} between two random variables $X$ and $Y$ and $H(X)$ is the \textit{entropy} of a variable: which are both metrics coming from the information-theoretical domain. The measure $SU(X, Y)$, then, is the \textit{symmetrical uncertainty} of two features $X$ and $Y$, where a feature $Y$ is more correlated to $X$ than to $Z$ if $IG(X, Y) > IG(Z, Y)$. In this way, a ranking can be constructed considering the measure $SU(X, Y)$, where feature are sought with high $SU$ scores. In a second phase of the algorithm, the scoring table is traversed yet again, to eliminate possible redundant features included in the selected feature subset.

\subsubsection{Wrapper methods}
Wrapper methods, on the other hand, use some learning algorithm to determine a suitable subset of features. A search is conducted over the space of possible feature subsets, eventually selecting the subset that has the highest validation score in the test set using a chosen learner as a predictor. Characteristics that define wrapper methods lend themselves similar characteristics to traditional optimisation problems; although an exhaustive search might yield an optimal solution, such a solution might not always be feasible due to its great time complexity. For this reason, in some applications a filter is applied first, before running a wrapper method.

Examples of such methods are numerous. Straight-forward methods include the range of \textit{sequential} feature selection methods, which aim to start (1) either with the full subset of dataset features or (2) with an empty set of features. The two approaches are called \textit{Forward} Feature Selection and \textit{Backward} Feature Elimination, respectively. To then facilitate a forward- or backward iteration step it is customary to use an estimator of some kind to retrieve a scoring on the features, selecting either the best- or eliminating the worst scored feature. \gls{rfe} is one such backward-elimination method, which might, for example, use a \gls{svm} to construct estimation scores \citep{maldonado_weber_2009}. Another option is to perform an exhaustive search, in which every feature combination is tried such to obtain the optimal feature subset, given the learning task and the estimator used. To obtain such a ranking, the chosen estimator might use an arbitrary method to compute it, be it a relatively simple learning step or a sophisticated model evaluation. Although such methods can perform reasonably well, such methods tend to be more time-consuming than filter methods.

\subsubsection{Embedded methods}
Embedded methods seek to combine the training task and feature selection. Given some suitable learner, features are weighted during the training process, producing either a feature ranking or a feature subset afterwards. e.g. some learners compute feature importance scores as part of their training process, which can then be used in combination with some threshold to select relevant features. Having already trained the model, subsequent prediction tasks can benefit from increased prediction speed by using less data.

Examples of embedded methods are \gls{lasso} and Ridge Regression, which perform feature selection alongside the process of finding optimal regression coefficients. For further explanations, see Section~\ref{section:ridge-regression}. Another notable embedded method is a \gls{dt}, which constructs, inherent during its fitting stage, a measure of importance on each variable. It does so by computing the probability of reaching a certain node - and determining the decrease in node impurity caused by weighting this probability value. To facilitate this computation, the probability of reaching a certain node is computed by considering the number of samples that reach the node during its decision-phase, divided by the total number of samples. Such, the leaves and depth obtained during the construction phase of the \gls{dt} can be used to determine a measure of feature importance, `embedded' into its learning phase.

\subsubsection{Hybrid methods}
A hybrid method is any method that is not classifiable by a single category, but rather lends from multiple categories. Hybrid methods can, for example, combine filter and wrapper methods \citep{hsu_hybrid_2011}, by first applying a computationally efficient filter and refining the result by using a wrapper method. Another paper \citep{das_filters_2001} describes its approach as hybrid due to both adding- and removing features in the feature selection process - exhibiting both forward- and backward selection at the same time. Lately, research was also put into examining \textit{Ensemble} feature selection methods \citep{bolon-canedo_ensembles_2019}, which combines the outputs of multiple selectors and decides useful features accordingly using some voting committee. Ensemble methods can be seen as hybrids or are seen as a category on its own.

\subsection{Types of features}
An important consideration in choosing a suitable feature selection method for any task is what kind of structure the concerning data has, if any at all. Data might exhibit tree, graph, or grouped structures, which is essential information when detecting feature interactions and determining useful features. Support for structured data is relatively new in the field and has not been an extensive point of concern for many feature selection algorithms in the past. Many traditional feature selection algorithms focused primarily `\textit{conventional flat}' features \citep{li_feature_2017}, in which the assumption is made that the data is independent and identically distributed (\textit{i.i.d.}). However, this assumption is widespread among many machine learning algorithms, since in many applications datasets are normalized to fit the i.i.d. condition before they are used.

Conventional data are opposed to more complex data structures, i.e., datasets with `structured features', as coined by the proposed taxonomy from Li et al (2017) \cite{li_feature_2017}, but also to linked data and \textit{streaming data}. In streaming data, the quality of an initially selected feature subset can be improved as more data comes in, and a feature selection algorithm is able to benefit from a larger distribution of samples. Adapting existing feature selection algorithms to fit the demands of streaming data proved to be a nontrivial problem. Nowadays, many companies and institutions have to deal with data volumes that easily exceed the boundaries of in-memory storage capacity - limiting the data scientist to train on only subsets of the entire datasets. Smart sampling is therefore needed to retain a representative sample distribution.

The scope of this research is limited to only the most common type of features, \textbf{conventional-, flat features} (\textit{i.i.d.}), or also \textit{tabular} data.

\subsection{Constructing Feature Rankings}
Although the number of existing feature ranking methods are numerous, a small set of example methods are explored to get a better understanding of the range of methods that do exist. First of all, a way of constructing a feature ranking that originates from classical statistics is examined. Next, a more algorithmic-type of approach is examined, which was specifically designed for the feature selection domain.

\subsubsection{Regularized Linear Regression}\label{section:ridge-regression}
One of the most fundamental methods in statistics is linear regression. It can be solved both analytically and numerically: in which the optimal approach is dependent on the amount of dataset dimensions at hand - where the amount of dataset dimensions $p$ gets very large, the analytic solution gets slower compared to an approximate method like Stochastic Gradient Descent. Recall that we can analytically solve Linear Regression by minimizing the Residual Sum-of-Squares cost function \citep{hastie_elements_2009}:

\begin{equation}\text{R}(\boldsymbol{\beta}) = (\mathbf{Y} - \mathbf{Z} \boldsymbol{\beta})^\intercal (\mathbf{Y} - \mathbf{Z} \boldsymbol{\beta}),\end{equation}

in which $\mathbf{Z}$ is our design matrix. Regression using this loss function is also referred to as `Ordinary Least Squares'. The mean of the cost function $\text{R}$ over all samples is called Mean Squared Error, or MSE. Our design matrix is built by appending each data row with a bias constant of 1 - an alternative would be to first center our data to get rid of the intercept entirely. To now minimize our cost function we differentiate $\text{R}$ with respect to $\boldsymbol{\beta}$, giving us the following unique minimum:

\begin{equation}\hat{\boldsymbol{\beta}} = (\mathbf{Z}^\intercal \mathbf{Z})^{-1} \mathbf{Z}^\intercal \mathbf{Y},\end{equation}

which results in the estimated least-squares coefficients given the training data, also called the normal equation. We can classify by simply multiplying our input data with the found coefficient matrix: $\hat{\mathbf{Y}} = \mathbf{Z} \hat{\boldsymbol{\beta}}$. Now, in the case where our model is fit using multiple explanatory variables, we are at risk of suffering from \textit{multicolinearity} - the situation where multiple explanatory variables are highly linearly related to each other causing non-optimal fitting of the model coefficients.

In \textbf{Ridge regression}, we aim to tamper the least squares tendency to get as `flexible' as possible to fit the data best it can. This might, however, cause parameters to get very large. We therefore like to add a penalty on the regression parameters $\boldsymbol{\beta}$; we penalise the loss function with a square of the parameter vector $\boldsymbol{\beta}$ scaled by new hyperparameter $\lambda$. This is called a \textit{shrinkage method}, or also: \textit{regularization}. This causes the squared loss function to become:

\begin{equation}\text{R}(\boldsymbol{\beta}) = (\mathbf{Y} - \mathbf{Z} \boldsymbol{\beta})^\intercal (\mathbf{Y} - \mathbf{Z} \boldsymbol{\beta})+\lambda \boldsymbol{\beta}^\intercal \boldsymbol{\beta},\end{equation}

where we can see that $\boldsymbol{\beta}^\intercal \boldsymbol{\beta}$ denotes the square of the parameter vector, thus supplementing the loss function with an extra penalty. This is called regularization with an $L^2$ norm; which generalization is called \textit{Tikhonov regularization}, which allows for the case where not every parameter scalar is regularized equally. If we were to now derive the solutions of $\boldsymbol{\beta}$ given this new cost function by differentiation w.r.t. $\boldsymbol{\beta}$:

\begin{equation}\hat{\boldsymbol{\beta}}^{\text {ridge }}=\left(\mathbf{Z}^{T} \mathbf{Z}+\lambda \mathbf{I}\right)^{-1} \mathbf{Z}^{T} \mathbf{Y},\end{equation}

in which $\lambda$ will be a scaling constant that controls the amount of regularization that is applied. Note $\mathbf{I}$ is the $p \times p$ identity matrix - in which $p$ are the amount of data dimensions used. An important intuition to be known about Ridge Regression, is that directions in the column space of $\mathbf{Z}$ with small variance will be shrunk the most; this behavior can be easily shown be deconstructing the least-squares fitted vector using a \gls{svd}. 

\textbf{\gls{lasso} regression} is a slightly modified variant of Ridge Regression, where instead of an $L^2$ norm an $L^1$ norm is used instead. This problem can be denoted as the following minimization problem:

\begin{equation}
\hat{\boldsymbol{\beta}}^{\text {lasso }}=\underset{\boldsymbol{\beta}}{\operatorname{argmin}}\left\{\frac{1}{2} \sum_{i=1}^{N}\left(y_{i}-\beta_{0}-\sum_{j=1}^{p} x_{i j} \beta_{j}\right)^{2}+\lambda \sum_{j=1}^{p}\left|\beta_{j}\right|\right\},
\end{equation}

in which the similarities between Ridge Regression are easily seen. The squared $\sum_1^p \beta^2_j$ ridge penalty is replaced by by the $L^1$ norm penalty of $\sum_1^p |\beta_j|$.

\textbf{Feature selection} can be employed using both \gls{lasso}- and Ridge regression. Because the dimensions whose coefficients are shrunk the most are presumably the least relevant to the prediction task, non-contributing features can be cut off from the design matrix using a threshold point. In fact, because \gls{lasso} does not square the weights vector but takes the absolute value, some coefficients might even be shrunk to near-zero values: removing the need for defining a threshold at all, since the coefficients have zero contribution already. In line with the feature subset definition given in Section~\ref{section:feature-support-definition}, such a feature support set can be constructed by thresholding the \gls{lasso} coefficients like so:

\begin{equation}
\hat{\mathbb{S}}^{lasso} = \{ j \mid \lvert \beta_j \rvert \geq \epsilon \},
\end{equation}

which will result in a feature support set containing only the indices of features where the coefficients were at least larger than $\epsilon$.

Although the former methods are suited for regression tasks only, regularization schemes are employed widely and have similar mechanics in these applications. Examples are numerous and include regularized Logistic Regression, regularized \glspl{svm} and regularized Neural Networks: employing regularization in any Machine Learning model is standard practice. In this way, we gained insight into a fundamental tool to estimate feature importance and shrinking the model coefficients using a regularization term - and more importantly, the fact that it can be employed for feature selection.

\subsubsection{Relief-Based Feature Selection algorithms}
The Relief-family of feature selection algorithms originates from a seminal paper by Kira et al. \citep{kira_feature_1992}, introducing the original version of the Relief algorithm. Over time, many variations on the Relief algorithm have been made, most notably ReliefF \citep{kononenko_estimating_1994}. In fact, so many variations on the algorithm were made that one can speak of \glspl{rba} in the literature \citep{urbanowicz_relief-based_2018}. Due to the overall architectural design of the algorithm, \glspl{rba} manage to rank features and optionally select a feature subset within a reasonable time-complexity domain and is often competitive when it comes to validation estimator performance.

The algorithm works by usage of an \textit{instance-based} learning method, considering one data sample at a time and updating statistics on feature importance. Given a training dataset $\mathbf{Z}$, $n$ samples, $p$ dimensions and a relevancy threshold $\epsilon$, Relief can compute a measure of feature importance in a finite amount of time. The essential concept is to compute the $p$-dimensional Euclidean distance for a randomly picked instance, iterating over all dataset samples and computing the \textit{nearest `miss'} and \textit{nearest `hit'} for every instance. Such a miss- or hit is defined to be either one of the positive- or negative dataset instances. To now compute how much `relevance' should be added for each feature given the closest hit- and miss, a subroutine is used, in which for every feature the \textsc{diff} between the randomly picked instance and nearest- hit and miss is computed and squared.

Once the iterations are complete, the summed weights for each feature are stored in a vector $\boldsymbol{w}^{relief}$. Once all $n$ samples have been traversed, the feature weights vector $\boldsymbol{w}^{relief}$ is normalized by dividing by the amount of samples $n$. In this way, a ranking is constructed - which can be converted into a feature subset by applying the threshold $\epsilon$ to the ranking. i.e., using the notation introduced in \ref{section:methods-ranking-types}, a feature subset can be constructed like so:

\begin{equation}
\hat{\mathbb{S}}^{relief} = \{ i \mid \boldsymbol{w}^{relief}_i \geq \epsilon \},
\end{equation}

given some threshold $\epsilon$.

\clearpage
\section{Evaluating Feature Rankings}\label{section:evaluation}
Like seen in Chapter~\ref{section:related-work}, the evaluation of feature- ranking and selection methods is a nontrivial process, that has been conducted in many ways in the literature. Because of this reason, it is desired to acquire a general evaluation method that is applicable to \textit{all} feature- ranking and selection methods. The algorithms should be evaluated comprehensively: highlighting all aspects relevant to the performance of the method. In this chapter, a reasoning is given on which evaluation metrics are sensible to use, and which evaluation methods are recommended to use for different scenarios.

\subsection{Cross-Validation and Bootstrapping}\label{section:cv}
Several steps can be undertaken to provide more reliable estimates of the feature ranker performance. The goal is to prevent the practitioner to get promising but misleading results, whilst keeping the complexity of the benchmarking process at a reasonable level.

\subsubsection{Cross-Validation}
For almost any Machine Learning experiment, it is recommended to use some form of \gls{cv}. Because a model has the possibility to get `learn' the training data, evaluating the performance of an estimator on just the set of training data is dangerous practice. Evaluation metrics can be misleading, suggesting better model performance than is actually the case. For example, an improperly cross-validated model that is employed outside the realm of the training data might turn out to have very poor generalization performance. Thus, also in the context of benchmarking feature ranking and selection algorithms, validation must be performed by holding out a set of the available data samples at hand.

The most simple form of \gls{cv} is a training/testing split, in which the practitioner holds out one set of the example data for use in the testing phase - it is to be remain unseen by the model during the training phase. More robust methods include training the model multiple times on various training datasets and evaluating on the held out testing data. This can be done using 5-fold or 10-fold \gls{cv}, where, for example, in the case of 5-fold \gls{cv} \sfrac{4}{5}th of the data is reserved for training and \sfrac{1}{5}th for testing, repeated for each fold - so 5 times.

What is especially important in the process of conducting \gls{cv}, is that no operations are performed \textbf{before} having split the data, that can influence the experimental results. For example, in the scenario where one want to select variables to be included in some feature subset and run some prediction estimators afterward, a practitioner must be careful to split the data before the variable selection process. This is because otherwise, the predictive estimators might gain an unfair advantage due to having already seen the left out samples, i.e. they have already gained an advantage from the variable selection that was performed on the full dataset. This might cause skewed and misleading estimations of the error rate of the estimators - causing faulty models. Therefore, it is at all times important to keep in mind the `right' way of performing \gls{cv}: split first before running any operations that regard the dataset samples \citep{ambroise_selection_2002}.

\subsubsection{Bootstrapping}\label{section:bootstrapping}
Bootstrapping, on the other hand, is a similar but different process. Performing a bootstrap is a classical method for estimating statistical quantities regarding the learning process, such as variance, prediction error, or bias. The process works by resampling the dataset with replacement $B$ times, such to create $B$ different permutations of the dataset. Then, the learning process is repeated for each of the $B$ bootstrap datasets, i.e., refitting the estimators for each of the permuted datasets. If, for example, the designated dataset is denoted like $\mathbf{Z}=\left(z_{1}, z_{2}, \ldots, z_{b}\right)$ with each sample $z_{i}=\left(x_{i}, y_{i}\right)$, then the $b$-th bootstrap permutation of the dataset can be denoted as $\mathbf{Z}^{* b}$. Such, estimates can be made of, for example, the variance of some statistical quantity $S$ computed over the dataset $\mathbf{Z}$:

\begin{equation}\label{eq:variance-bootstrap}
\widehat{\operatorname{Var}}[S(\mathbf{Z})]=\frac{1}{B-1} \sum_{b=1}^{B}\left(S\left(\mathbf{Z}^{* b}\right)-\bar{S}^{*}\right)^{2},
\end{equation}

which can be seen to be the unbiased average of the statistical quantity $S$ over the $B$ bootstrap permutations of $\mathbf{Z}$. Note that the average value of the statistic $S$ is computed like so:

\begin{equation}\label{eq:average-bootstrap}
\bar{S}^{*}=\sum_{b} S\left(\mathbf{Z}^{* b}\right) / B,
\end{equation}

i.e. by averaging $S$ over the $B$ bootstraps. Like such, a reasonable estimate can be made of any statistical quantity computed over the dataset, as long as the quantity can be computed for any permutation on the dataset and enough computational resources are available to run $B$ repeated experiments on each of the permutations.

The practice of bootstrapping comes useful to the evaluation of feature- ranking and selection algorithms, allowing the practitioner to better estimate quantities that would previously be less reliable estimates. Examples of such quantities are the feature ranking algorithm stability, variance, or fitting time. Especially in assessing the algorithm stability, there exists an interest to know how the designated algorithm functions under conditions of varying samples. For this, bootstrapping is especially useful, since the exact data generating distribution at hand is often not available- meaning no more samples can be drawn from the distribution to generate a larger data population. In the lack of a data generating distribution, resampling with replacement offers a solution.

\subsection{Validation estimators}\label{section:evaluation-validation-estimators}
A straight-forward way to evaluate the performance of a feature selection algorithm is to run the ranking algorithm, and subsequently run a `validation' estimator on the selected feature subset. The quality of the selected feature subset is then quantified through the performance of the validation estimator: when more informative and relevant features are selected, validation estimator performance presumably goes up. The validation estimator can be configured to be any classifier or regressor, dependent on the prediction task at hand, though often choices are made from a common set of estimators as can be seen in Table~\ref{table:evaluation-metrics-table}. 

This idea can be extended to feature rankings. Many feature selection algorithms are in fact feature ranking algorithms and do not provide a built-in mechanism to perform feature subset selection - the algorithms allow a user to define the desired amount of features to be selected using a hyper-parameter, which then uses its feature ranking internally to construct a feature subset, like explained in Section~\ref{section:methods-ranking-types}. It is therefore desired to also benchmark such algorithms in a systematic way: allowing more flexibility with respect to the evaluation process.

\begin{figure}[ht]
    \centering
    \includegraphics[width=\linewidth]{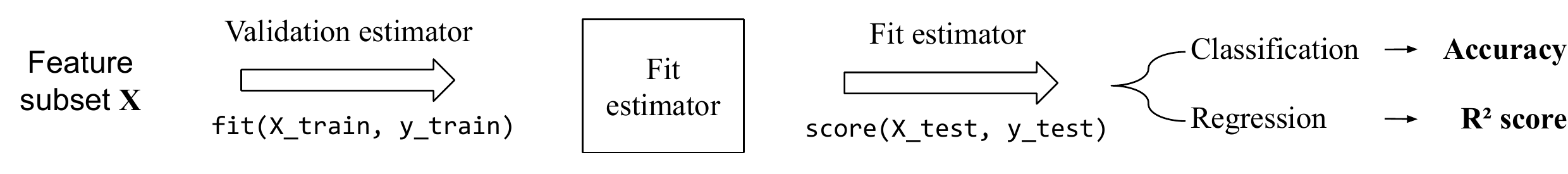}
    \caption{The process of running a validation estimator on a feature subset. In the case of a feature ranking, a validation estimator is fit for some various amount of feature subsets. Often, feature subsets with some number of the highest ranked features are evaluated.}
    \label{fig:schematic-validation-estimators}
\end{figure}

As can be seen in Figure~\ref{fig:schematic-validation-estimators}, the process of running a validation estimator is straight-forward. Once a feature subset has been determined by thresholding a feature ranking or using the feature subset computed by a feature selection algorithm, a validation estimator is trained and evaluated on the held-out test set. Afterwards, a suitable evaluation metric is used depending on the learning task at hand: classification or regression. Note - the \hyperref[section:introduction]{research scope} was confined to just classification and regression, not clustering. Exactly how many validation estimators are fit to evaluate a feature ranking depends on the ranking type (Section~\ref{section:methods-ranking-types}). In the case of a feature support vector, only the feature subset might have to be evaluated, whilst for feature importance or ranking vectors one might want to evaluate the first $k$ best feature subsets. Generally, each of the vectors are evaluated like so:

\begin{itemize}
    \item Feature importance (Section~\ref{section:feature-importance-definition}). Fit $k$ validation estimators, for the first $k$ best feature subsets, starting with a feature subset including only the highest ranked feature and subsequently including lower-ranked features. $k$ might be chosen depending on the dataset size at hand.
    \item Feature support (Section~\ref{section:feature-support-definition}). A feature subset was already selected by the feature selection algorithm and such the feature subset can be directly evaluated by the validation estimator.
    \item Feature ranking (Section~\ref{section:feature-rankings-definition}). A feature ranking can fit $k$ validation estimators; similarly to the feature importance vector.
\end{itemize}

Notably, in the case where a feature ranking algorithm supports computing both feature importance or ranking vectors and the feature support vector, simply the $k$ best feature subsets are evaluated by the validation estimator. It is thereby assumed that the selected feature subset is included in one of the $k$ best feature subsets. If for some reason the selected feature subset $\hat{\mathbb{S}}$ is not in any of the $k$ best feature subsets, the feature subset will be evaluated by the validation estimator separately.

A bootstrap can be used to generate more accurate estimations of the validation process. Especially for validation estimators that are sensible to varying permutations of the training data this process is useful - since otherwise validation metrics could be particularly misleading. In this way, given either an accuracy- or R\textsuperscript{2} score for classification and regression, respectively, an averaged bootstrapped score can be computed. This is done simply by plugging in either the R\textsuperscript{2} score or accuracy (or any other metric suitable for regression- or classification) into Equation~\ref{eq:average-bootstrap}.

\subsection{Apriori knowledge on relevant features}\label{section:evaluation-apriori-knowledge}
In the scenario where we know \textit{\gls{apriori}} which features are relevant, more direct evaluation can be applied on the constructed feature ranking. Such apriori ground-truth information can be gained in a couple of ways: (1) by utilising human domain expert knowledge, (2) by augmenting real-world datasets with noisy dimensions and assuming all other dimension to be uniformly relevant or lastly (3) by generating synthetic datasets with known feature importance levels (see Section~\ref{section:pipeline-components-datasets}. In the context of this section, the assumption is made that the feature importance scores are simply `known' apriori: no restriction is made on exactly how feature relevance ground truth was retrieved.

Knowing the relevant dataset features apriori can be useful information to the feature ranking evaluation process. This is because even though the goal of feature- ranking and selection algorithms is to separate the useful from the irrelevant features, the performance of such a ranking is nowadays most commonly evaluated by training yet another estimator; the `validation' estimator (Section~\ref{section:related-work} and Section~\ref{section:evaluation-validation-estimators}). This makes the evaluation scores dependent on the validation estimator, requiring publications to have run the same validation estimator to make the results comparable between papers. Also, for some datasets, the chosen validation estimator might be sophisticated enough to make up for any noisy dimensions that were added - reducing the differences between the feature rankings, which is instead desired to be amplified to determine which feature ranker is the best for which dataset. Lastly, a sophisticated validation estimator might also perform a form of feature selection on its own, which might also skew the performance of faulty feature rankings that included lots of noisy dimensions in its feature subset. For these reasons, it is interesting to investigate the possibilities to find robust and reliable evaluation methods that do not require a validation estimator.

Using the ground-truth relevant features, an accompanying evaluation metric can be constructed. Taking into consideration how different feature ranking algorithms construct various types of feature rankings as described in Section~\ref{section:methods-ranking-types}, a suitable metric can be created for each.

\textbf{Feature importance} (Section~\ref{section:feature-importance-definition}) scores are defined to be a real-valued $p$ dimensional vector, $\hat{\boldsymbol{w}}$. In order to evaluate the closeness of the vector $\hat{\boldsymbol{w}}$ to the dataset ground truth $\boldsymbol{w}$, a range of metrics can be used. A first approach would be to regard the estimated vector as an estimation of a continuous target vector, i.e., to regard the problem as a regression type of task. In such a perspective, each predicted feature importance would pose as a data sample. In this way, usual metrics related to the regression task can be used.

The \textbf{R\textsuperscript{2}-score} is one such metric that can be used in this context. This can be formalized like so:

\begin{equation}
\begin{aligned}
R^{2}&=1-\frac{\mathrm{RSS}}{\mathrm{TSS}} \\
\text{where}&\\
\mathrm{RSS} &= \sum_{i=1}^{n}\left(y_{i}-f\left(x_{i}\right)\right)^{2} =\sum_{i=1}^{p}\left(w_i - \hat{w_i} \right)^{2}, \text{ and}\\
\mathrm{TSS} &= \sum_{i=1}^{n}\left(y_{i}-\bar{y}\right)^{2} =\sum_{i=1}^{p}\left( w_i - \bar{\boldsymbol{w}} \right)^{2}, \\
\end{aligned}
\end{equation}

which can be used to evaluate the closeness of the predicted feature importance vector $\hat{\boldsymbol{w}}$ to the ground-truth feature importance vector $\boldsymbol{w}$.

The \textbf{logistic-loss}, or \textit{cross-entropy} score is another metric which can be employed to quantify the quality of a feature importance vector $\hat{\boldsymbol{w}}$. In this scenario, however, the target variables are regarded to be binary labels, which therefore means the predicted feature importance vector $\hat{\boldsymbol{w}}$ is to be compared against $\boldsymbol{s}$ instead of $\boldsymbol{w}$. Remember, from Section~\ref{section:feature-support-definition}, that $\boldsymbol{s}$ is the $p$-dimensional vector indicating with a boolean whether or not the feature is relevant, i.e. $\boldsymbol{s} \in \mathbb{B}^p$. Such, a zero indicates a feature is not relevant; and a one indicates the feature is informative or relevant. In this way, a feature ranker approximates the true binary relevance labels $\boldsymbol{s}$ with probability values arranged in the probability vector $\hat{\boldsymbol{w}}$. The logistic loss can be formalized like so:

\begin{equation}
\begin{aligned}
L_{\log }(y, p) &= -\log \operatorname{Pr}(y \mid p)=-(y \log (p)+(1-y) \log (1-p)) \\
&\text{substituting for } s_i \text{ and } \hat{w_i}\\
L_{\log }(s_i, \hat{w_i}) &= -\log \operatorname{Pr}(s_i \mid \hat{w_i})= -(\hat{w_i} \log (\hat{w_i})+(1-s_i) \log (1-\hat{w_i})) \\
&\text{averaged over all vector components }\\
L_{\log }(\boldsymbol{s}, \hat{\boldsymbol{w}}) &= - \frac{1}{p} \sum_{i=1}^p (\hat{w_i} \log (\hat{w_i})+(1-s_i) \log (1-\hat{w_i})) \\
\end{aligned}
\end{equation}

\textbf{Feature support} (Section~\ref{section:feature-support-definition}) is evaluated differently. Just like in the logistic-loss metric, the ground truth relevant labels are encoded as binary labels, i.e. the vector $\boldsymbol{s}$ is used as the ground-truth. Now, however, the predicted targets are also encoded as binary labels, i.e. the vector $\hat{\boldsymbol{s}}$ is used. This means that the problem is to be viewed in the supervised classification perspective and all metrics accompanying such task are available.

Classification \textbf{accuracy} is such a metric, measuring simply the average number of correct predictions were made between the predicted- and true target labels. It can be defined as such:

\begin{equation}
\begin{aligned}
\operatorname{accuracy}(y, \hat{y}) &= \frac{1}{n} \sum_{i=1}^{n} \boldsymbol{1} \left(\hat{y}_{i}=y_{i}\right) \\
&\text{substituting for } s_i \text{ and } \hat{s_i}\\
\operatorname{accuracy}(s, \hat{s}) &= \frac{1}{p} \sum_{i=1}^{p} \boldsymbol{1} \left(\hat{s}_{i}=s_{i}\right), \\
\end{aligned}
\end{equation}

where $\boldsymbol{1}(x)$ is the indicator function \citep{davis_undecidable_2004}. Using this metric, the amount of useful features included in a feature subset is rewarded with higher accuracy scores accordingly.

\textbf{Feature rankings} (Section~\ref{section:feature-rankings-definition}) can be evaluated similarly to feature importance scores - thereby also requiring normalization. Presume that besides the relevance of the features is known as a binary value, also the \textit{order} of relevance is known, i.e. which features are more relevant than others. This is very similar to the feature importance scores: though the difference is that the feature importance scores must first be converted to a ranking such to allow for meaningful comparison. In this reasoning, both the predicted- and the ground truth feature ranking vectors, which are $\hat{\boldsymbol{r}}$ and $\boldsymbol{r}$ respectively, can be normalized using Equation~\ref{eq:normalize-feature-ranking} to obtain $\hat{\boldsymbol{w}}$ and $\boldsymbol{w}$, respectively.

In this way, the normalized vectors $\hat{\boldsymbol{w}}$ and $\boldsymbol{w}$ can again be considered probability vectors, such that metrics like the R\textsuperscript{2}-score can be used: just like for the feature importance scores.

One has to keep in mind, however, not to intermix the feature importance and feature ranking scores with each other: even though the same metric is used to convert to summarize the predicted feature- importance and ranking vectors into a single scalar, the scorings are built from different vectors to begin with. The R\textsuperscript{2}-score coming from the feature importance vectors might have an unfair advantage due to the fact that they have floating-point precision on their approximations, whilst the feature ranking vectors $\boldsymbol{r}$ are converted from the integer domain to be normalized into floating-point numbers.

\subsection{Stability}
The stability of any algorithm is an important facet of the total method performance - which must not be overlooked or forgotten. This is for in many applications, only robust algorithms can be systematically relied on. Although it is only natural for algorithms to vary in fitting behavior under different permutations of the sample set, it is at all times desired to get an algorithm that is as stable as possible. To evaluate the stability of feature ranking algorithms, an easy method is to take several bootstrap permutations of the dataset (Section~\ref{section:bootstrapping}), to simulate drawing new samples from the data generating distribution at hand. Assuming such a bootstrapping procedure to be in place, several metrics can be used to quantify stability.

\subsubsection{Stability of feature importance vectors}\label{section:feature-importance-stability}
Feature importance scores are defined to be $p$ dimensional vectors in the domain of real numbers $\mathbb{R}$ (Section~\ref{section:feature-importance-definition}). Accordingly, the matrix containing $B$ such vectors is the $B \times p$ dimensional matrix in $\mathbb{R}$, denoted as $\hat{\mathbf{W}}$. A straight-forward way to assess the stability of the feature importance matrix is to compute the variance for each of the $p$ dimensions over $B$ bootstraps, i.e. Eq~\ref{eq:variance-bootstrap} is used by plugging in a column of the feature importance matrix $\hat{\mathbb{W}}$ as the statistical quantity $S$:

\begin{equation}
\widehat{\operatorname{Var}}[\hat{\mathbf{W}}_{:,i}]=\frac{1}{B-1} \sum_{b=1}^{B}\left(\hat{W}_{b,i}-\bar{\hat{\mathbf{W}}}_{:,i}^{*}\right)^{2},
\end{equation}

where $\bar{\hat{\mathbf{W}}}_{:,i}^{*}$ is the average feature importance score for the $i$-th feature over $B$ bootstraps. In this way, the variance over each dimension can be computed for a single feature ranker.

To summarize the variances over all dimensions, their summation might be taken, i.e., the variances are summarized as the scalar $\sum_{i=1}^p \widehat{\operatorname{Var}}[\hat{\mathbf{W}}_{:,i}]$. Alternatively, one might want to weight such summation, in order to reflect the desire to penalize instabilities in the ranking of relevant features more so than instabilities in the ranking of irrelevant features. In the case where the ground-truth feature importance scores are given, i.e. $\boldsymbol{w}$ is known, this vector might be used to apply a weighting scheme to the summation of the variances. One can do so by taking the inverse normalized ground truth feature importance score, i.e. $\frac{1}{w_i}$ for the $i$-th feature. Such, the summarization of $\hat{\mathbf{W}}$ now becomes $\sum_{i=1}^p \frac{1}{w_i} \widehat{\operatorname{Var}}[\hat{\mathbf{W}}_{:,i}]$.

\subsubsection{Stability of feature support vectors}
This time around, a way to quantify the stability of a feature subset is desired, i.e., the feature support vector $\hat{\boldsymbol{s}}$. A measure for the quantifying stability of set permutations that comes easily to mind might be the Hamming distance between multiple pairs of algorithm runs, i.e. ran on the same dataset with varied sample populations. Indeed, in \citep{dunne_solutions_2002} a measure based on Hamming distance is proposed. Reports also exist on numerous other approaches, like an entropy based measure \citep{krizek_improving_2007}, a measure based on the cardinality of intersection measure \citep{kuncheva_stability_2007} and a measure based on correlation coefficients \citep{kalousis_stability_2007}.

\textbf{Desired properties} of any measure are important to define in concrete manner. To compare the strength of any of these measures, the general objective for exactly what information we want to convey in a stability metric should be verbalized first. In \citep{mohana_chelvan_survey_2016}, three desired stability metric properties are expressed: (1) \textit{Monotonicity}, (2) \textit{Limits} and (3) \textit{Correction for chance}. Given the proposed measures in the literature, only few measures satisfied all properties. In another paper \citep{nogueira_stability_2018}, however, the authors extend the set of desired properties to include another two: (4) the stability estimator must be \textit{Fully defined} and (5) \textit{Maximum Stability $\leftrightarrow$ Deterministic Selection} - meaning that a maximum stability value should be achieved if-and-only if all feature sets are exactly identical.

\textbf{A measure} for quantifying feature subset stability was proposed in \citep{nogueira_stability_2018}, using the set of newly proposed desired properties. Having explored a statistically sound method to define a metric satisfying all five properties, the authors proposed a novel stability estimator, adjusted to this paper's terminology:

\begin{equation}\label{eq:stability-measure}
\hat{\Phi}(\hat{\mathbb{S}}^{boot})=1-\frac{\frac{1}{p} \sum_{i=1}^{p} \sigma_{i}^{2}}{\mathbb{E}\left[\frac{1}{p} \sum_{i=1}^{p} \sigma_{i}^{2} | H_{0}\right]}=1-\frac{\frac{1}{p} \sum_{i=1}^{p} \sigma_{i}^{2}}{\frac{\bar{k}}{p}\left(1-\frac{\bar{k}}{p}\right)},
\end{equation}

where $\hat{\Phi}$ resembles the stability estimate of the feature subsets arranged in $\hat{\mathbb{S}}^{boot}$ (Eq~\ref{eq:feature-support-superset}). Remember, that each set in $\hat{\mathbb{S}}^{boot}$ is a feature subset containing the indices of the selected features. Each of the sets in $\hat{\mathbb{S}}^{boot}$ is created by running the feature ranker on a permutation of the datasets, i.e. a resampling of a dataset equal probability distribution and dataset properties. Furthermore, the authors define $\sigma_{i}^{2}$ as the unbiased sample variance of the selection of the $i^{t h}$ feature and $\bar{k}$ as the average number of features selected over the $B$ feature sets. Given this paper's goal, there is no necessity for going into further mathematical details - for this we refer to the paper itself.

What is important is that this new measure satisfies all desirable properties for quantifying stability, as was proven in the paper. Accompanying the novel definition are instructions for computing confidence intervals and for performing a hypothesis testing for comparing various method stabilities. Exploring feature selection stability values given a set of parameter choices not only allows for choosing better hyperparameters, but also allows for comparing stabilities over various feature selection methods.

\subsubsection{Stability of feature ranking vectors}
Lastly, the stability of feature ranking vectors is also to be quantified. Since the stability measure for feature support was all about working with sets instead of vectors, the best option is to again normalize to a feature importance vector and compute the variance thereof. The summarization of the variances might again be weighted using the ground-truth feature importance vector $\boldsymbol{w}$, if it is available. See Section~\ref{section:feature-importance-stability}.

\subsection{Time complexity}
Another metric to be taken into account is algorithm complexity, which manifests itself in three interlinked aspects: time-, storage- and algorithm simplicity. Like the classical principle \textit{Occam's razor} implies - there at all times exists a preference for simpler models over more complex ones, especially in the case both accomplish the same feat. So, an understandable model is preferred that performs limited computational steps in order to restrain time- and storage complexity from rising too high.

Although measuring algorithm time- and storage requirements does provide some understanding, more insightful would be a theoretical description of the complexity in Big O notation. Theoretical complexities are harder to obtain though - leaving many authors to resort solely to measurements. Nonetheless, complexity analysis is recommended to be part of any feature selection method comparison, for it is a critical aspect to consider.

\subsection{Statistical integrity}
Given that the above discussed evaluation methods are computed, a statistical test is ought to be applied to provide convincing evidence for any algorithm's superiority. Like explained in \citep{demsar_statistical_2006}, many papers make implicit hypotheses acclaiming improved performance over existing methods. Although the chosen metrics might have been appropriate to statistically show performance gains as significant, the results are less reliable when left not validated by a statistical test. Therefore, a statistical verification step is required in the feature selection evaluation process.

Comparing multiple feature selectors is a non-trivial problem, which can be seen as the problem of comparing multiple \textit{classifiers}. To compare multiple classifiers, \citep{demsar_statistical_2006} recommends the Wilcoxon signed ranks test for comparison of two classifiers and the Friedman test for comparison of more classifiers given multiple datasets. Accompanying the Friedman test, it is recommended to perform corresponding post-hoc tests, such as the Nemenyi test \citep{nemenyi_distribution-free_1963}. The usually popular ANOVA test was advised against because, given the context of machine learning, ANOVA assumptions are violated, e.g. ANOVA assumes samples are drawn from normal distributions and the requirement of random variables having equal variance.

A \textbf{Wilcoxon signed ranks test} and the \textbf{Friedman test}, are for these reasons recommended for two- or more methods respectively, to statistically verify significant differences in performance of feature selection algorithms over the others using some summarizing scalar per feature selection method per dataset. The Friedman test is recommended to be accompanied with a post-hoc test, like Nemeyi's post-hoc test.

\clearpage
\section{Building a benchmarking pipeline}\label{section:pipeline}
To facilitate conducting a comprehensive experiment, an accompanying software pipeline is to be built. Such a pipeline must be able to run feature ranking methods, select feature subsets, and validate the quality of the feature subsets using some validation estimator. After both the feature ranking- and validation step, evaluation metrics have to be computed - assessing the performance of the feature ranking both directly- and indirectly. After the metrics have been computed, results should be stored either on-disk or be uploaded to the cloud, after which they should be made interpretable by visualizing them. Furthermore, because the `direct' evaluation of feature rankings is only possible when a ground-truth reference of the relevant features exists (Section~\ref{section:evaluation-apriori-knowledge}), it is also desired to be able to generate synthetic datasets built in to the framework. In this way, a practitioner using the pipeline can easily design synthetic datasets suited for the purpose at hand.

\textbf{Challenges} lie mainly in the scalability of the system. To conduct an experiment of moderate size, a single machine can be used for all stages of the pipeline: run the algorithms, collect and visualize the data. In such a small-scale context, the data- collection and visualization stage could as well be manual- no huge amounts of data have to be processed, so such a collection- and visualization stage could be run from a single script. When a larger experiment is desired to be run, however, a more sophisticated system is desired. Practitioners must not be restricted in running the pipeline on a single CPU or single machine, i.e., the system should support both \textit{horizontal}- and \textit{vertical} scaling. To design a system that can run experiments in such \textbf{scalable} and distributed ways, the system must be modular to facilitate distributing jobs over multiple machines, although it is also necessary to aggregate results of individual jobs as part of a pipeline step itself - requiring a piece of code to be aware of every job having finished.

An \textbf{implementation} of such a pipeline was made in Python, for the purposes of this paper. Many of the challenges described above were addressed and overcome by using an architecture that was built from the ground up to be scalable. Both the data- collection and aggregation processes were automated - after having run the pipeline for a collection of feature ranking methods and datasets, results are automatically stored, aggregated, and visualized. In this chapter, the reader is explained the general architecture of the system and the components that make the system work - after which the usage of the pipeline is shown along with its capabilities for scaling. The chapter is meant to be explanatory at first and provide concrete instructions and examples afterward.

\subsection{Architecture}
To first get an idea of the structure of the system, comments are made on its architecture. The pipeline has two main components: an encapsulating pipeline `run' script and the pipeline itself. The pipeline run script is to be called the \textit{main} script: allowing us to differentiate easily between the two. Whilst the main script is designed to be generic and work for any \gls{ml} related benchmarking task, the pipeline itself is to be tailor made for the specific task at hand. Ideally, one could swap out the pipeline for another (custom built) one, but still using the main script mechanics: allowing a practitioner to create various but related tasks using a single framework.

\subsubsection{The main script}\label{section:pipeline-main-script}
\begin{figure}[ht]
    \centering
    \includegraphics[width=\linewidth]{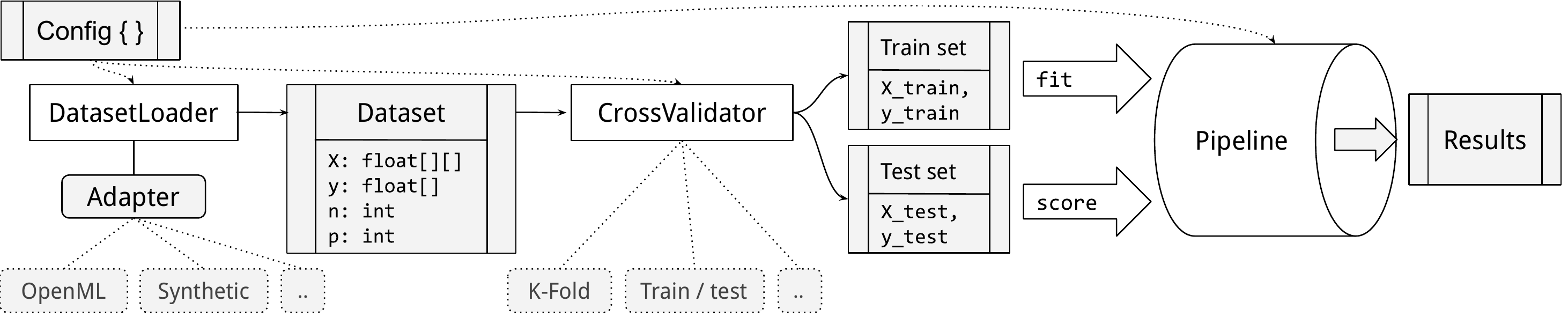}
    \caption{A general architecture schematic of the built benchmarking pipeline's \textit{main} script: providing data and instructions to the pipeline. Once the designated dataset is loaded using a configured adapter, the data is fed to the pipeline after a \gls{cv} split. The pipeline is first fit using the training data and then scored using the testing data, after which the results are collected and stored.}
    \label{fig:schematic-main-architecture}
\end{figure}

The sole purpose of the main script is to do three things: (1) load a dataset, (2) split the dataset using a \gls{cv} method and lastly (3) run the pipeline. In this procedure, any dataset is loaded into the system using an \textit{adapter} - allowing data to come in from any source by use of a plug-able architecture (Section~\ref{section:pipeline-components-datasets}). Immediately after the dataset has been loaded a \gls{cv} split is conducted. In this way, because the cross-validation step is separated from all the pipeline steps itself, it is less likely that a pipeline can go faulty on the \gls{cv} process - since the data splitting operation was already conducted once a pipeline implementer gets their hands on the data. An illustration of the main script, with as part of it the pipeline, can be seen in Figure~\ref{fig:schematic-main-architecture}.

An important facet facilitating the operations in the main script is the \textbf{config} object (as seen at the top-left in Figure~\ref{fig:schematic-main-architecture}). The config object contains all configuration and parameters relevant to all pipeline steps, essentially storing all instructions required to perform a single pipeline run. It is thereby important to store \textbf{all} configuration and not to leave any parameters set at random or use temporary variables - the pipeline execution is desired to be \textit{deterministic} and therefore making its results \textit{reproducible}. In this way, experiments conducted using the pipeline are better suited for scientific experiments - a situation in which a user desires to be able to store exactly the state that produces a certain result and explain its findings in minute detail.

To facilitate configuring the system in such a way, several options exist in the Python ecosystem. One could use the built-in \texttt{optparse} module, or its Python 3 counterpart \texttt{argparse}. The built-in packages leave much room for improvement, however. The modules do not allow loading config from a file, do not allow nesting, has only basic input validation and provides no tools for more complex configuration scenarios like the pipeline at hand. A solution for this is the \textit{Hydra} library \citep{yadan_hydra_2019}, a framework providing powerful utilities for configuring a Python application. The package allows one to create hierarchical and composable configurations, to be set using either the command-line or using yaml files. A simple yaml definition configuring the main script to use the `Iris Flowers' dataset and perform a K-fold \gls{cv} can be seen in Listing~\ref{code:pipeline-main-config}.

\begin{lstlisting}[caption={A simple configuration for the main script, using the `Iris Flowers' dataset and K-fold \gls{cv}.}, label={code:pipeline-main-config}]
defaults:
  - base_config
  - base_rank_and_validate
  - dataset: iris
  - cv: kfold
  - callbacks:
    - wandb
  - storage_provider: wandb
\end{lstlisting}

It can also be observed, in Listing~\ref{code:pipeline-main-config}, that there exist configuration possibilities for `callbacks' and a `storage provider'. These allow customizing a back-end for sending the result data to and an adapter for storing- or caching pipeline steps, respectively. Callbacks and the storage provider will be elaborated upon in Section~\ref{section:pipeline-callbacks} and Section~\ref{section:pipeline-storage-provider}, respectively. First, a look is taken at the pipeline implementation for feature ranking and subset validation, which is called \textit{Rank and Validate}.

\subsubsection{The `Rank and Validate' pipeline}\label{section:pipeline-rank-and-validate}
With the main script now defined, a better look can be taken at the pipeline implementation itself. Whereas the main script was still a general-purpose \gls{ml} benchmark runner, an implementation of the system by means of a pipeline is task-specific. i.e., one such pipeline is meant to execute a specific \gls{ml} task, such as performing evaluation on a feature ranker. The pipeline does so in a couple steps:

\begin{enumerate}
    \item \textbf{Resample} the dataset. Using the configured settings, the dataset is resampled. Options that are ought to be supported are shuffling, and more importantly; bootstrapping. It is thereby important to fixate a random `seed', which allows one to reproduce the resampling exactly.
    \item \textbf{Rank} features. Next, the feature ranker is set to work. The resampled data is passed to the ranker's \texttt{fit} function, after which the stored ranker might be cached.
    \item \textbf{Validate} feature subset. Finally, after the ranker has been fit, its ranking is used to validate its feature subsets, using some validation estimator. Again, the fit validation estimator might be cached. The final results are stored to disk and/or uploaded to the cloud.
\end{enumerate}

\begin{figure}[ht]
    \centering
    \includegraphics[width=\linewidth]{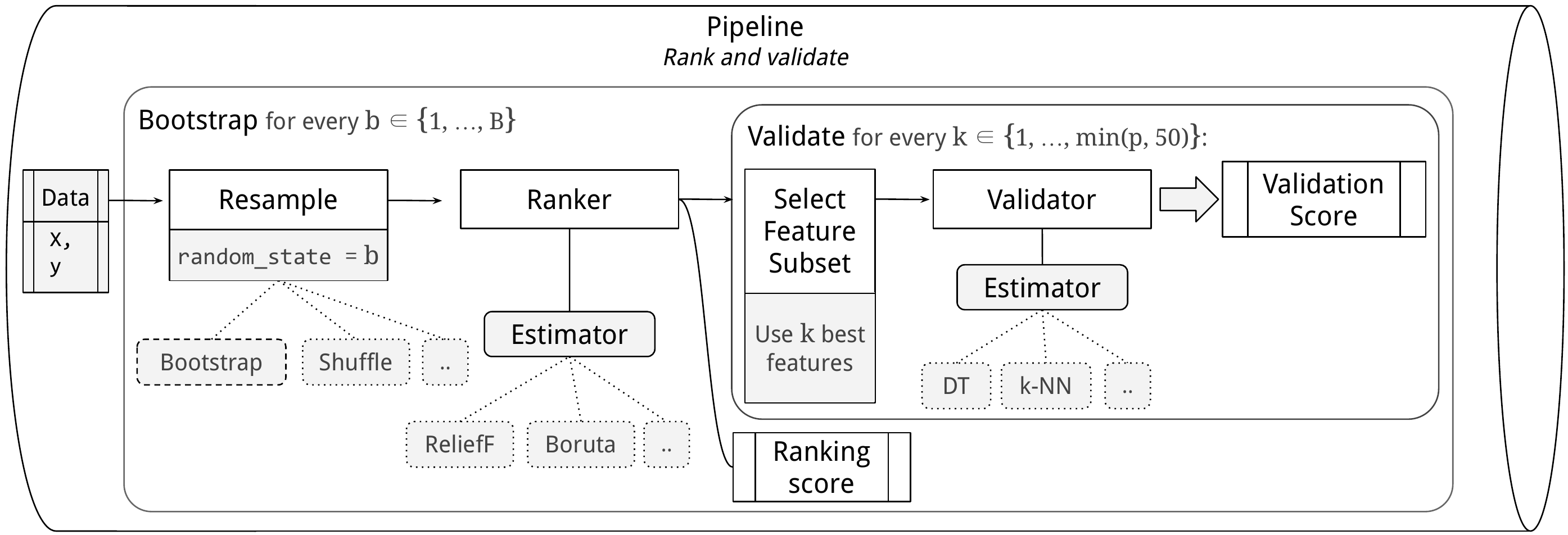}
    \caption{The `Rank and validate' pipeline. The pipeline performs feature ranking, subset selection and subset validation, in a sequence of loops, such to facilitate bootstrapping. The pipeline is first fit and then scored.}
    \label{fig:schematic-pipeline-architecture}
\end{figure}

This process is illustrated in Figure~\ref{fig:schematic-pipeline-architecture}.

As can be seen illustrated by rectangles in Figure~\ref{fig:schematic-pipeline-architecture}, the above steps are executed in a specific sequence of loops. First of all, all steps (1-3) are passed through \textbf{twice}: once for the training data using the \texttt{fit} function and once using the testing data using the \texttt{score} function. The reasoning behind the division into these two steps is to force keeping application logic separate, i.e., to enforce a separation of concerns. Because the fitting step is essentially different than the scoring step, it makes sense to separate them. This separation also allows better implementations of caching, because a practitioner is enforced to store all estimator related state inside the instance itself. This distinction is on par with the sci-kit learn \citep{pedregosa_scikit-learn_2011} API.

For each of the two passes, the entire process is run $B$ separate times: once for each \textbf{bootstrap}. The random resampling seed is changed with each bootstrap iteration accordingly: making sure each bootstrap dataset is a different random resampling of the dataset, but yet keep the pipeline to be reproducible. The results from each bootstrap are then stored with each bootstrap random state attached, allowing a practitioner to differentiate between results and possibly aggregate- and group them.

Finally, each of the bootstrap iterations performs a number of feature subset validations. Ideally, in the case where a dataset has $p$ dimensions and a ranker has computed a feature importance score for each of them, it would be desired to validate $p$ feature subsets - the first subset starting with only the best ranked feature and subsequently including lesser-ranked features. However, this would fast become computationally intractable: in the case where $p$ is of considerable size, the amount of feature subsets to validate becomes vast, fast. A compromise between the thoroughness and computational tractability is to only validate some number of the best feature subsets up to a certain dimension. The choice that is made in this paper is to validate at most 50 feature subsets and possibly less if $p < 50$. i.e., the $\min (p, 50)$ best feature subsets are evaluated. This means for every $k \in \{1,\ldots,\min (p, 50) \}$ the $k$ best features are used for subset validation.

The \textbf{scoring} process happens only after all estimators in the pipeline have been fit. Once the fitting phase is complete, the testing sets will be passed to the pipeline's \texttt{score} function. Each estimator gets the opportunity to generate some score given the testing dataset. Because both the rankers \textit{and} estimators are considered estimators in the system, a ranker will also get such scoring opportunity. More specifically so, rankers and validators are scored in the following ways:

\begin{itemize}
    \item \textbf{Rankers} are scored using the dataset ground-truth relevant features, if available. Thereby metrics as described in Section~\ref{section:evaluation-apriori-knowledge} are used: the \textbf{R\textsuperscript{2} score} and the \textbf{logistic loss}.
    \item \textbf{Validators} are scored like described in Section~\ref{section:evaluation-validation-estimators}. i.e., the scoring depends on the dataset task: the \textbf{R\textsuperscript{2} score} is used in the case of \textit{regression}, and the \textbf{accuracy} score is used in case of \textit{classification}.
\end{itemize}

That said, a general picture of the pipeline architecture and data flow is now obtained. The pipeline is, however, more complex. The pipeline supports multiprocessing, distributed computing, caching, and data visualization. Moreover, each pipeline component has very different functionalities and can be configured separately from the rest of the pipeline. Therefore, a look is taken into the separate pipeline components first in Section~\ref{section:pipeline-components}, to then take an in-depth view of the pipeline execution afterwards, in Section~\ref{section:pipeline-execution}. Finally, an elaboration is made on data visualization in Section~\ref{section:pipeline-visualization}.

\subsection{Components}\label{section:pipeline-components}
Each pipeline component can be configured using their own yaml files and has a reasonable level of isolation with respect to its module structure. The main components are discussed briefly.

\subsubsection{Datasets}\label{section:pipeline-components-datasets}
The system allows using datasets from numerous sources, using \textit{adapters}. Adapters are designed to be generic interfaces for fetching data from any source, as long as it can be expressed in two matrices $\mathbf{X}$ and $\mathbf{y}$. Although some adapters are available built-in to the framework, new adapters can easily be configured. Any such adapter must only implement a \texttt{get\_data} function, which returns the two matrices $\mathbf{X}$ and $\mathbf{y}$, given a configuration object facilitated by Hydra (see Section~\ref{section:pipeline-main-script}). Such adapters can implement any logic that retrieves the data themselves, be it logic for loading a dataset from local disk or from an internet platform.

\textbf{OpenML} is one such platform. OpenML \citep{vanschoren_openml_2014} is a platform for sharing and organizing data, completely open to all. Due to the flexible adapter structure of the system, and thanks to the availability of a Python package providing interfacing support with the platform, an integration with OpenML was made possible. Using the interface, datasets are fetched from a remote and cached to disk once downloaded. In this way, a cached version of the dataset is used in subsequent requests - making sure a dataset is not unnecessarily downloaded again. Integration with the platform allows users of the pipeline to access a large library of datasets, varying in type and domain. The dataset library features both regression and classification datasets, and of various target types, i.e., both univariate and multivariate. Also, the platform has multiple existing benchmark suites that can be used, such as the OpenML-CC18 \citep{bischl_openml_2019}. An example definition of a configuration file for loading an OpenML dataset is to be found in Listing~\ref{code:pipeline-openml-example}.

\begin{lstlisting}[caption={A dataset config for loading the `Iris Flowers' dataset from OpenML.}, label={code:pipeline-openml-example}]
name: Iris Flowers
task: classification
adapter:
  _target_: fseval.adapters.OpenML
  dataset_id: 61
  target_column: class
\end{lstlisting}

It can be seen, in Listing~\ref{code:pipeline-openml-example}, that a `target' class can be defined. This is passed to Hydra, which in turn instantiates the class with the given configuration. This is one of the core features that makes possible the modular adapter architecture that is in place.

\textbf{Synthetic} datasets can also be generated using the modular adapter-structure in place. By viewing a synthetic data generator as yet another adapter, providing data given a configuration object, both real-world and synthetic datasets can be used and generated using a single interface. A straight-forward way to generate synthetic datasets is to use the sci-kit learn \texttt{datasets} module, which provides functions for drawing samples from a wide variety of distributions. Aside from generic generators, there also exists support for drawing from more specific distributions, such as `two interleaving half circles', an `S curve' or a `swiss roll' distribution.

Most interesting for the current purposes, however, are two functions for generating classification- and regression datasets, \texttt{make\_classification} and \texttt{make\_regression}, respectively. Besides allowing one to configure the amount of samples and dimensions to generate, one can also define the desired informative-, redundant-, repeated- and irrelevant features to generate. In this way, a ground-truth for the relevant features can be constructed. In the case of regression, one can also retrieve the coefficients, or weighting, for each feature which are to be approximated, i.e., the desired feature importance for each feature. This means that whilst in the case of classification one has access to a \textit{binary} ground-truth, an \textit{exact} measure of the desired feature importance can be retrieved in the case of regression. An example definition of a synthetic classification dataset can be found in Listing~\ref{code:pipeline-synthetic-example}.

\begin{lstlisting}[caption={A dataset config generating a synthetic dataset using the sci-kit learn \texttt{make\_classification} function.}, label={code:pipeline-synthetic-example}]
name: Synclf hard
task: classification
adapter:
  _target_: sklearn.datasets.make_classification
  class_sep: 0.8
  n_classes: 3
  n_clusters_per_class: 3
  n_features: 50
  n_informative: 4
  n_redundant: 0
  n_repeated: 0
  n_samples: 10000
  random_state: 0
  shuffle: false
feature_importances:
  X[:, 0:4]: 1.0
\end{lstlisting}

Worth noting in Listing~\ref{code:pipeline-synthetic-example} is the presence of a `feature\_importances' attribute. This is the exact definition of the ground-truth relevant features, used by the pipeline. The system allows one to define the feature importances as numpy-indexed selectors, operating on some matrix the same size as $\mathbf{X}$. Every instance and every accompanying dimension can have a specific ground-truth weighting, although for the current purposes only \textit{global} feature- ranking and selection are considered. In the example above, it can be seen that the first 5 features are defined to be equally relevant: and therefore all have a uniform weighting of 1.0.

\subsubsection{Cross-Validation}
The Cross-Validator can be configured to be any class that has a \texttt{split} function, taking in the matrices $\mathbf{X}$ and $\mathbf{y}$ and returning a generator. The generator must then be an iterator over the number of configured folds, with each split returning the designated training- and testing data. An example configuration for a 5-Fold \gls{cv} split can be seen in Listing~\ref{code:pipeline-cv-example}.

\begin{lstlisting}[caption={A config for 5-Fold \gls{cv}, with shuffling. The split is reproducible due to the fixed random seed.}, label={code:pipeline-cv-example}]
name: K-Fold
splitter:
  _target_: sklearn.model_selection.KFold
  n_splits: 5
  shuffle: True
  random_state: 0
  fold: 0
\end{lstlisting}

As is seen in Listing~\ref{code:pipeline-cv-example}, a \gls{cv} technique can be configured to be a sci-kit learn module. Another important thing to note is the fixed `fold' attribute, set in this case to 0. Because the pipeline always executes exactly one fold, performing a K-Fold \gls{cv} is done by executing the pipeline multiple times using different values of \texttt{fold}.

\subsubsection{Resampling}
Built in, there are two options for resampling. There exist options for (1) bootstrap sampling and (2) shuffle resampling. Whilst the latter only changes the permutation of the dataset, the former performs resampling \textit{with} replacement, therefore actually changing the dataset distribution. \textbf{Bootstrap} resampling is configured like as can be seen in Listing~\ref{code:pipeline-bootstrap-example}.

\begin{lstlisting}[caption={A config for bootstrap sampling using a sample size of 1.0, i.e, using the full dataset.}, label={code:pipeline-bootstrap-example}]
name: Bootstrap
replace: true
sample_size: 1.00
\end{lstlisting}

Noteworthy, in Listing~\ref{code:pipeline-bootstrap-example}, is the `sample size' parameter. If needed, the dataset can be down- or up-sampled. Although up-sampling generally does not make much sense, down-sampling can increase the stochasticity of the bootstrapped dataset permutation. This means that, if desired, one can create more `random' distributions by setting \texttt{sample\_size} to a lower number. Lastly, a random seed can be fixed, like as can be seen in the pipeline schematic, Figure~\ref{fig:schematic-pipeline-architecture}: the resampling \texttt{random\_state} is set to the bootstrap number. In this way, the resampling is exactly reproducible.

\subsubsection{Estimators}
In the context of the current system, an `estimator' can mean two things. An estimator is either: (1) a feature ranker or (2) a validation estimator. Still, the two are combined into a single estimator interface: which is because both can share much of the same API for interacting with them. Both implement a fitting- and scoring step, and can be of regressor- or classifier type. Also, to support these two learning tasks in a better way, any estimator is encapsulated in a generalized estimator interface. An example configuration for a \textbf{ranking} estimator can be seen in Listing~\ref{code:pipeline-estimator-example}.

\begin{lstlisting}[caption={An estimator config for ReliefF \citep{kononenko_estimating_1994}.}, label={code:pipeline-estimator-example}]
name: ReliefF
classifier:
  estimator:
    _target_: skrebate.ReliefF
regressor:
  estimator:
    _target_: skrebate.ReliefF
estimates_feature_importances: true
estimates_feature_support: false
\end{lstlisting}

Both a regressor and classifier were defined in a \textit{single} interface, as can be observed in Listing~\ref{code:pipeline-estimator-example}. Although the two were in this case defined to be the same, they might have as well pointed to different modules. Under the hood, the system detects the configured dataset task at initialization and instantiates either one of the two modules. Furthermore, the system has to be told the capabilities of the estimator, e.g., whether it estimates targets, feature importance or feature support. This is done using the boolean attributes starting with `estimates\_': where in this case the ReliefF estimator estimates just the feature importance, but no feature subset.

\textbf{Validation} estimators are configured in much the same way - and can also have both a regressor and classifier defined. The interface also allows indicating an estimator's support for \textit{multioutput} learning or the estimator requiring a \textit{strictly positive} dataset to work.

\subsubsection{Storage providers}\label{section:pipeline-storage-provider}
\begin{figure}[ht]
    \centering
    \includegraphics[width=\linewidth]{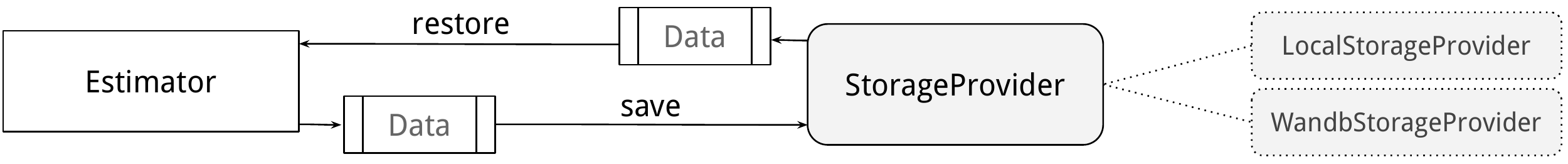}
    \caption{An illustration of the Storage Provider functionality. The storage provider can be used to save- and restore files, using both a local- or remote file system.}
    \label{fig:schematic-storage-provider}
\end{figure}

A \textit{storage provider} is defined to be a bridge between the pipeline and a file system, independent of whether this file system is on the local disk or in the cloud. Because in the pipeline it is desired that the fit estimators can be cached, a storage system must be built accordingly. Independent of exactly where the files are stored, a storage provider can be constructed by implementing routines for \textbf{saving} and \textbf{restoring} files. As long as the storage provider returns a usable file object given a file path, the system is ambivalent about how the file is loaded. A schematic of how this storage provider works is to be seen in Figure~\ref{fig:schematic-storage-provider}.

In the example in Figure~\ref{fig:schematic-storage-provider}, it can be seen that the storage provider is used by an estimator to save- and restore data. Such data can, for example, be a \textit{pickle} file used to serialize- and deserialize a fit estimator object.

\textbf{Caching} is supported in this way. By storing a reference to the file-path right in the job configuration itself, cached files can be reused via either the local- or remote file system. The pipeline allows configuring exactly \textit{which} parts of the pipeline allow using cached versions of their fit estimators, such that in any subsequent run of a job, some estimators can be reused and others overwritten. This functionality facilitates running multiple validation estimators given a single fit feature ranker, for example.

\subsubsection{Callbacks}\label{section:pipeline-callbacks}
The final component to discuss is the \textit{Callback}. A callback can be used to send data to a back end during the execution of the pipeline. Such data can be configurations, objects containing metrics or entire tables with results. The requirements for implementing such a callback are low, since any such data can be stored as an opt-in. If a pipeline user desires so, the usage of a callback can be completely ignored, storing only data to the disk using the storage provider. Sending metrics to a callback can, however, be a powerful way to send data to a database or cloud back-end right in the pipeline. In this way, the usual data collection phase can be automatized. About an implementation of such a callback can be read in Section~\ref{section:pipeline-visualization}. First, however, a closer look is taken at the pipeline execution and its scalability.

\subsection{Execution}\label{section:pipeline-execution}
The pipeline execution happens in a predefined number of steps. Because the pipeline is desired to be run in parallel and in a distributed way, the execution of such a system is nontrivial. Support for scaling in both the horizontal- and vertical directions are required, i.e., the pipeline must be able to scale over more machines but must also scale over more system resources such as CPU or RAM. Systems that fulfil such desires are \gls{hpc} systems, which are designed to support scalable computing \citep{ristov_superlinear_2016}. An elaboration is made on an implementation of such a scalable architecture.

\subsubsection{Multiprocessing}
\begin{figure}[ht]
    \centering
    \includegraphics[width=0.8\linewidth]{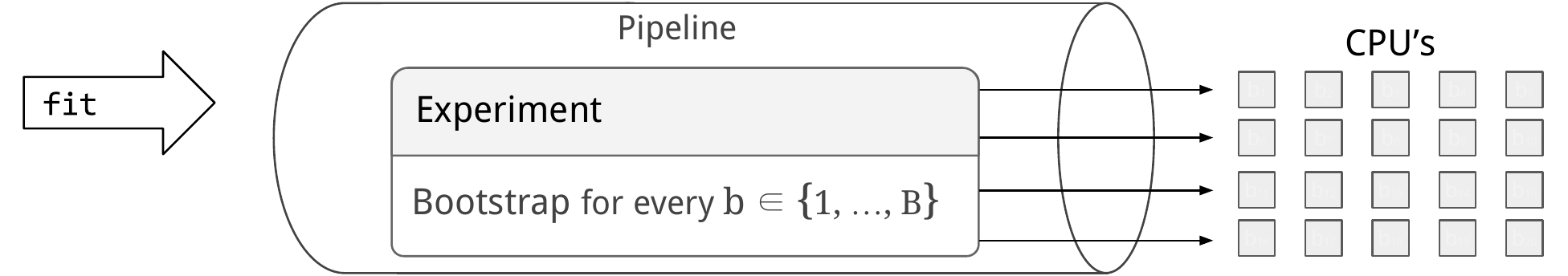}
    \caption{The pipeline multiprocessing distribution process. In the pipeline fit step the set of bootstraps is distributed over the available CPU's.}
    \label{fig:schematic-pipeline-multiprocessing}
\end{figure}

To facilitate \textit{vertical} scaling, a multiprocessing implementation is built. It is thereby the goal to best utilize the system resources at hand: maximizing the usage of the available processors. To do so, certain iterative parts of the pipeline were wrapped in an encapsulating `Experiment' module. The task of the module is, then, to pick up wherever the pipeline would iterate over a number of steps, and add multiprocessing support. As could be read in Section~\ref{section:pipeline-rank-and-validate} and seen in Figure~\ref{fig:schematic-pipeline-architecture}, the pipeline has two main loops: the first (1) runs $B$ bootstraps and the second (2) validates $\min (p, 50)$ feature subsets. In this implementation, the choice was made to distribute the former loop, i.e., to distribute the bootstraps over multiple CPU's. An illustration making clear the functionality of such a multiprocessing approach can be seen in Figure~\ref{fig:schematic-pipeline-multiprocessing}.

As could be seen in Figure~\ref{fig:schematic-pipeline-multiprocessing}, every bootstrap resampling job is assigned to a processor. In case more bootstraps are to be run than the number of available processors, i.e., $B >$ \# CPU's, some CPU's are assigned more than one bootstrap resampling job. In \gls{hpc} environments, it is common to have a dozen CPU's available in a single job. Given that a reasonable but still feasible number of bootstraps to run is also a few dozen, it can be straight-forward option to configure the number of bootstraps to run as the number of CPU's available in a job on the designated \gls{hpc} environment.

Important to note in such a multiprocessing job, is the available \textit{scope} of variables accessible inside any subprocess. Because each subprocess runs isolated on a CPU, it cannot access variables exclusively available to the main thread process. In the case that such a subprocess requires access to variables from the main thread, either one of two options can be considered. One option is to pass down a copy of the variables in question to the subprocess: this, however, does not apply when such variables change during the program execution. In that case, the second option is to communicate between the subprocess and the main thread. In this way, functionality such as storing cache files (Section~\ref{section:pipeline-storage-provider}) can still be done in the main thread, if it is deemed necessary.

\subsubsection{Distributed computing}
To make the pipeline \textit{horizontally} scalable, an approach must be constructed for distributing the processes over multiple machines. Although running multiple jobs in a \gls{hpc} environment can be a relatively straight-forward practice, a choice must be made on the amount of processing to put in a single job. For example, if a system were in place that automatically distributes jobs over a number of processors and machines by itself, the horizontal- and vertical scaling approaches could be tackled in a single solution. In this scenario, an automatic job scheduler would assign very small jobs to nodes and processors by itself; the smallest experimental unit could be running one bootstrap on a single feature ranker. However, such an approach is rather complex and often unsupported by existing \gls{hpc} systems. Therefore, the choice is to go with a hybrid approach - a single \gls{hpc} job is assigned multiple experiments. In this way, long queue waiting times as a consequence of running many small jobs are prevented.

\begin{figure}[ht]
    \centering
    \includegraphics[width=\linewidth]{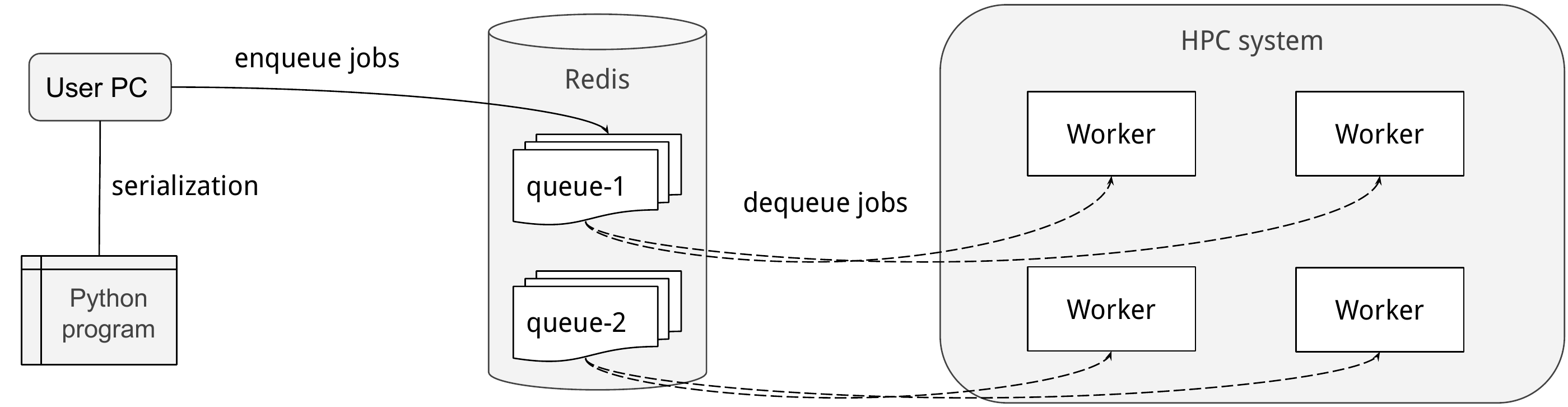}
    \caption{The \gls{rq} enqueueing- and dequeueing process. Jobs are serialized on the client and stored in Redis FIFO queue. Workers on the \gls{hpc} system process the jobs on the designated queues.}
    \label{fig:schematic-pipeline-redis-queue}
\end{figure}

Such an approach can be accomplished by the use of a second job queue, next to the \gls{hpc} job queue. The queue of choice is \gls{rq}, a library for keeping track of an arbitrary number of job queues in Redis which then can be executed by \textit{workers}. Workers are run on the \gls{hpc} system itself and are built solely to execute work coming from the queues. The library is built to be fault-tolerant, scalable, and easy to use. The basic idea is that during the enqueueing phase, a piece of Python code is serialized into a binary format and stored in Redis into a First-In-First-Out queue. Then, once the time has come that the job can be executed, the job execution code is deserialized from Redis back into Python code, and executed. This functionality can be seen illustrated in Figure~\ref{fig:schematic-pipeline-redis-queue}.

It can be seen in Figure~\ref{fig:schematic-pipeline-redis-queue}, that the Redis database acts as an intermediary between the user and the \gls{hpc} system. Whilst the queues in Redis may contain hundreds or thousands of jobs, the amount of jobs on the \gls{hpc} system only need to be several: one job for each worker. Each worker may run on its own node, or share a node by means of a virtual machine. In overall, the \gls{rq}-based approach has a number of benefits.

\begin{itemize}
    \item First of all, the \gls{hpc} system is not overloaded by small jobs. Instead, \gls{rq} is better designed to store and process a multitude of small jobs - preventing overhead and long queuing times in the \gls{hpc} system. This is because the only jobs that are launched on the \gls{hpc} system are the workers - which can be configured to last an arbitrary amount of time, or to terminate once its queue is empty.
    \item A second benefit is that \gls{rq} has good support for handling job failures. \gls{rq} allows automatically retrying failed jobs, with optionally set time intervals between tries, such that in the case of a network time-out or system error the job is be restarted without further human input. Furthermore, if the job still fails execution after its pre-configured number of retries were completed, the job is not just removed from all queues. Instead, the job is moved to a separate `failed job' queue. The error logs for jobs in this queue are easily inspected using a dashboard interface for \gls{rq}.
    \item A last benefit is the serialization- and deserialization process in \gls{rq}. Because Redis does not just store instructions to execute the code, but the actual execution code itself, code on the \gls{hpc} system might be changed during the time a job is in queue. This gives a practitioner more flexibility to change program code and run the pipeline at the same time.
\end{itemize}

As a side benefit, the Hydra library used for configuring the system has built-in support for working with \gls{rq}. By use of a plugin, the pipeline can be configured to enqueue its jobs to \gls{rq} instead of executing right away. An important note is that due to the serialization process the Python version during enqueueing and dequeueing have to be exactly equal.

\subsection{Data- collection and visualization}\label{section:pipeline-visualization}
Finally, all results coming out of the pipeline should be collected and visualized. Since the pipeline has support for plugging in any arbitrary `Callback' (Section~\ref{section:pipeline-callbacks}), primitives are in place for sending the data to a data- collection and visualization back-end. Any such callback can implement functions for processing the pipeline configuration, individual metric objects, and result tables. Because the pipeline is built to be executed at scale, a clear distinction is made between the pipeline execution phase and the data collection phase: the two are separated by means of the \textit{Callback} interface.

\begin{figure}[ht]
    \centering
    \includegraphics[width=0.9\linewidth]{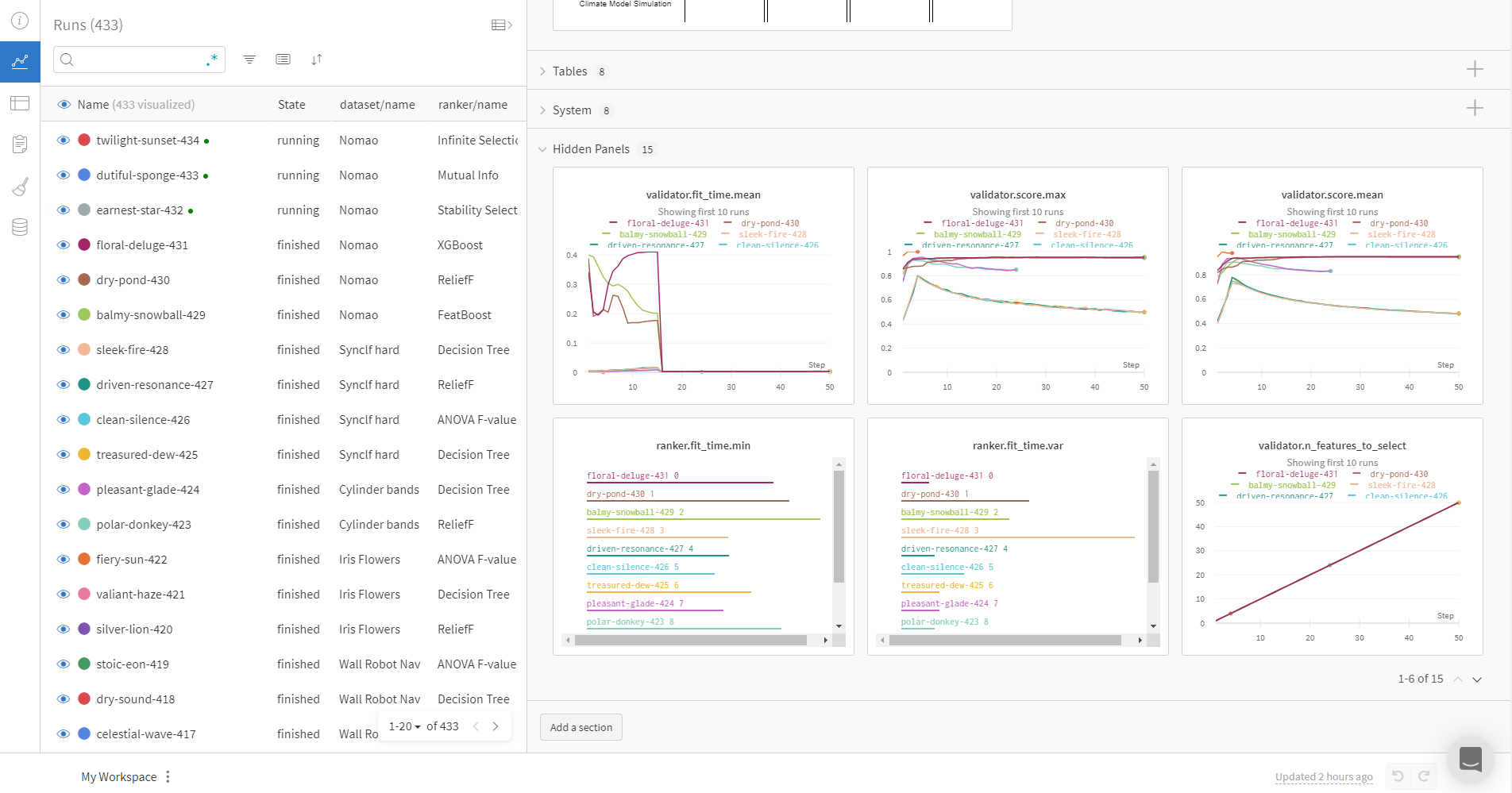}
    \caption{The Weights and Biases dashboard front-end. The platform allows tracking experiments, aggregating data, and visualizing results.}
    \label{fig:pipeline-wandb-frontend}
\end{figure}

One callback implementation is the `Wandb' callback, providing integration with the Weights and Biases platform \citep{biewald_experiment_2020}. Weights and Biases is a platform for real-time experiment tracking and visualization, with built-in primitives in Python. The platform allows you to upload tables, metrics, or configuration objects, and subsequently visualize them in the front-end interface. The platform also allows one to upload- and download raw files, allowing a user to integrate remote caching functionality (Section~\ref{section:pipeline-storage-provider}). A screenshot of the platform front-end can be seen in Figure~\ref{fig:pipeline-wandb-frontend}.

An important feature on the Weights and Biases platform is the ability to collect- and aggregate data interactively in the front-end. This is done by means of `Vega-Lite' \citep{satyanarayan_vega-lite_2017}, a visualization grammar for creating interactive visualizations in a front-end environment. This allows a user to send `raw' data to the dashboard, and aggregate it on the front-end itself. Any aggregation operation can be applied, be it taking averages, sums or computing the variances. The front-end then allows you to visualize the aggregated data using a simple grammar. In this way, the pipeline requires much less small adjustments and modifications for the user to be satisfied with the data aggregation process - the data can be summarized in a later stage after all results are already in.

\clearpage
\section{Experiments}\label{section:experiments}
Accompanying the proposal of a new feature- ranking and selection methodology, there is an experiment. The aim of the experiment is to show an example of what is possible with both the pipeline, and the newly proposed evaluation methodology. In this way, \gls{ml} practitioners and authors of new methods are able to conduct experiments themselves with the same setup and configuration: therefore allowing comparison between the experiments.

\subsection{Experiment setup}
The experiment closely follows the outlines of Section~\ref{section:evaluation} and Section~\ref{section:pipeline}. The setup can be summarized like so:

\begin{itemize}
    \item The dataset at hand is split into a 80\% training- and 20\% testing dataset. Whereas the training set is reserved for fitting the feature ranker and validator, the testing set is reserved for scoring the validator.
    \item For each Feature Ranker and dataset, $B = 25$ bootstrap resampling iterations are run.
    \item In each bootstrap resampling iteration, $\min (p, 50)$ feature subsets are evaluated, with each including the $k$ best features.
    \item Validation estimators are evaluated with the R\textsuperscript{2} score in case of regression and with accuracy in the case of classification. The validation estimators used are \gls{knn} with $k=5$ and a \gls{dt} at default sklearn settings.
    \item In the aggregation process, in case several experimental runs are found with the same configuration, i.e., for one Feature Ranker executed on one dataset, only the `best' run is considered. The best run is considered to be the run with the highest average mean score over all features, considering all bootstraps. This aggregation step will be elaborated upon in Section~\ref{section:experiments-example}.
\end{itemize}

All experiments were run on a SLURM \gls{hpc} environment. Specifically, the University of Groningen has a `Peregrine' compute cluster, with machines of various types, of which the most common one is a 24 core machine powered by two Intel Xeon E5 2680v3 CPUs running at 2.5 GHz. Per node, 128 GB memory and 1 TB internal disk space is available, but 10 GB was requested per CPU instead. This accounts for a total of 100 GB memory for 10 CPU processes. In total, the experimentation facilitating these results took 7682 hours of processing time on the above-mentioned machines.

A line-up of feature rankers and datasets was constructed to conduct benchmarking on. The full list of both is available in the Appendix Section~\ref{section:appendix-experiment-lineup}. See Table~\ref{table:experiments-ranker-specification} for the Feature Ranking line-up, and Table~\ref{table:experiments-dataset-specification} for the datasets line-up. The distribution of classes can be found in Appendix Figure~\ref{fig:experiments-datasets-class-distributions}. Most datasets are balanced, though also some imbalanced ones are included. Imbalanced datasets are not treated specially, it is up to the ranking- and validation estimators to correct for this. It can also be seen that rankers of various types are included: both regressors and classifiers, and rankers that support both learning tasks. Furthermore, some rankers were included that support multioutput datasets, i.e., datasets with multivariate targets.

\subsection{Experimental results for the `Synclf hard' dataset}\label{section:experiments-example}
To best understand the format of the experimental results and the accompanying metrics, a look is taken at the results for one dataset. In this way, a better understanding of the charts is gained first.

First of all, the considered dataset is the `Synclf hard' dataset. It is a synthetically generated dataset, created using the sklearn \texttt{make\_classification} function (Section~\ref{section:pipeline-components-datasets}). Its full configuration specification is defined in Listing~\ref{code:pipeline-synthetic-example}. The dataset is defined to have $n=10000$ samples and $p=50$ dimensions. After the \gls{cv} step was conducted, 8000 samples are left for training. The dataset has 4 relevant features and 3 target classes, which are perfectly balanced. That said, observations are first made on running one feature ranker on the dataset. For this, ReliefF \citep{kononenko_estimating_1994} is chosen.

\subsubsection{ReliefF performance}

\begin{figure}[ht]
    \centering
    \includegraphics[width=\linewidth]{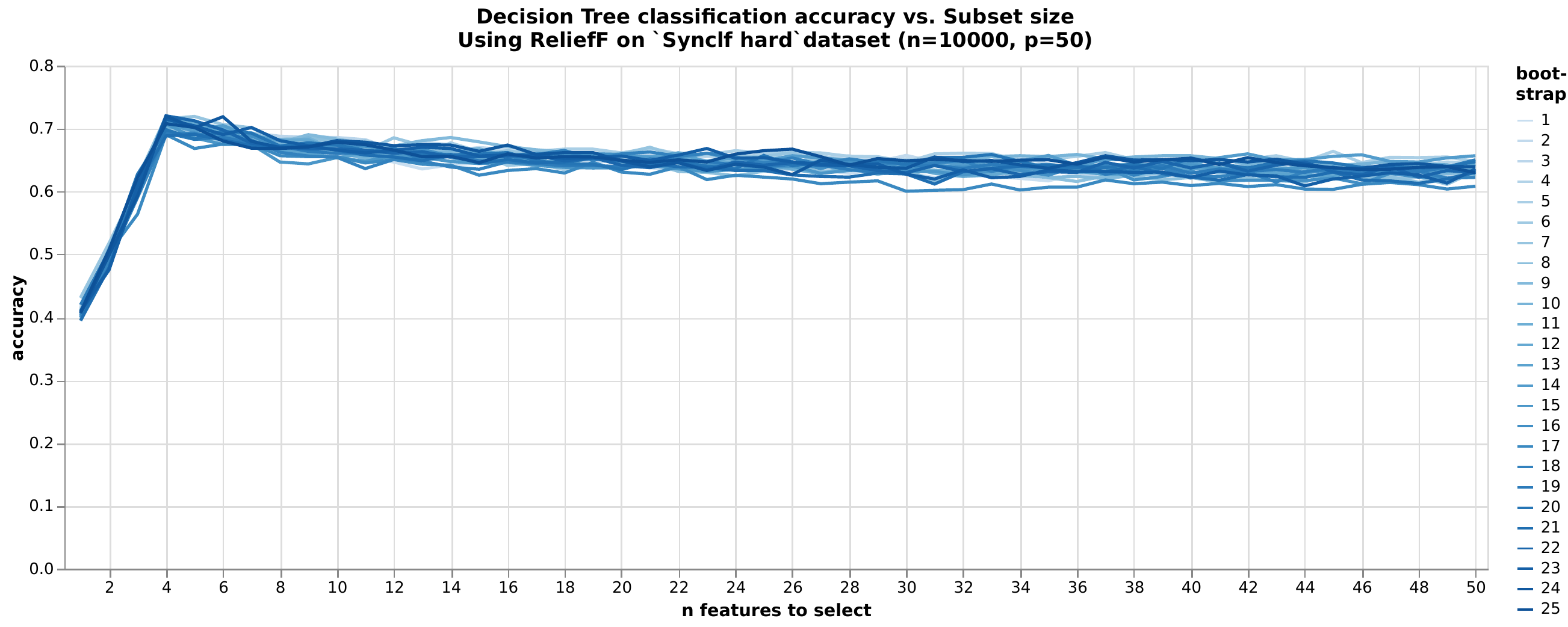}
    \caption{Validation performance for ReliefF on the `Synclf hard' dataset, for all 25 bootstraps.}
    \label{fig:results-validation-dt-relieff}
\end{figure}

To start, a look is taken at a plot that explicitly plots \textbf{all bootstraps}. See Figure~\ref{fig:results-validation-dt-relieff}.

As can be seen in Figure~\ref{fig:results-validation-dt-relieff}, the various bootstraps have had an effect on the validation performance per subset. Due to the random resampling with replacement taking place in the bootstrap phase, the feature ranker has to deal with permutations of the dataset each run. Indeed, this randomness is reflected in the validation performance.

One clear pattern is the fact that the validation performance first goes \textbf{up}, peaks at 4 features, and goes gradually down again. The fact that the peak happens at 4 features is clarified by the fact that the dataset had 4 informative features defined, meaning that the feature ranker correctly ranked the four informative features in its top-4 in most bootstraps. An intuition for the performance degradation is the fact that adding noisy dimensions can actually cause the estimator performance to degrade.

\begin{figure}[ht]
    \centering
    \includegraphics[width=\linewidth]{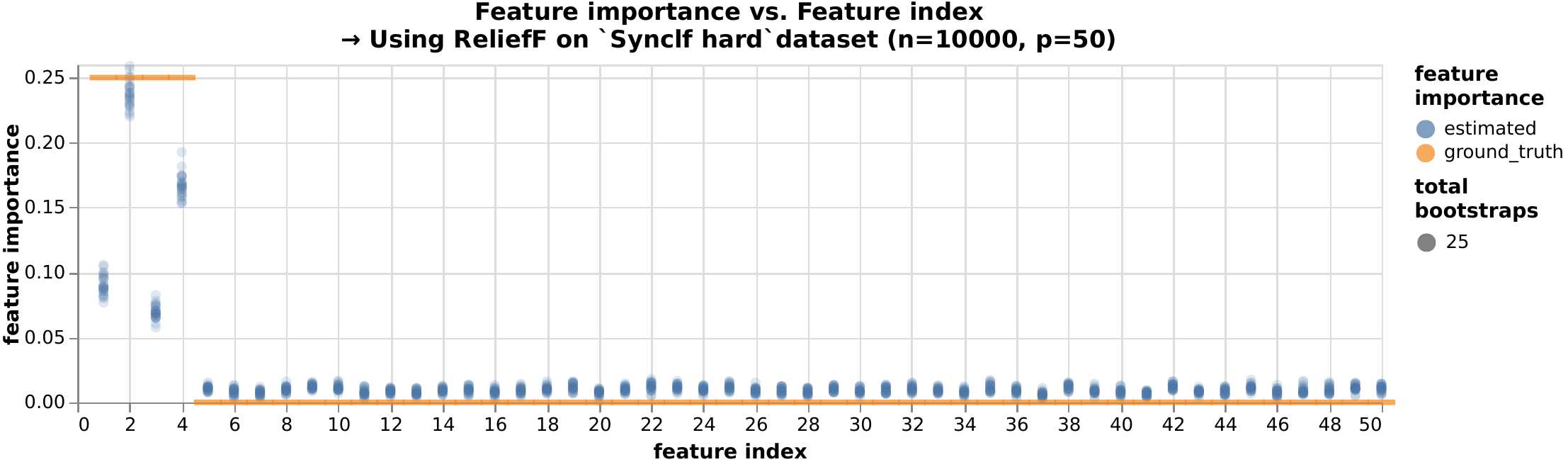}
    \caption{Estimated feature importances for ReliefF on `Synclf hard'.}
    \label{fig:results-importances-relieff}
\end{figure}

Next, a better look is taken at the estimated \textbf{feature importances}. Since the validation performance suggests that the ranker correctly identifies the importance of the informative feature, it is expected that this is reflected in the feature importance estimates. Indeed, this is the case. See Figure~\ref{fig:results-importances-relieff}.

It can be seen in Figure~\ref{fig:results-importances-relieff} that the first four features were assigned a larger importance than the others. Indeed, these are the ground-truth relevant features, as is known \gls{apriori} because the dataset was synthetically generated. Besides the `estimated' feature importances, also the \gls{gt} feature importance is visible. At all times, the \gls{gt} gets the same treatment as the estimated feature importances: it is also normalized to a probability vector. In this way, the two vectors can be compared fairly.

Using the information shown in Figure~\ref{fig:results-importances-relieff}, the R\textsuperscript{2} score and Log loss can now be computed between the estimated- and \gls{gt} feature importances. The computation is elaborated upon in Section~\ref{section:evaluation-apriori-knowledge}. The scores are shown in Table~\ref{table:experiments-relieff-gt-metrics}.

\renewcommand\theadalign{bl}
\begin{table}[ht]
    \centering
    \begin{tabular}{| l | l | l | l | l | l | l |}
    \hline
    \thead{Metric} & \thead{Mean} & \thead{Stdev} \\
    \hline
    R\textsuperscript{2} score & $0.6962$ & $\pm 0.0151$  \\ 
    \hline
    Log loss & $0.1749$ & $\pm 0.002517$\\ 
    \hline     
    \end{tabular}
    \caption{The R\textsuperscript{2}- and log loss metrics for ReliefF on `Synclf hard'. Scores were aggregated over 25 bootstraps, see Figure \ref{fig:results-importances-relieff}. Scores were computed like explained in Section~\ref{section:evaluation-apriori-knowledge}.}
    \label{table:experiments-relieff-gt-metrics}
\end{table}

The metrics in Table~\ref{table:experiments-relieff-gt-metrics} are to be interpreted as follows. In the case of the R\textsuperscript{2} score, a \textit{higher} score means a better result, i.e., means the ranker estimated the feature importances closer to the ground truth. The log loss, on the other hand, is desired to have a \textit{low} value; meaning the difference between the estimated- and the \gls{gt} importances were smaller. Next, lets see how to quantify the stability of the run.

\begin{figure}[ht]
    \centering
    \includegraphics[width=0.9\linewidth]{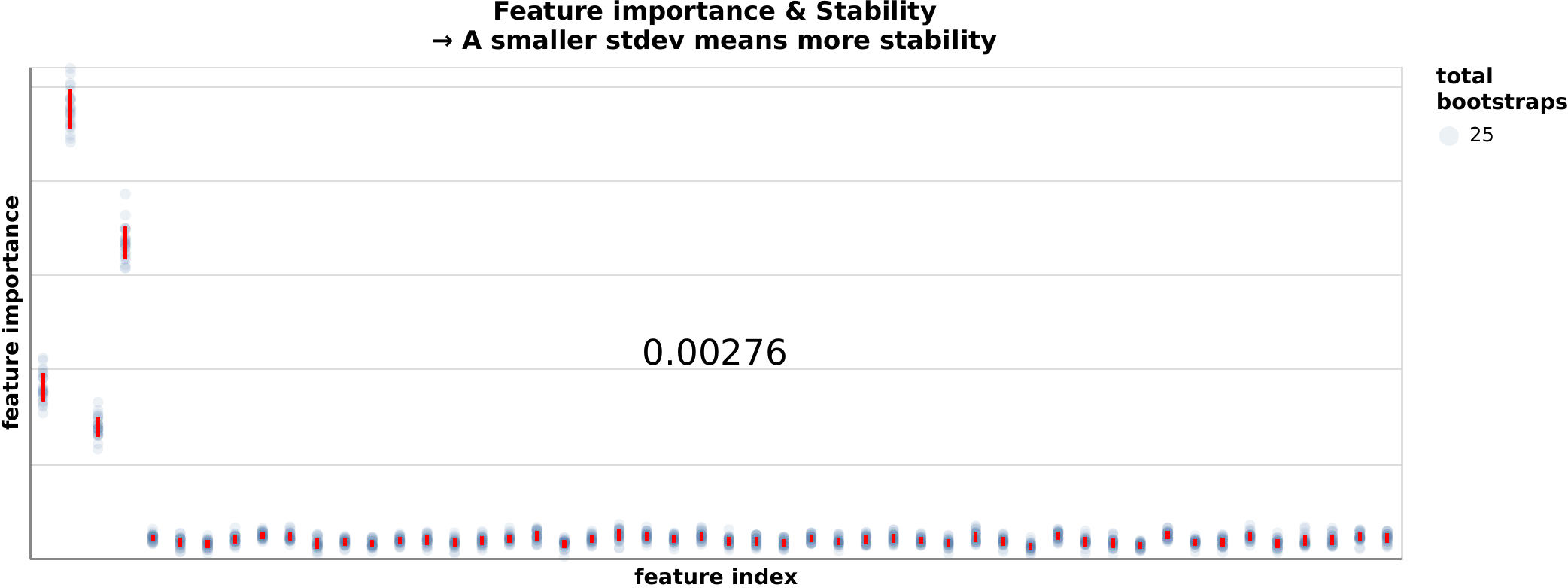}
    \caption{Feature importance scores with stdev error bars and the total mean score visible. A higher stdev score means less stability. Shown for ReliefF on the `Synclf hard' dataset.}
    \label{fig:results-importances-stability-relieff-example}
\end{figure}

To quantify the \textbf{stability} of estimated feature importance scores, the standard deviations of the feature importances are taken. First, the stdev score is computed for each feature over the $B$ bootstraps, and then the mean is taken over over the stdev scores. This is like it was explained in Section~\ref{section:feature-importance-stability}. The score can be plotted like it can be seen in Figure~\ref{fig:results-importances-stability-relieff-example}.

It can be seen in Figure~\ref{fig:results-importances-stability-relieff-example}, that the instabilities take place mainly in the relevant features. This means the algorithm is confident on which features \textit{not} to choose, but it is not sure about the weighting of the relevant ones. The mean stdev over all features is illustrated as text on the plot. Thanks to the red error bar lines, it can be quickly observed where the algorithm is instable. 

\subsubsection{Performance of multiple rankers}

\begin{figure}[ht]
    \centering
    \includegraphics[width=\linewidth]{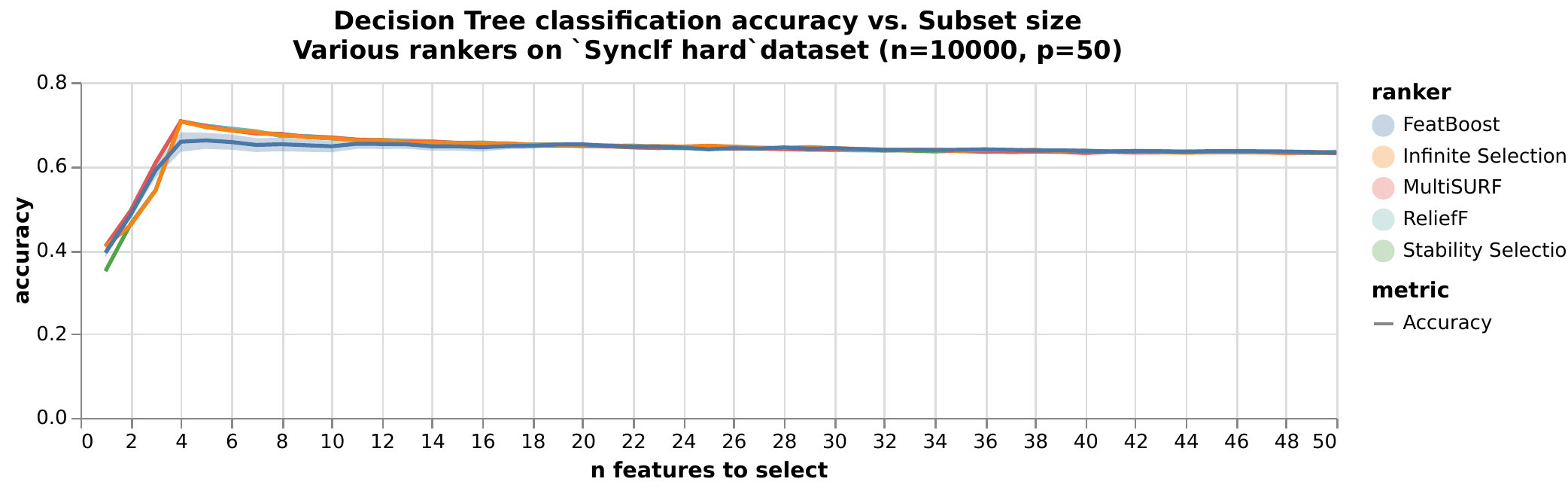}
    \caption{\gls{dt} validation performance of several rankers on the `Synclf hard' dataset.}
    \label{fig:results-validation-dt-various-rankers}
\end{figure}

Next, multiple rankers are considered. Again, the dataset at hand is the `Synclf hard' dataset. The validation performance of all rankers can now be compared in a single plot, see Figure~\ref{fig:results-validation-dt-various-rankers}.

In this case (Figure~\ref{fig:results-validation-dt-various-rankers}), the validation curve is \textbf{aggregated} over the bootstraps. That is, for each point on the curve, the average score is taken over the $B$ bootstraps. Looking at the plot, indeed, the various rankers show similar behavior like before in the validation performance. For most rankers, performance goes up, peaks at 4 features, and goes down again. Some rankers, however, do not manage to pick out the most useful features right away, and see their performance peak a bit later. It can also be the case that some rankers selected the most relevant features \textit{earlier} in their feature subsets, causing higher values at the start of the curve.

\begin{figure}[ht]
    \centering
    \includegraphics[width=\linewidth]{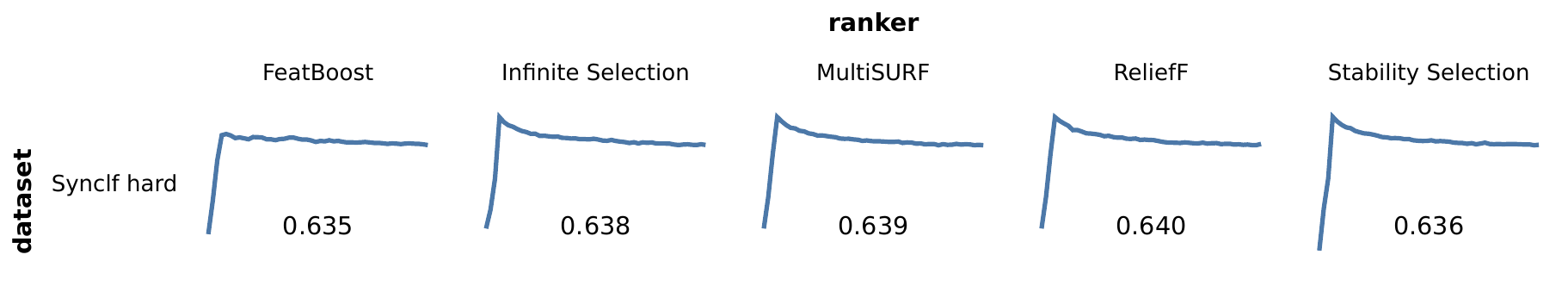}
    \caption{Aggregated \gls{dt} validation curves with their \textbf{mean validation score} shown. The mean score is the average curve value, allowing summarizing the curve with a single value. Curves are the same as in Figure~\ref{fig:results-validation-dt-various-rankers}.}
    \label{fig:results-validation-with-mean-score}
\end{figure}

Another way to plot the validation curves is to put the curves side by side, instead of stacked upon each other. This allows illustrating an important aggregation metric: the \textbf{mean validation score}. As could be seen before, the validation curves are already assumed to be aggregated over the bootstraps. This means, the results for each ranker are now reduced to $\min(p, 50)$ values. To further reduce this value and be able to summarize the performance with a single \textit{scalar}, the mean of the aggregated validation curve is taken. See Figure~\ref{fig:results-validation-with-mean-score}.

Looking at the curves (Figure~\ref{fig:results-validation-with-mean-score}), it can be seen that the mean validation score scores higher where the curves scored a higher total score over all feature subsets. In this regard, the mean validation score could also be expressed as the sum of scores, or even the `area under the validation curve'. Because the tick steps on the x-axis are in this case always one, however, it is chosen to take the mean score instead. The detail lost by computing the interpolated curve pieces between the curve points is negligible. Taking the mean validation score is also a much easier metric to compute than the AUC, which in this case would require lots of interpolation. The mean validation score can be seen to represent the performance of the feature ranker reasonably well. When the ranker scores the relevant features as highly informative, the validation score appropriately reflects this. Feature rankers that have a low area under the curve are also seen to have a lower score.

\begin{figure}[ht]
    \centering
    \includegraphics[width=\linewidth]{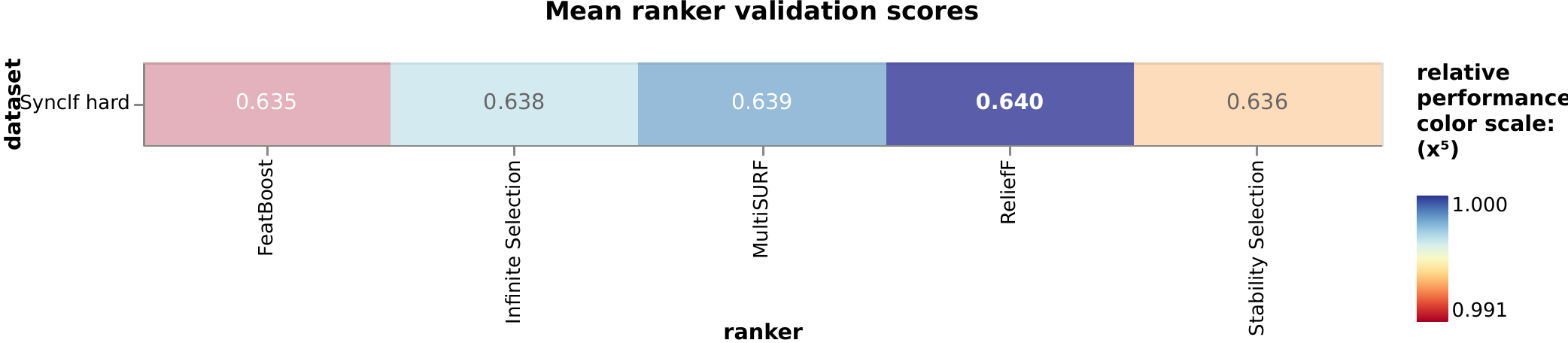}
    \caption{Mean \gls{dt} validation scores plotted with emphasized \textbf{relative performance}. The \textit{relative performance} is a ranker's mean validation score as a fraction of the highest achieved score by any ranker for some dataset. Same curves as in Figure~\ref{fig:results-validation-dt-various-rankers} and Figure~\ref{fig:results-validation-with-mean-score}.}
    \label{fig:results-validation-with-mean-score-tabular}
\end{figure}

Next, the previously plotted mean validation scores are presented in yet another format. This time, it is desired to emphasize the relative performance of a ranker, compared to the others. This can be done by using adaptive cell background colors, see Figure~\ref{fig:results-validation-with-mean-score-tabular}.

This time (Figure~\ref{fig:results-validation-with-mean-score-tabular}), the summarized performance of each feature ranker is visible in an instant. The coloring of the chart adjusts according to the \textbf{relative performance} of the ranker, with respect to the best performing ranker. This is computed simply by taking the ranker's mean validation score as a fraction of the highest score achieved by any ranker. This means that a relative performance of $1.00$ resembles the highest performing ranker, and a relative performance of $0.50$ would mean a ranker gets only half of the highest performing ranker's score. An accompanying color scale illustrates this fact, where blue resembles a higher score and red a lower score. Furthermore, the differences are further exacerbated by use of an \textit{exponential} color scale. In this case, an exponent of $x^5$ is chosen, to be able to better differentiate between the top performers. The best score also gets its font weight displayed as \textit{bold}, such to emphasize the best performing ranker.
It is thereby important to note that in case \textit{multiple} datasets are considered, each dataset gets its own relative performance scale. This is because the scores across multiple datasets ought not to be compared directly to each other.

\begin{figure}[ht]
     \centering
     \begin{subfigure}[b]{0.49\textwidth}
        \centering
        \includegraphics[width=\linewidth]{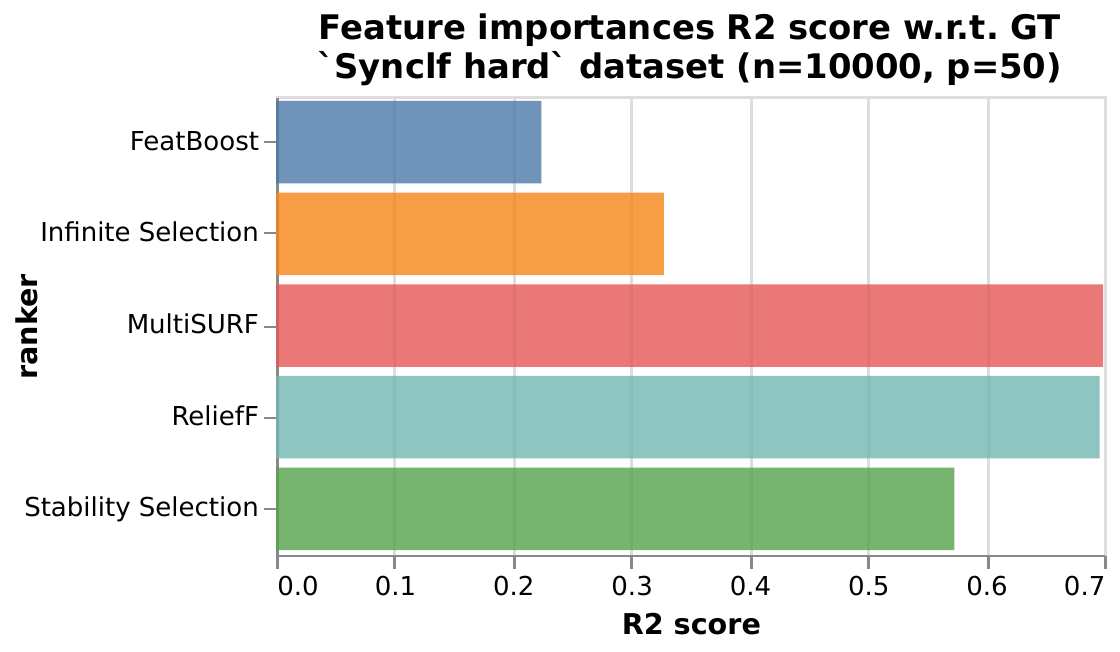}
        \caption{The average R\textsuperscript{2} score over all bootstraps. A \textbf{higher} score is better. Computed like explained in Section~\ref{section:feature-importance-definition}.}
        \label{fig:results-r2-score-various-rankers}
     \end{subfigure}
     \hfill
     \begin{subfigure}[b]{0.49\textwidth}
        \centering
        \includegraphics[width=\linewidth]{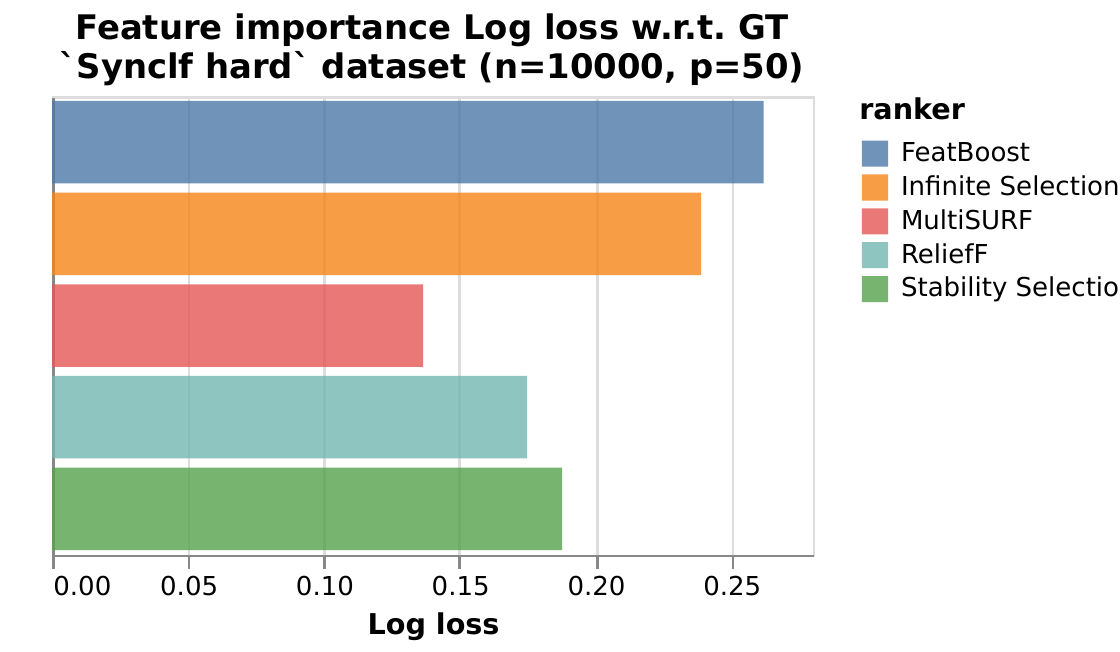}
        \caption{The average log-loss score over all bootstraps. \textbf{Lower} is better. Computed like explained in Section~\ref{section:feature-importance-definition}.}
        \label{fig:results-log-loss-various-rankers}
     \end{subfigure}
     
    \caption{Log loss and R\textsuperscript{2} scores, computed by comparing the \textit{estimated} feature importance scores against the feature importance \textit{ground-truth}. Such a ground-truth is available when datasets are synthetically generated. Scores shown for `Synclf hard' dataset on multiple rankers.}
    \label{fig:results-ground-truth-various-rankers}
\end{figure}

Again the R\textsuperscript{2}- and log loss scores are investigated, which are computed using the ground-truth relevant features. The hypothesis is that the scores are able to `predict' the validation estimator's performance. This is useful because, by evaluating the quality of the feature ranking directly, one could already make meaningful conclusions, just by having run the ranker alone. A comparison chart for these metrics is plotted in Figure~\ref{fig:results-ground-truth-various-rankers}.

For Figure~\ref{fig:results-ground-truth-various-rankers} it is, first of all, important to note that the R\textsuperscript{2}- and log loss scores were computed as an average score over all bootstraps. This means that after running a single ranker on a single dataset, $B$ such scores are obtained - which are then aggregated into a single score by taking its mean. In the figure it can be seen, that in the case of the R\textsuperscript{2} score the scorings represent the scores presented in Figure~\ref{fig:results-validation-with-mean-score} and Figure~\ref{fig:results-validation-with-mean-score-tabular} very well. The order of best performance w.r.t. mean validation score was predicted by the R\textsuperscript{2} score: wherever the R\textsuperscript{2} score is higher, the mean validation performance is also higher. The same applies to the log loss scores: more loss means worse mean validation performance. Although in this example the differences are very small, still a measure of proportionality between the scores exists, for this dataset.

\begin{figure}[ht]
    \centering
    \includegraphics[width=\linewidth]{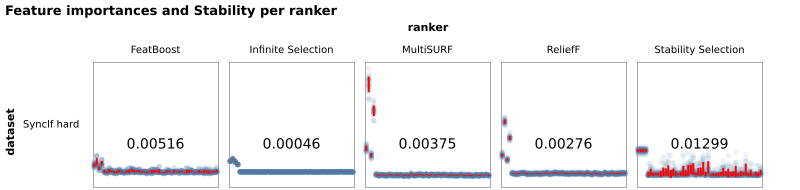}
    \caption{The estimated feature importances and their stabilities, quantified using the mean standard deviation over all feature importances. Plots shown for `Synclf hard' dataset.}
    \label{fig:results-importances-stability-multiple-rankers}
\end{figure}

To assess the \textbf{stability} of the rankers, the feature importances of the various rankers can be put side-by-side. Then, the standard deviations of each feature estimated over $B$ bootstraps can be shown. This can be seen illustrated in Figure~\ref{fig:results-importances-stability-multiple-rankers}.

As can be seen in Figure~\ref{fig:results-importances-stability-multiple-rankers}, the stability of the ranking algorithms varies. As illustrated by the red standard deviation error bars, the algorithms have varying amounts of deviation in their feature importance estimations. Whereas Infinite Selection is very stable, and shows hardly any variance over the bootstraps at all, Stability Selection is very unstable, despite its name. MultiSURF can be seen to be varying mainly in the relevant features, similarly to ReliefF. Looking at the summarized stability value, indeed the expected instabilities are captured in the scalar. Stability Selection has the highest mean standard deviation and Infinite Selection has the lowest. This indicates the metric might be successful in capturing the instability of the algorithms.

Now that all relevant plots are clarified and understood, a broader view can be taken. In the next section, the experiment in its entirety will be considered.

\subsection{Experimental result for all datasets}
\begin{figure}[ht]
    \centering
    \includegraphics[width=\linewidth]{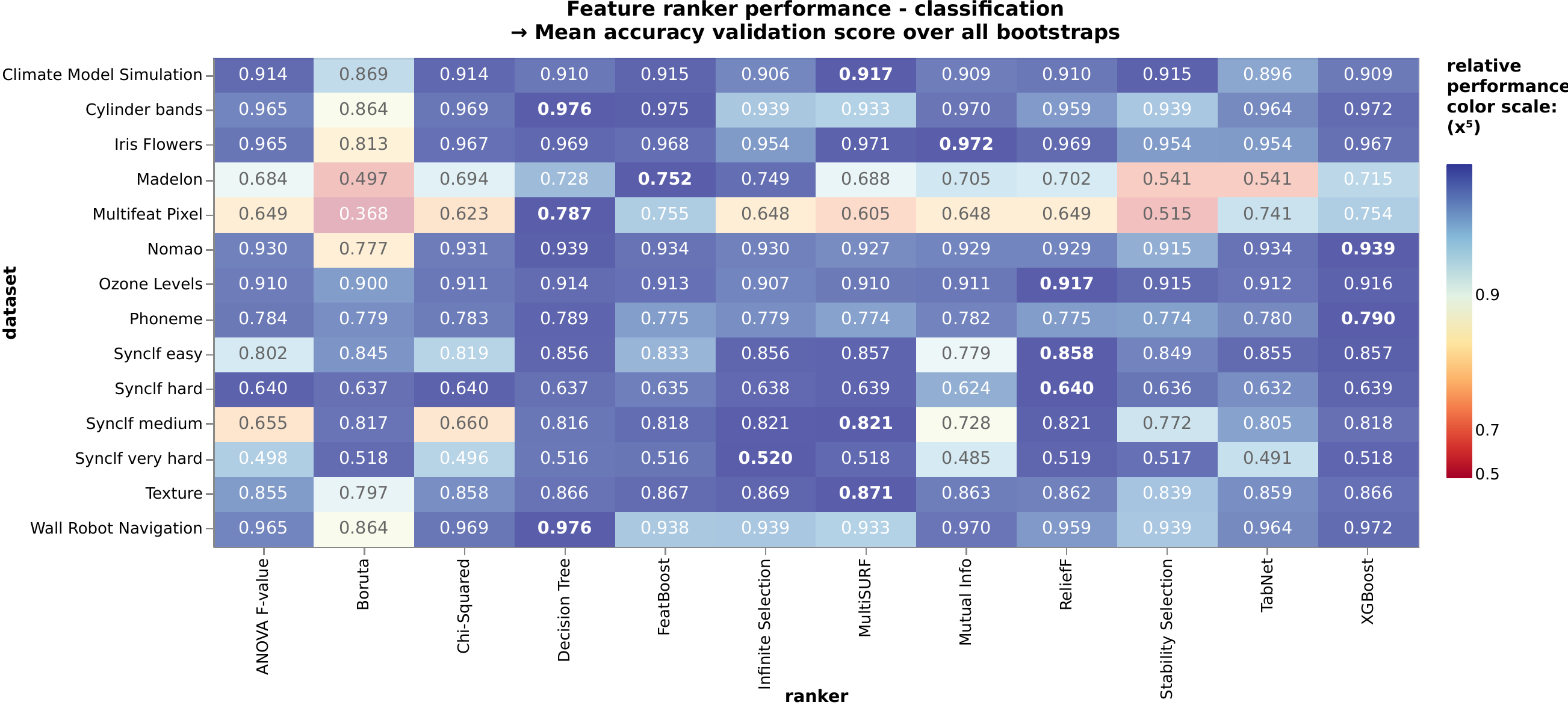}
    \caption{Mean \textbf{classification} accuracy validation scores for all datasets and all rankers. Color scale is configured with \textit{relative performance}, like explained in Section~\ref{section:experiments-example}. Validation scores shown are for classification datasets using a \gls{dt} as a validator.}
    \label{fig:results-all-datasets-mean-validation-dt}
\end{figure}

Now, a look is taken at the results of \textit{all} datasets. See Figure~\ref{fig:results-all-datasets-mean-validation-dt}, which is explained below.

To illustrate the performance of the complete experiment, different ways of visualization are required. With this many combinations of feature rankers and datasets, no longer can the line curves showing the mean validation scores be used to explain the findings. Instead, it is chosen to display this data in a heatmap, exactly similar to Figure~\ref{fig:results-validation-with-mean-score-tabular}, but with more data. Various aspects of performance are discussed, starting with the \textbf{mean validation} performance.

In Figure~\ref{fig:results-all-datasets-mean-validation-dt} it can be seen that the table coloring reveals much about the performance of the feature ranker. The more the coloring is toward dark \textit{blue}, the better the ranker performed. On the other hand, red indicates bad performance: the worst case in this plot is Boruta, which only achieves about half the score of the best performer for the `Multifeat Pixel' dataset. Boruta's bad performance is almost certainly due to a configuration anomaly, since the performance is worse than random in some cases. A notable performer is the \textbf{\gls{dt}}, used as a Feature Ranker by making use of its computed feature importance scores. The algorithm scores high on nearly every dataset, making it a good contender for the best performer. It might seem odd, however, that a \gls{dt} is run for Feature Selection first, and then \textit{also} for validation. For this reason, the experiment was also run with another validation estimator, \gls{knn}. For this, see Figure~\ref{fig:results-all-datasets-mean-validation-knn} in the Appendix.

\begin{figure}[ht]
    \centering
    \includegraphics[width=\linewidth]{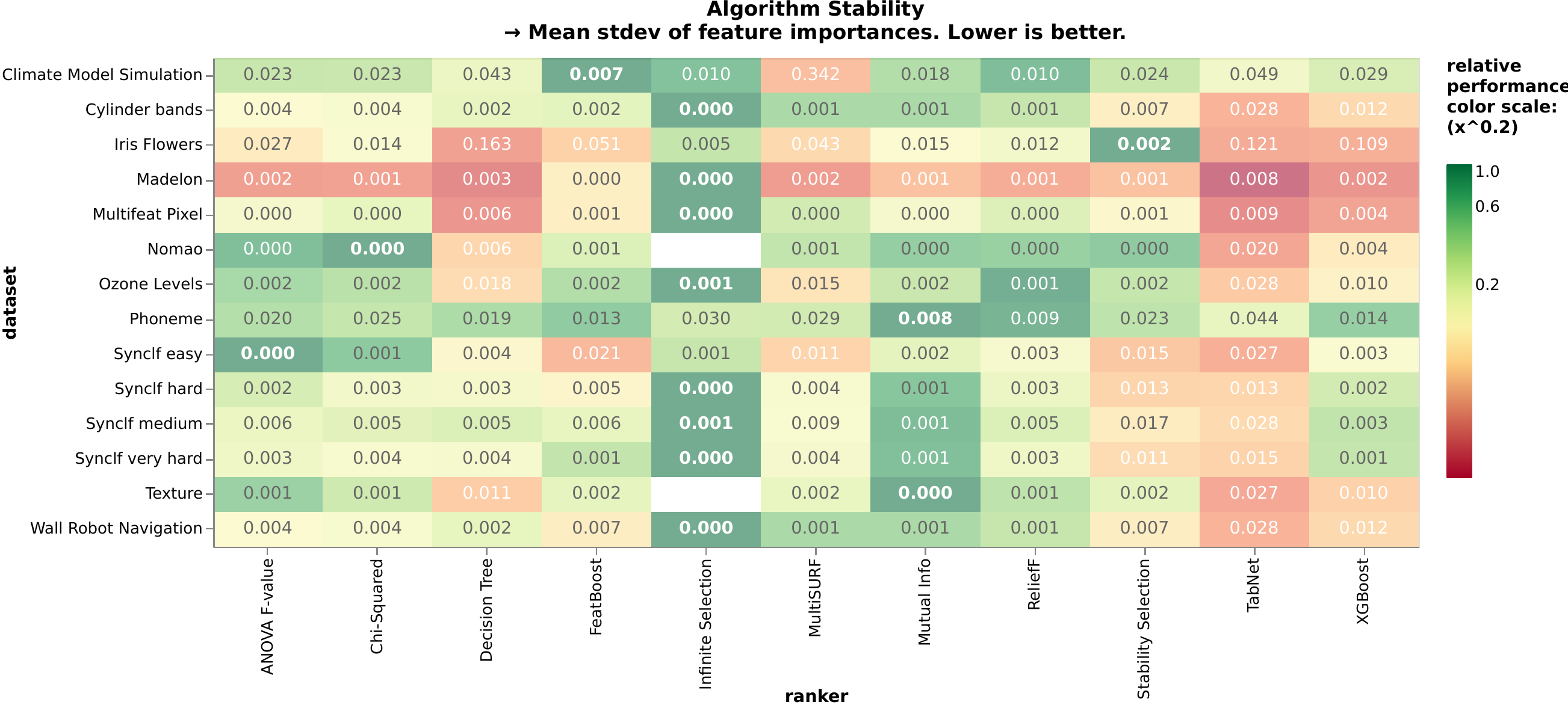}
    \caption{Algorithm stability scores. Obtained by computing the standard deviation over the feature importance vectors, like illustrated in Section~\ref{section:experiments-example} and explained in Section~\ref{section:feature-importance-stability}. A custom color scheme was chosen to better emphasize the differences. A darker shade of red means worse stability. A darker shade of green means better stability. The empty cells are due to an experiment failure.}
    \label{fig:results-all-datasets-stability-scores}
\end{figure}

Next is the algorithm \textbf{stability}. The algorithm stability scores were computed like illustrated in the previous section, but now for more datasets. The scores were represented in a table and have a custom color scheme applied to it to better emphasize the differences. See Figure~\ref{fig:results-all-datasets-stability-scores}.

Like can be seen in Figure~\ref{fig:results-all-datasets-stability-scores}, the stability scores differ per ranker. Especially TabNet, XGBoost and Decision Tree can be seen to be relatively unstable. Infinite Selection, Mutual Info and ReliefF, on the other hand, have a better stability.

\begin{figure}[ht]
    \centering
    \includegraphics[width=\linewidth]{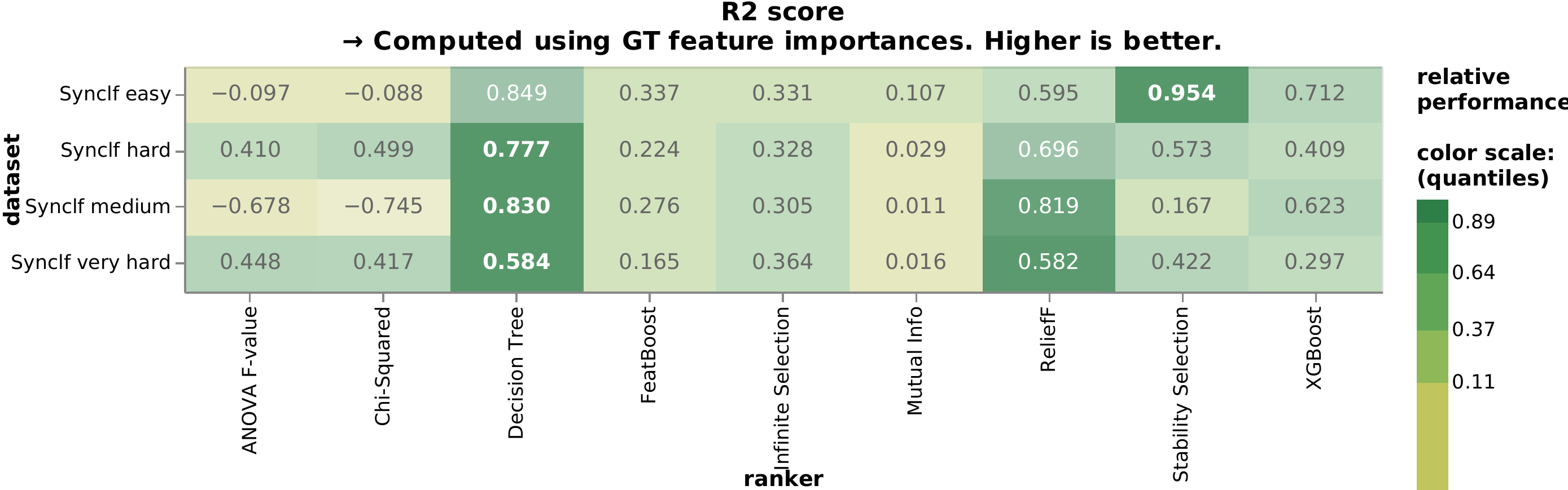}
    \caption{Algorithm R\textsuperscript{2} scores, computed using the \gls{gt} feature importanes. The computation was explained in Section~\ref{section:evaluation-apriori-knowledge} and illustrated in Section~\ref{section:experiments-example}. A darker shade of green means a better score.}
    \label{fig:results-all-datasets-r2-scores}
\end{figure}

Next, a look is taken at the computed R\textsuperscript{2} scores. The scores were computed using the \gls{apriori} known relevant features, i.e., the datasets \gls{gt} feature importances. A comparison heatmap plot can be seen in Figure~\ref{fig:results-all-datasets-r2-scores}.

As could be seen in Figure~\ref{fig:results-all-datasets-r2-scores}, the differences in R\textsuperscript{2} scores are notable, and correlate slightly with the mean validation scores for the designated datasets. Both \gls{dt} and ReliefF have high R\textsuperscript{2} scores for the chosen datasets. FeatBoost, however, had a high overall mean validation score but cannot be seen to have a high R\textsuperscript{2} score. This might indicate that the scoring does not always foretell validation performance.

Especially in the case of FeatBoost and Infinite Selection, there exist big discrepancies between the mean validation score and the computed R\textsuperscript{2} scores. Whereas for both rankers the mean validation scores are high for the `Synclf' datasets, the R\textsuperscript{2} scores did not follow. A reason for this might be the fact that these rankers do often rank features as having an importance of above zero. Whereas an algorithm might still get the feature importance \textit{order} correct, the algorithm does a worse job in the R\textsuperscript{2} scoring due to always giving features scores of well above zero. Other rankers, that assign more importance to some presumably relevant features but less to the others, will be better off with the R\textsuperscript{2} scoring. The metric might therefore pose an unfair advantage to how rankers behave in this regard.

\begin{figure}[ht]
    \centering
    \includegraphics[width=\linewidth]{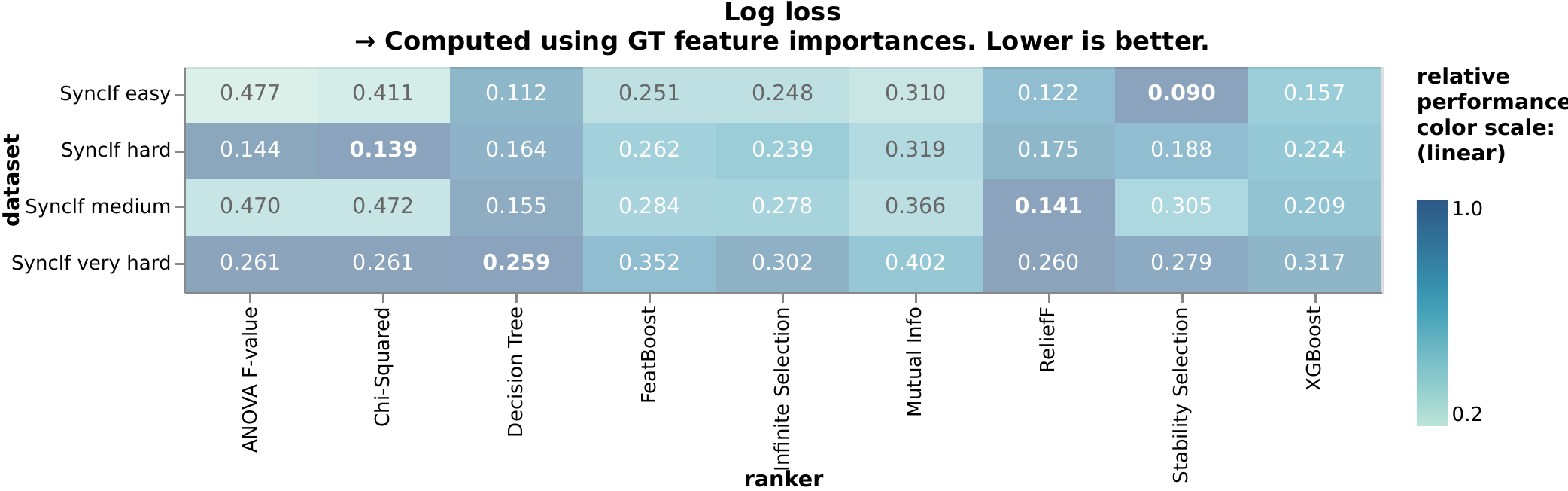}
    \caption{Algorithm Log loss scores, computed using the \gls{gt} feature importanes. The computation was explained in Section~\ref{section:evaluation-apriori-knowledge} and illustrated in Section~\ref{section:experiments-example}. A darker shade of blue means a better score.}
    \label{fig:results-all-datasets-log-loss}
\end{figure}

In a similar fashion, the log loss scores are plotted. See Figure~\ref{fig:results-all-datasets-log-loss}.

In the case of the log loss metric, a different pattern is visible (Figure~\ref{fig:results-all-datasets-log-loss}). Indeed, well performing rankers are \gls{dt} and ReliefF. The scoring did miss FeatBoost and Infinite Selection, however, as being overall good performers. Whereas according to the \textit{mean} validation scores FeatBoost and Infinite Selection are among the top performers, this is not reflected in the log loss score. This might, again, be due to an unfair advantage given to rankers that tend to score irrelevant features closer to zero. This is similar to the R\textsuperscript{2} score, though arguably the effect is exaggerated in the case of the R\textsuperscript{2} score. The metrics might therefore not be ideal for direct algorithm comparison: it is best the mean validation scores are included at all times. The metrics can, however, reveal interesting information in specific scenarios. This might be the case, for example, when no feature selection is applied at all in the evaluation process, and all one has is the feature importance ground truth and estimations.

Additionally, some plots can be seen in the Appendix. A plot showing the validation scores using \gls{knn} as the validation estimator can be found in the Appendix Figure~\ref{fig:results-all-datasets-mean-validation-knn}. To see the R\textsuperscript{2} scores and Log loss scores for the \textit{regression} datasets, see Figure~\ref{fig:results-all-datasets-log-loss-regression} and Figure~\ref{fig:results-all-datasets-r2-score-regression}, respectively. The regression dataset mean validation scores can be seen in Figure~\ref{fig:results-all-datasets-mean-validation-dt-regression}.

\subsection{Learning curves and time complexity}
\begin{figure}[ht]
    \centering
    \includegraphics[width=\linewidth]{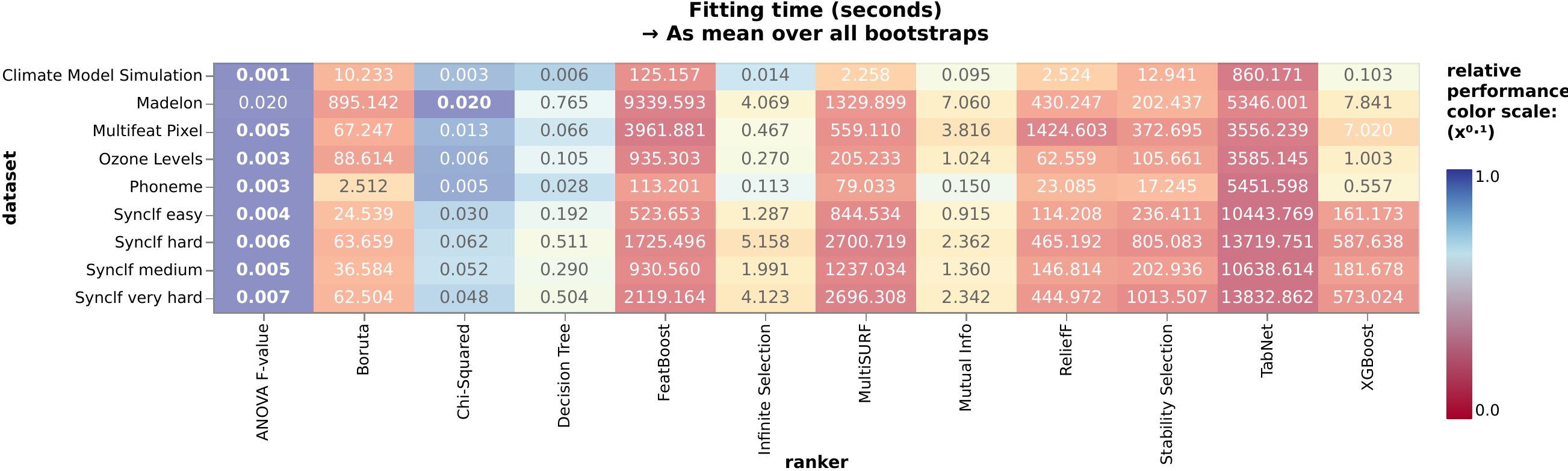}
    \caption{Mean fitting times from running the experiment $B=25$ times. Fitting times are represented in seconds. A custom color scheme and scaling factor was chosen to emphasize the differences in fitting times.}
    \label{fig:results-all-datasets-fitting-time}
\end{figure}

To first get an idea of the algorithm time complexity, an overview plot is taken into consideration. In this plot, multiple datasets and multiple rankers are considered. For each benchmark, the fitting time in seconds is plotted. See Figure~\ref{fig:results-all-datasets-fitting-time}.

In Figure~\ref{fig:results-all-datasets-fitting-time} it can first of all be seen that the algorithms differ immensely in fitting times. TabNet is (by far) the most complex and time-consuming estimator to fit. This is much due to the fact that it fits a sophisticated Neural Network and uses PyTorch to do so. The two algorithms that are next in time complexity are MultiSURF and FeatBoost. Some statistical estimators can be seen to have very low fitting times.

To investigate algorithm performance under varying conditions of \textbf{sample size}, a separate experiment was run. In this experiment, only the `Synclf hard' dataset is considered. Then, the sample size for this dataset was varied over a number of intervals, set to a fixed number each time. The sample size was set to range from 100 to 2,100 with intervals of 200. Note, that the total sample size for this dataset is 10,000. For every sample size, 10 bootstraps were run.

\begin{figure}[ht]
    \centering
    \includegraphics[width=\linewidth]{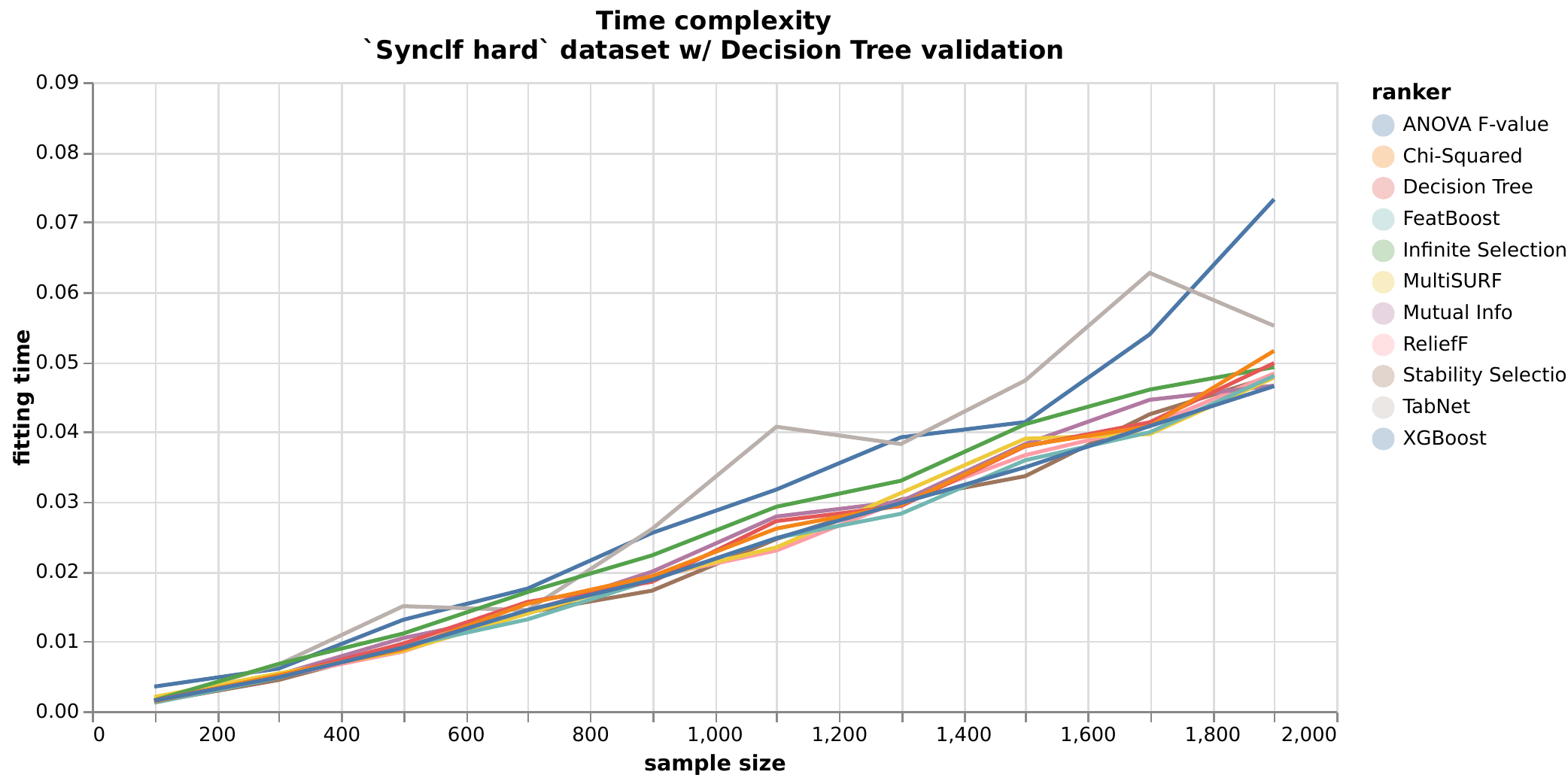}
    \caption{Time complexity plot for `Synclf hard' dataset. Sample sizes from 100 to 2,100 were used, with intervals of 200. The plot shows the average fitting time over 10 bootstraps.}
    \label{fig:results-time-complexity}
\end{figure}

In this way, two things can be investigated. (1) the time complexity and (2) the learning curve behavior of the various algorithms. Both were plotted. First, a look is taken at the time complexity. To investigate the time complexity of the algorithms, the fitting time was recorded. The fitting time was measured to entail only exactly the ranker fitting step - no supplementary preprocessing steps whatsoever influenced the time recording. The time was stored in a high-accuracy float but represented as \textit{seconds}.  See Figure~\ref{fig:results-time-complexity}.

As can be seen in Figure~\ref{fig:results-time-complexity}, most algorithms scale about slightly worse than linear when increasing the number of samples. It can be seen that when 1,000 samples are used versus 2,000, the fitting time is about double. The only outliers are XGBoost and TabNet, which seem to need a bit longer than its competitors when the sample size exceeds 900.

The learning curve was omitted from the main text and can be seen in the Appendix. See Figure~\ref{fig:results-learning-curve}. It can be seen that the algorithms learn very similarly given an increasing number of available samples. Only stability selection and mutual info are having a slightly harder time learning the useful features with little data.

\subsection{Online dashboard}
In case the reader is interested in exploring the data interactively him- or herself, the plots above can also be accessed on the online dashboard\footnote{\href{https://wandb.ai/dunnkers/fseval/reports/Final-results-MSc-Thesis--Vmlldzo3ODEyODE}{https://wandb.ai/dunnkers/fseval/reports/Final-results-MSc-Thesis--Vmlldzo3ODEyODE}}. The dashboard lets the user see the data in more detail by the use of zooming and tooltips. Moreover, the visualizations contain hyperlinks, taking the user directly to a detailed page of a single benchmark run.

\clearpage
\section{Discussion}\label{section:discussion}
In the following section, a brief discussion is held on the project result. Instead of interpreting the results directly, a broad view will be taken on the achieved results. Because the contribution of the paper consists of multiple elements, each is discussed separately.

The newly proposed \textbf{evaluation methodology} consists out of two main aspects. On the one hand, new evaluation metrics were created and proposed, making use of \gls{apriori} knowledge on relevant features. On the other hand, recommendations were made for the inclusion of existing metrics in the evaluation process of feature- rankers and selectors. Like it was discussed in the related work- and motivation sections, there is an interest in creating better evaluation metrics. There indeed exists a gap in the literature which is to be filled: a discussion on the usefulness of new evaluation metrics w.r.t. the feature- ranking and selection benchmarking process. Seemingly, the new metrics are of some use, and can be used to foretell the validation performance. It must be said, however, that the new evaluation metrics must probably always be accompanied by real-world data analysis with the use of validation estimators, in order to confirm the found results using the \gls{apriori} knowledge.

An important caveat in the newly proposed evaluation metrics is the assumption of \textit{uniform} feature importance. In the synthetically generated \textit{classification} datasets, grouped as the `Synclf' datasets, the informative features are known \gls{apriori}, thanks to the \texttt{make\_classification} function in sklearn. It is assumed, however, that all informative features do weight equally, i.e., have a uniform distribution w.r.t. feature importance. All rankers are scored accordingly to this uniformly distributed feature importance scores. This assumption might unjustly rank feature rankers as being less performant. This happens, for example, when a ranker assigned many small scores to irrelevant features and thus also a smaller score to the relevant features - but still getting the feature ranking (order) correct. When combining the \gls{apriori} scores with the validation scores, however, a clearer picture be obtained. Nonetheless, for the reason stated above, caution should be taken into interpreting the R\textsuperscript{2}- and log loss scores for the synthetically generated classification datasets. What is certainly still useful to practitioners, however, are the charts obtained by plotting the estimated feature importances against the ground-truth feature importances. Such charts can help in qualitative interpretation of the algorithm's performance. Also, the mean validation scores have been shown to be of good use in the evaluation process - which is also a newly proposed metric.

In the case of the synthetically generated \textit{regression} datasets, this caveat does not hold. This is because in the case of synthetically generated regression datasets, the actual ground-truth desired feature importance coefficients are known. This is due to the \texttt{make\_regression} function. This is a big plus for the `Synreg' datasets, giving them an extra amount of trustworthiness.

Next, the created \textbf{pipeline} is discussed. Arguably, this is the biggest contribution of the project. The initial goals before building the pipeline were several. The pipeline had to be flexible, to support many different feature rankers, datasets, and validation estimators. It was also desired to have the data- collection and visualization steps largely automatized, such that new experiments could be added easily. And all this was desired to be scalable, allowing the pipeline to be run on many processors and on many machines simultaneously. In the pipeline implementation that was built, all above features were addressed. When the pipeline is run, a practitioner has to complete only little steps to visualize and interpret the data, due to integration with the online dashboard. The pipeline is open-source and released as a PyPi package, such that a practitioner can also easily extend the pipeline to their own needs.

A noteworthy remark about the pipeline is to be made, however. The pipeline might have tried to tackle too many problems in a single codebase. Although the pipeline is, in the end very flexible, lots of code exists to support both regression- and classification datasets. This does make the pipeline more useful for a wider audience, but also forces the programmer to give in to simplicity. Evaluation metrics have to be implemented twice and validation estimators have to account for both learning tasks. All result data also has to be kept apart such not to intermingle the two. An upside of having included both learning tasks, however, is that it is now relatively easy to add support for another learning task. Since the codebase already deals with \textit{multiple} learning tasks, the difference between choosing between two and more is not big. Learning tasks such as clustering could be supported in this way.

Lastly, the \textbf{experiment} is discussed. The experiment is to be an implementation of the proposed methodology, and an application of the built pipeline. Although at first outset, the goal in this paper was to benchmark as many as possible feature rankers, at a specific point in time the scope was reduced to contain a more finite set of feature rankers. This is mainly due to the fact that even though the literature on feature- rankers and selectors is very extensive, the implementations are scarce. Only in rare cases, an up-to-date software package for the designated feature- ranker or selector exists. In most cases, software packages are out dated and work with older versions of sklearn. In other cases, only an R implementation can be found.

Nonetheless, the experiment does allow a practitioner to draw useful conclusions. Still, the experiment is more extensive than what is seen in many algorithm proposals in the literature. Many datasets are included, both of the real-world and synthetic type. But most importantly, using the newly proposed evaluation methodology and pipeline implementation that is freely available, new experiments can easily be added. Even, direct comparison is possible when data is uploaded to the same dashboard - allowing this experiment to be a starting place for further experimentation. In other words, with a proof-of-concept experiment in place, the field is open for more comprehensive experiments.

\clearpage
\section{Conclusion}\label{section:conclusion}
First, a concluding note will be given. Lastly, possible points of improvement for future work will be elaborated upon.

\subsection{Concluding note}
In this paper, it was investigated how best to evaluate feature- rankers and selectors. A selection of literature was considered, and their evaluation methods were compared. Traditionally, papers paid little attention to using \gls{apriori} knowledge nor performing stability analysis. Given this lack, one can speak of a gap in the literature - there is an interesting opportunity for investigating new methods.
After this opportunity was identified, new evaluation methods were proposed. In this way, a new methodology for evaluating feature- rankers and selectors was compiled in this paper. The main opportunities lie in the use of \gls{apriori} knowledge, performing stability analysis, and summarizing the validation performance in a novel way using a scalar metric.

An extensive benchmarking pipeline was constructed, implementing this new evaluation methodology. The pipeline is scalable, modular, and easy to configure. The pipeline is available as open-source free software, called `fseval'. A comprehensive experiment was also run using the pipeline, illustrating the pipeline capability's in a concrete benchmark. Using an online dashboard, the benchmark can easily be extended in the future. The results can also be interactively explored. Strong overall performers for classification datasets are Decision Tree, XGBoost, FeatBoost, and Infinite Selection. 

Feature ranking is an ever more relevant problem in a world where data and machine learning pose a prevalent role, useful for both feature selection and interpretable AI. With many feature ranking algorithms available, a comprehensive analysis is required to pick a suitable method given the context. But, when all relevant facets are highlighted and analyzed in a comprehensive feature ranking evaluation pipeline, authors and users can be more deliberate in arguing for any algorithm’s superiority.

\subsection{Limitations and Future work}
In the future, authors might build upon the work done in this paper. The following outlines the limitations of this paper and ideas for extending the work done in this project.

\textbf{Limitations} of the paper were the following.
\begin{itemize}
    \item The proposed methodology, including new evaluation metrics, could have enjoyed more mathematical argumentative support. Currently, the evaluation metrics \textit{are} outlined mathematically, but no theoretical predictions about their behavior are made. Instead, their usefulness is proven only in the experiment section, through empirical observation. Quite possibly, the usefulness of certain evaluation metrics could have been foretold mathematically.
    \item Another important caveat is the fact that the considered feature rankers were not hyper-parameter optimized for the experiment datasets. All feature rankers were run at their default settings. Fair to say, this completely invalidates some rankers. Since some rankers do not function properly without hyper-parameter optimization, their performance can be misleadingly bad. In this experiment, this is the case with Boruta, for example. One could also argue, however, that the algorithms should be able to function well at their defaults. A practitioner does not always have the luxury to perform a computationally expensive hyper-parameter optimization process.
    \item The feature support and feature rankings were less extensively evaluated than feature importance estimations. For example, the feature support estimations could have also had a more sophisticated metric devised for its evaluation. Currently, only the mean validation scores of the feature subsets are taken into account, but ideally, also the \textit{amount of selected features} would have been taken into account in the metric.
    \item Lastly, the experiment could have been more extensive. Currently, only three datasets are multioutput, meaning the rankers supporting multioutput get little comparison material. All these datasets were also of regression type. Given enough processing power, ideally the entire OpenML-CC18 benchmark suite would be run. Moreover, more synthetic datasets would be tested using a range of parameter settings.
\end{itemize}

\textbf{Ideas} for future work are several.
\begin{itemize}
    \item The pipeline can be extended in several straight-forward ways. (1) Firstly, more dataset adapters could be added. An interesting platform to support is Kaggle, although adapters for loading local- or remote CSV or JSON would also be useful. (2) Secondly, the pipeline could integrate with more back-ends aside from WandB. OpenML can also be used, for example, to upload metrics and experimental results to. Alternatively, data could be uploaded directly to a database of some kind, like MySQL. In this way, by adding more integrations to the existing benchmarking pipeline, the framework could become a truly versatile benchmarking tool for any feature ranking algorithm. This also the last pipeline improvement idea. (3) Lastly, one might even extract the general pipeline and benchmarking capabilities of the framework and use them for general-purpose \gls{ml} benchmarking. This would turn the framework from being just a feature ranking evaluation framework into a generic framework for testing \gls{ml} algorithms. 
    \item The evaluation process with regards to the feature importances ground-truth can enjoy more research. One idea is to normalize the feature importances vectors $\boldsymbol{w}$ and $\boldsymbol{\hat{w}}$ by a \textit{softmax} operation, instead of normalizing only by the sum of the vector. Using the current normalization method, negative values are not allowed in the feature importance vector. Although this is rare, it might occur. Furthermore, authors might also want to try to \textit{weight} the apriori scorings such as the R\textsuperscript{2} and log loss scores. A weighting could make sure, then, that the relevant features are assigned more weight in the scoring process, i.e., algorithms are rewarded more for getting those features right rather than the irrelevant features.
    \item Another idea is to apply the new evaluation methodology to interpretable AI algorithms in a more extensive way. Currently, only one method related to interpretable AI was included, TabNet. No special considerations were made for such rankers, but many interpretable AI methods support ranking features on a per-instance basis. Instance-based feature importance scoring is not considered in this paper's context. Because, however, the goal of ranking feature importances is similar in the domain of interpretable AI, the benchmarking framework can be made compatible relatively easily with instance-based methods.
\end{itemize}

\clearpage
\bibliographystyle{unsrtnat}
\bibliography{references}

\clearpage
\begin{appendices}
\section{Experiment line-up}\label{section:appendix-experiment-lineup}

\renewcommand\theadalign{bl}
\begin{table}[ht]
    \centering
    \begin{tabular}{| l | l | l | l | l | l | l |}
    \hline
    \thead{Name method} & \thead{Classif-\\ication} & \thead{Regr-\\ ession} & \thead{Multi-\\output} & \thead{Feature \\ importance} & \thead{Feature\\support} & \thead{Feature\\ranking} \\
    \hline
    ANOVA F-value (\href{https://scikit-learn.org/stable/modules/generated/sklearn.feature_selection.f_classif.html}{sklearn}) & \checkmark & \checkmark &  & \checkmark &  &  \\ 
    \hline
    Boruta \citep{kursa_feature_2010} & \checkmark &  &  &  & \checkmark & \checkmark \\ 
    \hline
    Chi-Squared (\href{https://scikit-learn.org/stable/modules/generated/sklearn.feature_selection.chi2.html}{sklearn}) & \checkmark &  &  & \checkmark &  &  \\ 
    \hline
    Decision Tree (\href{https://scikit-learn.org/stable/modules/generated/sklearn.tree.DecisionTreeClassifier.html}{sklearn}) & \checkmark & \checkmark & \checkmark & \checkmark &  &  \\ 
    \hline
    FeatBoost (\href{https://github.com/amjams/FeatBoost}{github}) & \checkmark &  &  & \checkmark & \checkmark &  \\ 
    \hline
    Infinite Selection \citep{roffo_infinite_2015} & \checkmark &  &  & \checkmark &  & \checkmark \\ 
    \hline
    MultiSURF \citep{urbanowicz_benchmarking_2018} & \checkmark & \checkmark &  & \checkmark &  &  \\ 
    \hline
    Mutual Info \citep{zaffalon_robust_2014} & \checkmark & \checkmark &  & \checkmark &  &  \\ 
    \hline
    ReliefF \citep{urbanowicz_benchmarking_2018} & \checkmark & \checkmark &  & \checkmark &  &  \\ 
    \hline
    Stability Selection \citep{meinshausen_stability_2009} & \checkmark &  &  & \checkmark & \checkmark &  \\ 
    \hline
    TabNet \citep{arik_tabnet_2020} & \checkmark & \checkmark & \checkmark & \checkmark &  &  \\ 
    \hline
    XGBoost \citep{chen_xgboost_2016} & \checkmark & \checkmark &  & \checkmark &  &  \\ 
    \hline     
    \end{tabular}
    \caption{Feature Ranker line-up. Both classifiers and regressors are considered, as well as multioutput estimators.}
    \label{table:experiments-ranker-specification}
\end{table}

\renewcommand\theadalign{bl}
\begin{table}[ht]
    \centering
    \begin{tabular}{| l | l | l | l | l | l | l |}
    \hline
    \thead{Name dataset} & \thead{$n$} & \thead{$p$} & \thead{Task} & \thead{Multi-\\output} & \thead{Domain} & \thead{Group} \\
    \hline
    \makecell[tl]{Boston house \\prices} & 506 & 11 & Regression & No & Finance & - \\ 
    \hline
    Additive & 10000 & 10 & Regression & Yes & Synthetic & Chen et al. \citep{chen_learning_2018} \\ 
    \hline
    Orange & 10000 & 10 & Regression & Yes & Synthetic & Chen et al. \citep{chen_learning_2018} \\ 
    \hline
    XOR & 10000 & 10 & Regression & Yes & Synthetic & Chen et al. \citep{chen_learning_2018} \\ 
    \hline
    \makecell[tl]{Climate Model \\Simulation} & 540 & 18 & Classification & No & Nature & OpenML-CC18 \citep{bischl_openml_2019} \\ 
    \hline
    Cylinder bands & 5456 & 24 & Classification & No & Mechanics & OpenML-CC18 \citep{bischl_openml_2019} \\ 
    \hline
    Iris Flowers & 150 & 4 & Classification & No & Nature & - \\ 
    \hline
    Madelon & 2600 & 500 & Classification & No & Synthetic & Guyon \citep{guyon_design_2003} \\ 
    \hline
    Multifeat Pixel & 2000 & 240 & Classification & No & OCR & OpenML-CC18 \citep{bischl_openml_2019} \\ 
    \hline
    Nomao & 34465 & 89 & Classification & No & Geodata & OpenML-CC18 \citep{bischl_openml_2019}\\ 
    \hline
    Ozone Levels & 2534 & 72 & Classification & No & Nature & OpenML-CC18 \citep{bischl_openml_2019} \\ 
    \hline
    Phoneme & 5404 & 5 & Classification & No & Biomedical & OpenML-CC18 \citep{bischl_openml_2019} \\ 
    \hline
    Synclf easy & 10000 & 20 & Classification & No & Synthetic & Synclf \\ 
    \hline
    Synclf hard & 10000 & 50 & Classification & No & Synthetic & Synclf \\ 
    \hline
    Synclf medium & 10000 & 30 & Classification & No & Synthetic & Synclf \\ 
    \hline
    Synclf very hard & 10000 & 50 & Classification & No & Synthetic & Synclf \\ 
    \hline
    Synreg easy & 10000 & 10 & Regression & No & Synthetic & Synreg \\ 
    \hline
    Synreg hard & 10000 & 20 & Regression & No & Synthetic & Synreg \\ 
    \hline
    Synreg medium & 10000 & 10 & Regression & No & Synthetic & Synreg \\ 
    \hline
    Synreg very hard & 10000 & 20 & Regression & No & Synthetic & Synreg \\ 
    \hline
    Texture & 5500 & 40 & Classification & No & \makecell[tl]{Pattern \\Recognition} & OpenML-CC18 \citep{bischl_openml_2019} \\ 
    \hline
    \makecell[tl]{Wall Robot \\Navigation} & 5456 & 24 & Classification & No & Mechanics & OpenML-CC18 \citep{bischl_openml_2019} \\ 
    \hline
    \end{tabular}
    \caption{All datasets considered in the experiment. Both real-world and synthetic datasets are considered. Whereas synthetic datasets are generated in the pipeline itself, real-world datasets are fetched from OpenML.}
    \label{table:experiments-dataset-specification}
\end{table}

\begin{figure}[ht]
    \centering
    \includegraphics[width=\linewidth]{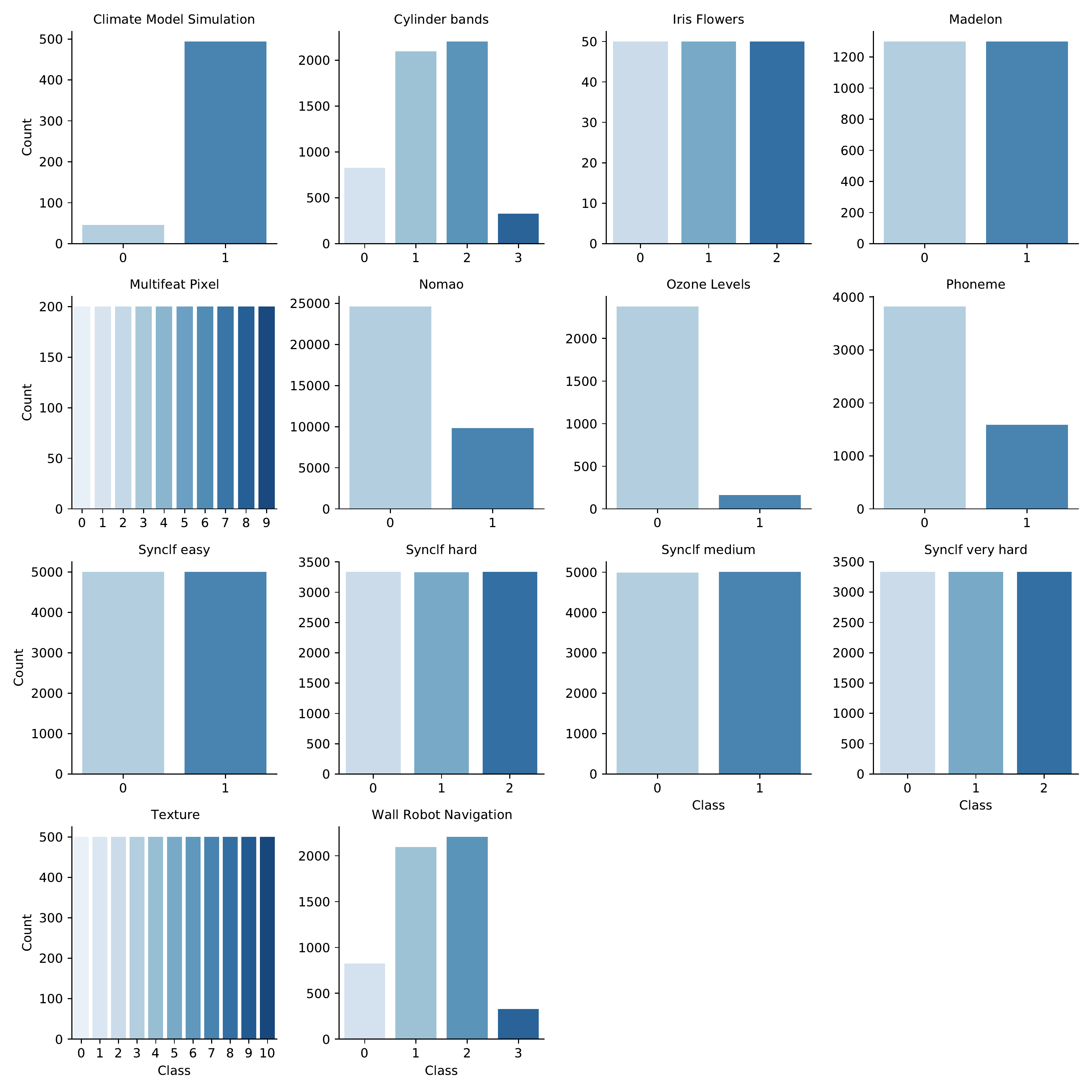}
    \caption{Class-distributions of the classification datasets described in Table~\ref{table:experiments-dataset-specification}. The x-axis shows the target classification classes and the y-axis shows the amount of samples with that target class. No binning was applied, the distributions are like shown.}
    \label{fig:experiments-datasets-class-distributions}
\end{figure}

\clearpage
\section{Additional results}\label{section:appendix-additional-results}

\begin{figure}[ht]
    \centering
    \includegraphics[width=\linewidth]{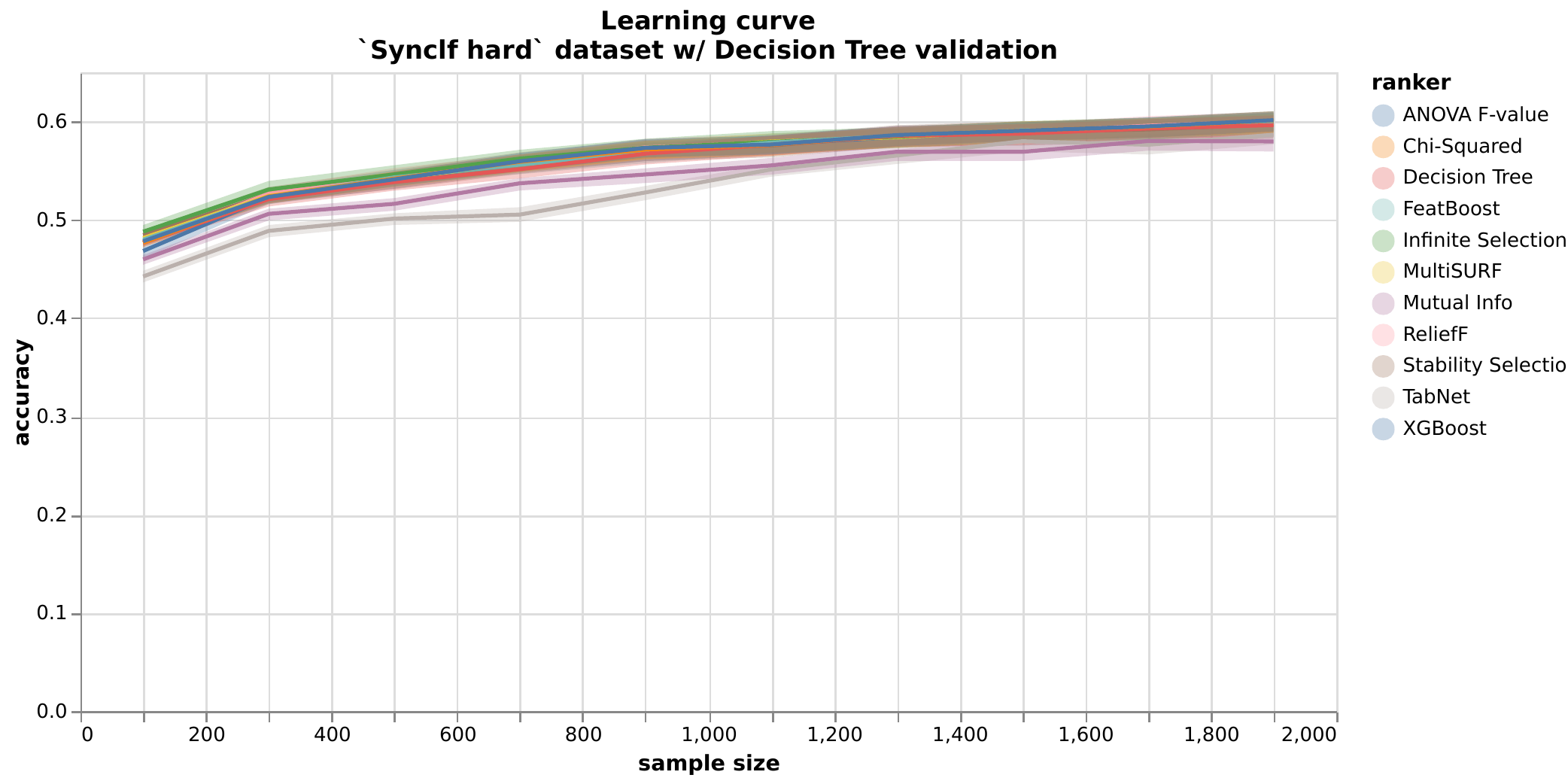}
    \caption{Learning curve for `Synclf hard' dataset. Sample sizes from 100 to 2,100 were used, with intervals of 200. The average accuracy over 10 bootstraps is shown.}
    \label{fig:results-learning-curve}
\end{figure}

\begin{figure}[ht]
    \centering
    \includegraphics[width=\linewidth]{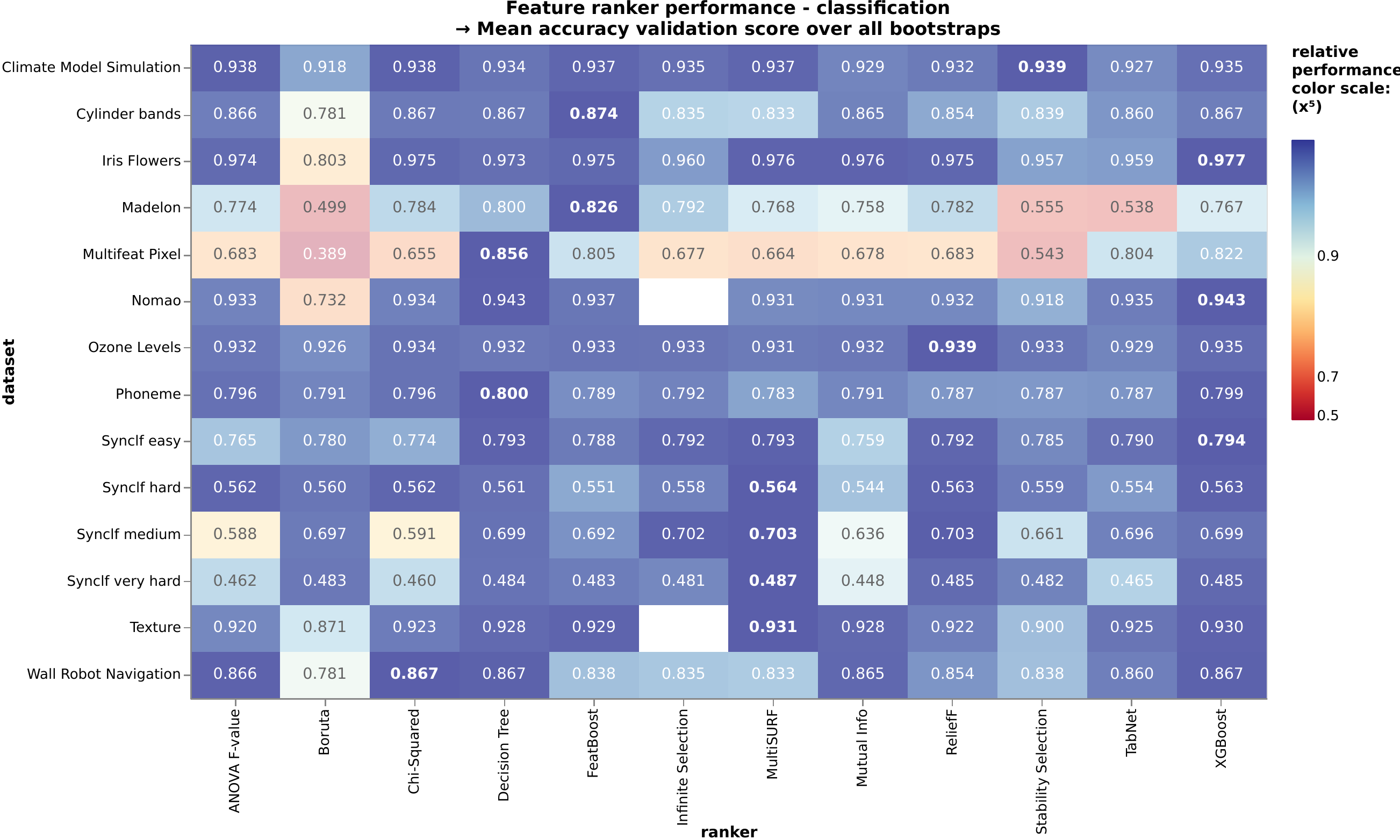}
    \caption{\textbf{\gls{knn}} mean validation scores for all \textbf{classification} datasets. Color scale is configured with \textit{relative performance}, like explained in Section~\ref{section:experiments-example}.}
    \label{fig:results-all-datasets-mean-validation-knn}
\end{figure}

\begin{figure}[ht]
    \centering
    \includegraphics[width=\linewidth]{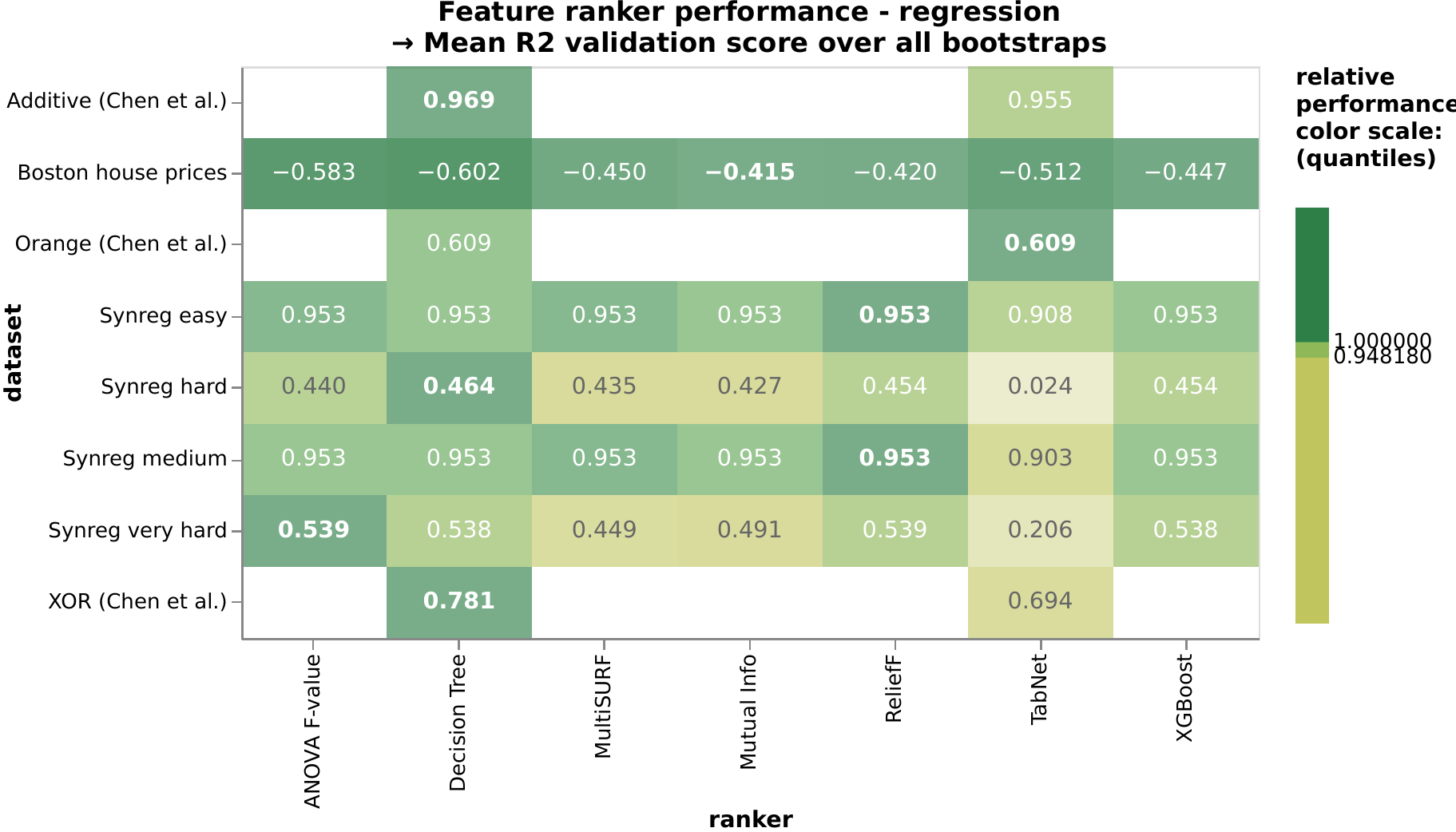}
    \caption{\textbf{\gls{dt}} mean validation scores for all \textbf{regression} datasets. Color scale is configured with \textit{relative performance}, like explained in Section~\ref{section:experiments-example}.}
    \label{fig:results-all-datasets-mean-validation-dt-regression}
\end{figure}

\begin{figure}[ht]
    \centering
    \includegraphics[width=\linewidth]{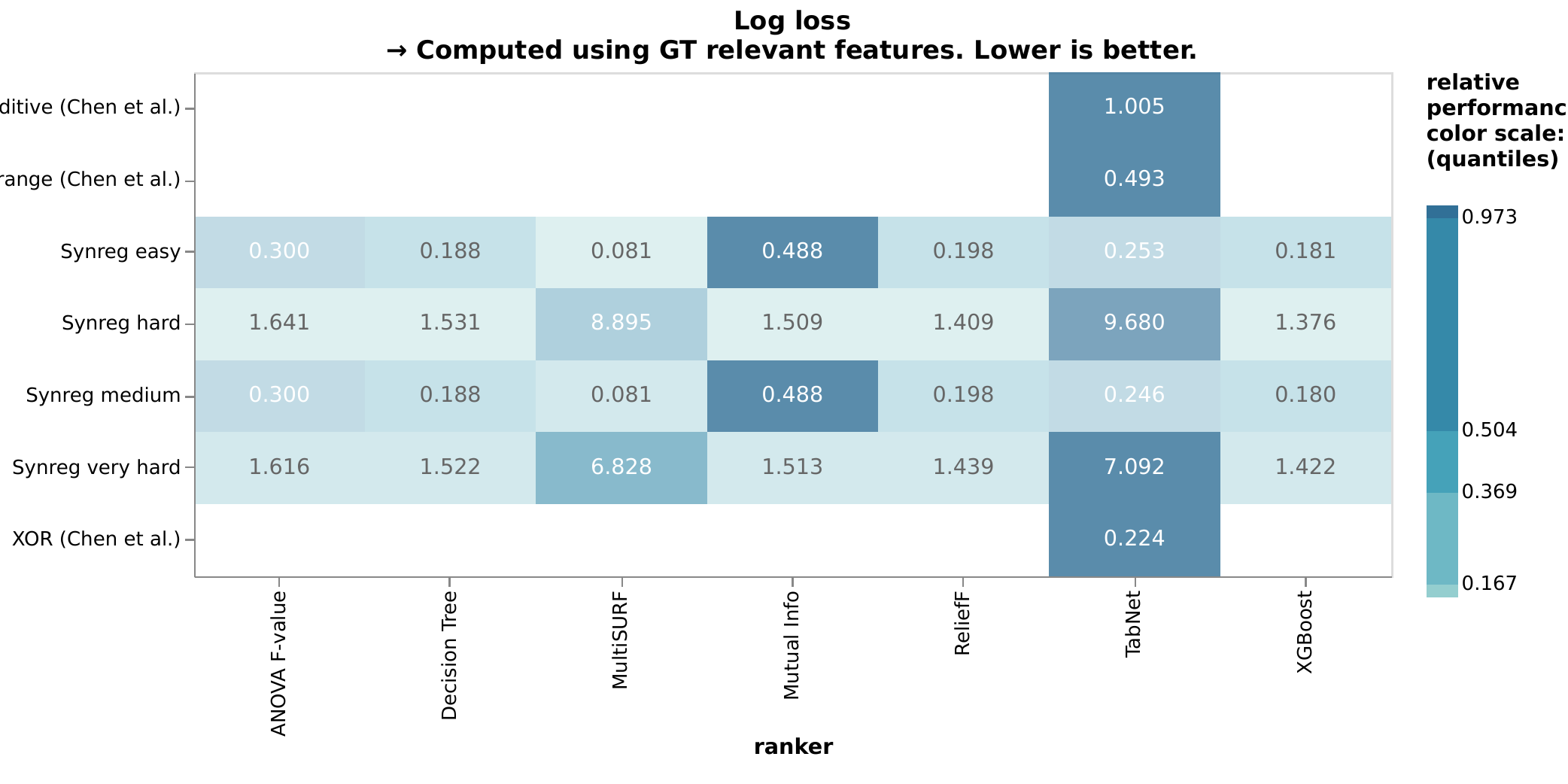}
    \caption{Algorithm Log loss scores for \textbf{regression} datasets, computed using the \gls{gt} feature importanes. The computation was explained in Section~\ref{section:evaluation-apriori-knowledge} and illustrated in Section~\ref{section:experiments-example}. A lighter shade of blue means a better score.}
    \label{fig:results-all-datasets-log-loss-regression}
\end{figure}

\begin{figure}[ht]
    \centering
    \includegraphics[width=\linewidth]{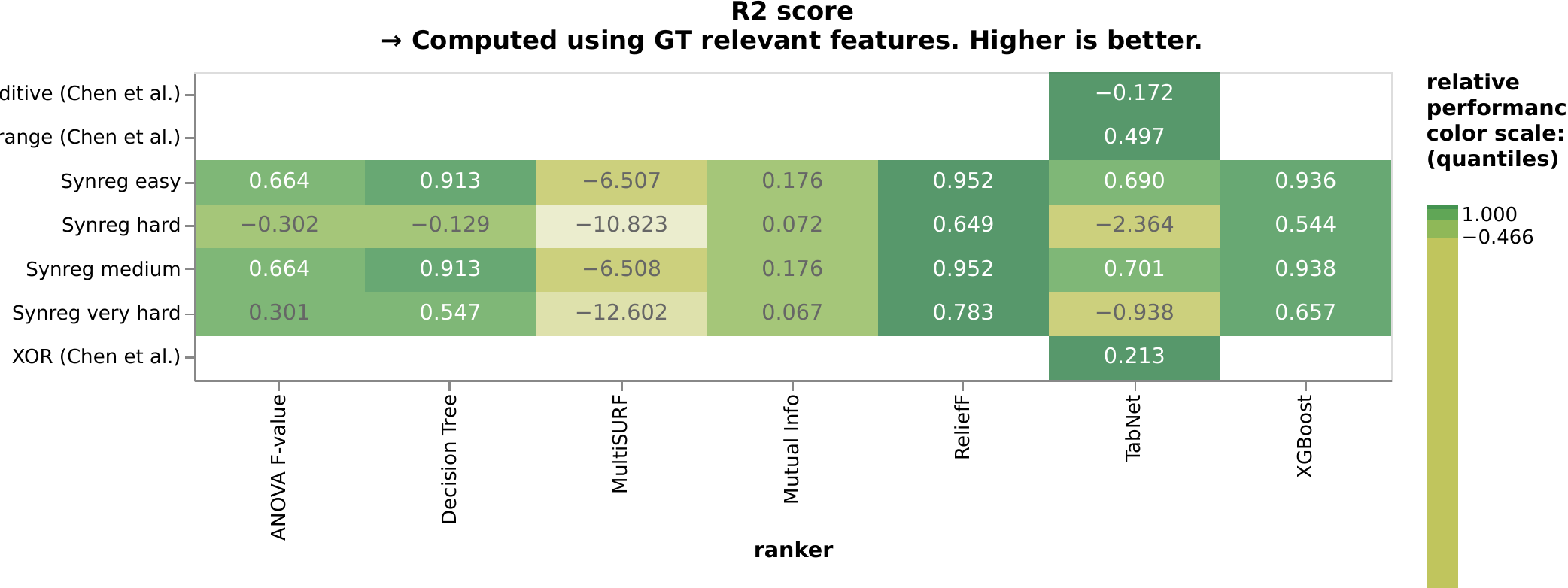}
    \caption{Algorithm R\textsuperscript{2} scores for \textbf{regression} datasets, computed using the \gls{gt} feature importanes. The computation was explained in Section~\ref{section:evaluation-apriori-knowledge} and illustrated in Section~\ref{section:experiments-example}. A darker shade of green means a better score.}
    \label{fig:results-all-datasets-r2-score-regression}
\end{figure}

\end{appendices}
\clearpage

\glsaddall
\printnoidxglossary

\end{document}